\pdfoutput=1
\documentclass[10pt,journal,letterpaper,compsoc]{IEEEtran}
\usepackage{authblk}

\usepackage[usenames,dvipsnames]{xcolor}
\usepackage{soul}

\newcommand{\be}{\begin{equation}}
\newcommand{\ee}{\end{equation}}

\usepackage{eso-pic}
\usepackage{xspace}

\usepackage{epsfig}
\usepackage{graphicx}
\usepackage[cmex10]{amsmath}
\usepackage{amssymb}

\usepackage[square,numbers,sort]{natbib}
\usepackage{wrapfig}

\usepackage{booktabs}

\usepackage[utf8]{inputenc}
\usepackage[T1]{fontenc}
\usepackage{url}
\usepackage{setspace}

\usepackage{multirow}
\usepackage{colortbl}
\usepackage{footnote}
\definecolor{lightgray}{rgb}{.9, .9, .9}

\newcommand{\E}{\mathbb{E}}
\newcommand{\V}{\mathbb{V}}
\newcommand{\R}{\mathbb{R}}
\newcommand{\N}{\mathbb{N}}
\newcommand{\Q}{\mathbf{Q}}
\newcommand{\B}{\mathbf{B}}
\newcommand{\VM}{\mathbf{V}}
\newcommand{\Z}{\mathbf{Z}}
\newcommand{\C}{\mathbf{C}}
\DeclareMathOperator*{\argmin}{argmin}
\DeclareMathOperator*{\argmax}{argmax}

\ifCLASSOPTIONcompsoc
\else
\fi
\ifCLASSINFOpdf
\else
\fi
\ifCLASSOPTIONcompsoc
\usepackage[tight,footnotesize,sf,SF]{subfigure}
\else
\usepackage[tight,footnotesize]{subfigure}
\fi

\usepackage[font=footnotesize,labelfont=sf,textfont=sf,justification=justified,singlelinecheck=false]{caption}

\begin{document}
%
\title{Bayesian Time-of-Flight for Realtime Shape, Illumination and Albedo}

\author[1]{Amit Adam \thanks{\tt\footnotesize amitadam@microsoft.com}}
\author[2]{Christoph Dann \thanks{{\tt\footnotesize cdann@cmu.edu}}}
\author[1]{Omer Yair \thanks{{\tt\footnotesize omeryair@microsoft.com}}}
\author[1]{Shai Mazor \thanks{{\tt\footnotesize smazor@microsoft.com}}}
\author[3]{Sebastian Nowozin  \thanks{{\tt\footnotesize Sebastian.Nowozin@microsoft.com}}}
\affil[1]{\small Microsoft AIT - Advanced Imaging Technologies, Haifa, Israel}
\affil[2]{\small School of Computer Science, CMU, Pittsburgh PA}
\affil[3]{\small Microsoft Research - Cambridge, Cambridge UK}

%
%

\markboth{~}%
{Adam \MakeLowercase{\textit{et al.}}: Bayesian Time-of-Flight}
%


\IEEEcompsoctitleabstractindextext{%
\begin{abstract}
We propose a computational model for shape, illumination and albedo inference
in a pulsed time-of-flight (TOF) camera.
%
In contrast to TOF cameras based on phase modulation, our camera enables general exposure profiles.
This results in added flexibility and requires novel computational approaches.
%
%
To address this challenge we propose a generative probabilistic model that
accurately relates latent imaging conditions to observed camera responses.
%
%
While principled, realtime inference in the model
turns out to be infeasible, and we propose to employ efficient non-parametric regression trees
to approximate the model outputs.
As a result we are able to provide, for each pixel, at video
frame rate, estimates and uncertainty for \emph{depth, effective albedo, and ambient
light intensity}.
These results we present are state-of-the-art in depth imaging.
%
%
The flexibility of our approach allows us to easily enrich our generative model. We
demonstrate that by extending the original single-path model to a two-path model,
capable of describing some multipath effects. The new model is seamlessly
integrated in the system
at no additional computational cost.
%
%
Our work also addresses the important question of optimal exposure design in pulsed
TOF systems.
Finally, for benchmark purposes and to obtain realistic empirical priors of
multipath and insights into this phenomena, we propose a physically accurate
simulation of multipath phenomena.
%
%
\end{abstract}
}

\maketitle

\IEEEdisplaynotcompsoctitleabstractindextext

%
\IEEEpeerreviewmaketitle


\section{Introduction}
The commercial success of depth cameras in recent years has enabled numerous
computer vision applications.  Notable applications are human pose
estimation~\cite{DBLP:conf/cvpr/ShottonFCSFMKB11,DBLP:journals/pami/ShottonGFSCFMKCKB13},
dense online 3D reconstruction of an
environment~\cite{DBLP:conf/ismar/NewcombeIHMKDKSHF11}, 
and other uses---an overview is available in a recent special
issue~\cite{DBLP:journals/tcyb/ShaoHXS13} and in the review
article~\cite{han2013computervisionkinect}.

Broadly speaking we may differentiate between depth cameras based on
triangulation and cameras which estimate depth based on time of flight
(TOF)~\cite{lefloch2013toffoundation,hansard2012tof-book}.
%
Furthermore, while in the context of TOF the cameras often operate using
modulated illumination and sensing, and the computational methods usually
employ phase-space reasoning~\cite{hansard2012tof-book}, in this paper we take
a different approach which we now describe.
%

\begin{figure}[t!] \centering
\includegraphics[width=0.995\linewidth]{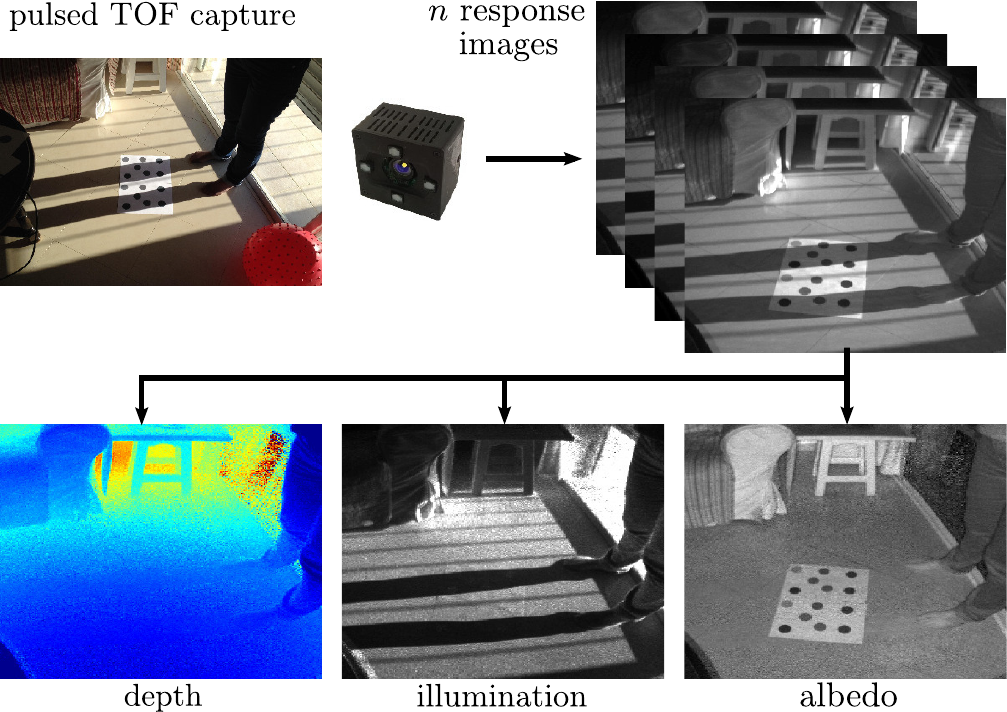}%
\caption{\footnotesize System overview: the inputs are $n$ pulsed TOF response images,
obtained concurrently using different exposure profiles.
In realtime (30fps) we separate \emph{depth}, \emph{ambient illumination}
and \emph{effective albedo} at every pixel.}
\label{fig:page1}%
\end{figure}

%
%
Figure~\ref{fig:page1} describes the inputs and outputs of our system.
We start with $n$ concurrently captured intensity images obtained under active
illumination of the scene, using $n$ different exposure profiles.
%
Using these $n$ observations at every pixel, we infer the depth,
reflectivity, and ambient lighting conditions.
We achieve this by using a generative probabilistic model that relates the
unknown imaging conditions---\emph{shape}, \emph{illumination} and
\emph{albedo}---to the per-pixel camera observations.
To perform inference we use either Bayesian inference or maximum likelihood
estimation.

However, achieving realtime video rate by direct application of these
inference methods is infeasible under practical constraints on computation.
Therefore we use an approach inspired by model
compression~\cite{bucila2006modelcompression} and approximate the accurate but
slow inference methods using regression trees, a fast non-parametric
regression method~\cite{breiman1984cart}.
%

The regression approach has two advantages;
first, it allows to approximate inference in principled probabilistic models
under tight compute and memory constraints;
second, it decouples the model from the runtime implementation, allowing
continuing improvements in the model without requiring changes to the
test-time implementation.

We demonstrate this important advantage in Section \ref{sec:twopath} where
we extend our generative model to a richer model which considers multipath effects.
Our decoupling of model+inference from runtime regression allows us seamless switching
between different generative models, at no additional computational cost.
To the best of our knowledge no other depth cameras have used a statistical
regression approach for online depth inference.


No matter which model we use for inference, at times there will be pixels that the model fails to explain.
Common reasons are mixed pixels (imaging a depth discontinuity), sensor saturation, complex
multipath, interference from another active device, or extreme image noise.
We propose a robust fit-to-model score that can be used to detect and invalidate
affected pixels from further processing.

Having described inference and pixel invalidation, we address an orthogonal but important
question in pulsed TOF systems: exposure profile design.
We are flexible to choose exposure profiles and we directly optimize the
expected accuracy of inferred depth using Bayesian decision theory.  This
yields a challenging optimization problem and we propose an approximate
solution.

Finally, we introduce an accurate TOF simulation procedure based on physically
accurate light transport simulation.
We use this capability for both exposure design and for synthetic but physically
accurate benchmarking.

\subsection{Related Work}
Most commercially available time-of-flight cameras (as of early 2015) work
using \emph{modulated} time of
flight~\cite{schwarte1997pmd,lange2001phasetof},
also known as \emph{phase-based} time-of-flight.
They generate a sinusoidal illumination signal and measure correlation of the
reflected signal with a sinusoidal-profiled gain function of the same
frequency, delayed by a phase shift~\cite{gupta2014phasorimaging}.
For a fixed frequency and phase shift a recorded frame does not contain
sufficient information to reconstruct depth.
Therefore, modern systems typically record a sequence of frames at multiple
frequencies and multiple phase shifts and use the combined set of frames to
unambiguously infer depth using so called \emph{phase unwrapping}
algorithms~\cite{hansard2012tof-book,gupta2014phasorimaging}.

In contrast, our camera uses \emph{pulsed} TOF, also known as \emph{gated}
TOF. This technology has differentiators in terms of hardware-related aspects (size, power, resolution)
which are not relevant here, but let us highlight an important computational aspect of this
camera: in contrast with the sine-like gain functions used in modulated TOF,
we are allowed to choose from a large space of possible gain functions.
Hence more general inference methods are required.
%

Our work on optimizing the gain profiles has not been addressed in pulsed TOF
systems, but for modulated TOF prior
work~\cite{payne2011waveformoptimizationtof} has attempted to optimize the
illumination profile to improve depth accuracy.

\textbf{Shape, Illumination, and Reflectance.}
Recovering the imaging conditions leading to a specific image---the inverse
problem of imaging, is a long standing goal of computer vision.
A recent modern treatment of this problem has been given in
\cite{DBLP:conf/cvpr/BarronM12,DBLP:conf/cvpr/BarronM13,BarronMalikTR2013,gehler2011intrinsicimages}, with a comprehensive
historical review. Conceptually the approach in these works is similar to ours: find the most likely
shape, illumination and albedo to give rise to the observed image. In contrast with these works, we do the inference at
the pixel level and not the image level, being able to do so due to the unique imaging process we employ.
Additionally our regression approach allows this inference to be done in realtime.
 Moreover, our shape output
actually gives the full posterior depth distribution. This allows direct usage
of our depth in incremental estimators or integrators such as
\cite{DBLP:conf/ismar/NewcombeIHMKDKSHF11} 
that specifically take care to  maintain the state distribution at all times
\cite{DBLP:journals/ijcv/IsardB98,thrun2005probabilistic}.

In the context of illumination estimation,  we remark that there have been specific works on shadow
removal~\cite{finlayson2002removingshadows,xiao2014shadowremovalrgbd}, which
is a nice byproduct of our approach (see Figure~\ref{fig:page1}).

\textbf{Multipath Interference.}
Multiple reflections (multipath) commonly occur in real
scenes~\cite{lefloch2013toffoundation}.
There is now a solid body of work on handling multipath in modulated TOF
systems, but to the best of our knowledge there is no published work on
handling multipath in pulsed TOF cameras.

We briefly discuss work that exists for modulated TOF and relate it to our
proposed solution.
The work of~\cite{fuchs2010multipath,fuchs2013multipath}
and~\cite{jimenez2014multipath}
model the light reflections in the scene globally to improve depth inference.
To do this, they assume planar Lambertian surfaces and iteratively minimize an
energy function.
The methods work in important settings but the expensive minimization
procedure precludes a realtime implementation.
%
The
work~\cite{dorrington2011twopath,godbaz2012twopath,DBLP:conf/icmcs/KirmaniBC13}
assumes two-path interference from close-to-specular surfaces.
The resulting methods are practical and efficient and our approach in
Section~\ref{sec:twopath} makes similar model assumptions.
However, we work with different signals (pulsed TOF) and also provide a
probabilistic model with uncertainty estimates.
%
The work~\cite{DBLP:conf/eccv/FreedmanSKLS14} generalizes the above two-path
models to signals which arise from either two-path specular or two-path
Lambertian reflectors; these signals are ``compressible'' and can hence be
described with few parameters; the resulting method can be implemented in real
time.
%
Another recent branch of literature originating from work on transient
imaging~\cite{wu2012globallighttransport} uses modulated TOF imaging with
Fourier-based reconstruction of the time-dependent light density.
The work of~\cite{bhandari2014multipath,lin2014multipath} reconstructs
the transient light density for each pixel from a large number of modulated
TOF images, each with a different modulation frequency.  While this line of
work could inspire practical multipath techniques and is computationally
efficient, currently the large number of required frequencies (several dozen)
and the large acquisition time precludes realtime applications in dynamic
scenes.

The robust invalidation of observations seems to have not been considered
before with the exception of~\cite{DBLP:conf/eccv/FreedmanSKLS14} who provide
an adhoc method for invalidation.  Because we use a sound probabilistic model
we can leverage and adapt standard methods in Bayesian
practice~\cite{bayarri2000compositenull} for this purpose.

\subsection{Contributions}
To summarize and as an aide in following the paper, our novel contributions are:

\textbf{Principled Framework}

-- A probabilistic generative model for pulsed TOF imaging;

-- Principled inference of all latent imaging conditions, given camera observations

-- Accurate depth uncertainty estimates;

-- Robust Bayesian per-pixel invalidation for outlier observations;

\textbf{Practicalities}

-- A novel use of regression to enable realtime inference under
tight compute and memory constraints;

-- Complete decoupling of runtime mechanism from model and inference

\textbf{Extensibility and Multipath}

-- A probabilistic model for depth inference in the presence of simple
multi-path;

\textbf{Results}

-- Experimental results showing robust video-rate inference of shape, illumination and
reflectance, both indoors and outdoors at direct sunlight;

\textbf{Computational Photography and Tools}

-- Design of exposure profiles to directly optimize depth accuracy under
task-derived imaging conditions;

-- A novel physically-based renderer for TOF simulation

-- Use of the simulator for both exposure design and benchmarking


\section{Modeling the Imaging Process}
We start with our camera's principle of operation, then formulate a
generative model relating the unknown imaging conditions to the observable
camera outputs.
Assume that a specific pixel images a point at a certain distance and denote by
$t$ the time it takes light to travel twice this distance.
The reflected signal is integrated at the pixel using a gain determined by a
shutter signal $S(\cdot)$.
If $P(\cdot)$ is the emitted light pulse, the reflected pulse arriving after
time $t$ is $P(u-t)$.
The observed response due to the reflected light pulse is
\be
	R_{\textrm{active}} = \int S(u) \: \rho \: P(u-t) \: d(t) \,\textrm{d}u. \label{active_part}
\ee

Here $\rho$ denotes the effective reflectivity%
\footnote{We use both the terms albedo and reflectivity. The quantity $\rho$
we use in the model actually contains the effect of foreshortening and
therefore we refer to effective reflectivity/albedo.}
of the imaged point,
and $d(t) = \frac{1}{t^2}$ denotes decay of the reflected pulse due to distance.
Therefore, the reflected pulse is downscaled by a factor of
$\rho \: d(t)$.
The quantity $\rho P(u-t) d(t)$ is integrated with an exposure-determined gain
$S(\cdot)$.

Let us now consider the effect of ambient illumination.
We denote by $\lambda$ the ambient light level falling on the imaged
point. Then the reflected light level is $\rho \lambda$, and we assume that
during the integration period, this level of ambient light is constant.
Therefore, the observed response due to ambient light is
$R_{\textrm{ambient}} = \int S(u) \: \rho \: \lambda \,\textrm{d}u$. 
%
The actual observed response is the sum of the responses due to active
illumination and due to ambient light,
\be
R = \int S(u) \: \left(\rho \: P(u-t) \: d(t) + \rho \: \lambda\right) \,\textrm{d}u. \label{response_model}
\ee
Equation~(\ref{response_model}) specifies the relationship between the unknown
imaging conditions $(t, \rho, \lambda)$ (depth, albedo, and ambient light
level), and the observation we obtain at the pixel, when using the exposure
profile $S(\cdot)$.
We \emph{concurrently} use $n$ different exposure profiles\footnote{The hardware
system enabling concurrent different exposure profiles
is described in~\cite{erez} and~\cite{patent1,patent2}.}
$S_1(\cdot),S_2(\cdot), \ldots, S_n(\cdot)$, and obtain $n$ observations as

{
\begin{eqnarray}
\!\!\!
\left[ \begin{array}{c}
R_1 \\
\vdots \\
R_n
\end{array}  \right] & \!\!\!\!\!=\!\!\!\!\! &
\rho \left[ \!\! \begin{array}{c}
\int S_1(u)  P(u-t) d(t)  \,\textrm{d}u \\
\vdots \\
\int S_n(u)  P(u-t) d(t)  \,\textrm{d}u
\end{array} \!\! \right]
\!\! + \!\! \rho \lambda \left[ \!\!\begin{array}{c}
\int S_1(u)    \,\textrm{d}u \\
\vdots \\
\int S_n(u)    \,\textrm{d}u
\end{array} \!\!\right] \nonumber\\
 & \!\!\!\!\!=\!\!\!\!\! & \rho \vec{C}(t) + \rho \lambda \vec{A}.
\end{eqnarray}
}
\noindent In short, we have the observed response vector
{
\be
\vec{R} = \rho\vec{C}(t) + \rho\lambda\vec{A}. \label{vec_response_model}
\ee
}
Here $\vec{C}(t)$ is the expected response from a point at distance equivalent
to time $t$, assuming unit reflectivity and no ambient light.
This response is scaled by the reflectivity $\rho$ and shifted in the ambient light
direction $\vec{A}$, the magnitude of the shift being the product of albedo
and ambient light level.  Equation~(\ref{vec_response_model})
is the model describing our imaging process.

We remark that $\vec{C}(\cdot)$ and $\vec{A}$ are determined by the
illumination and exposure signals and are estimated using a simple camera
calibration process which is outside the scope of this paper.


\begin{figure}[t!] \centering
	\subfigure[Response curve $\vec{C}$]{%
		\includegraphics[width=0.49\linewidth]{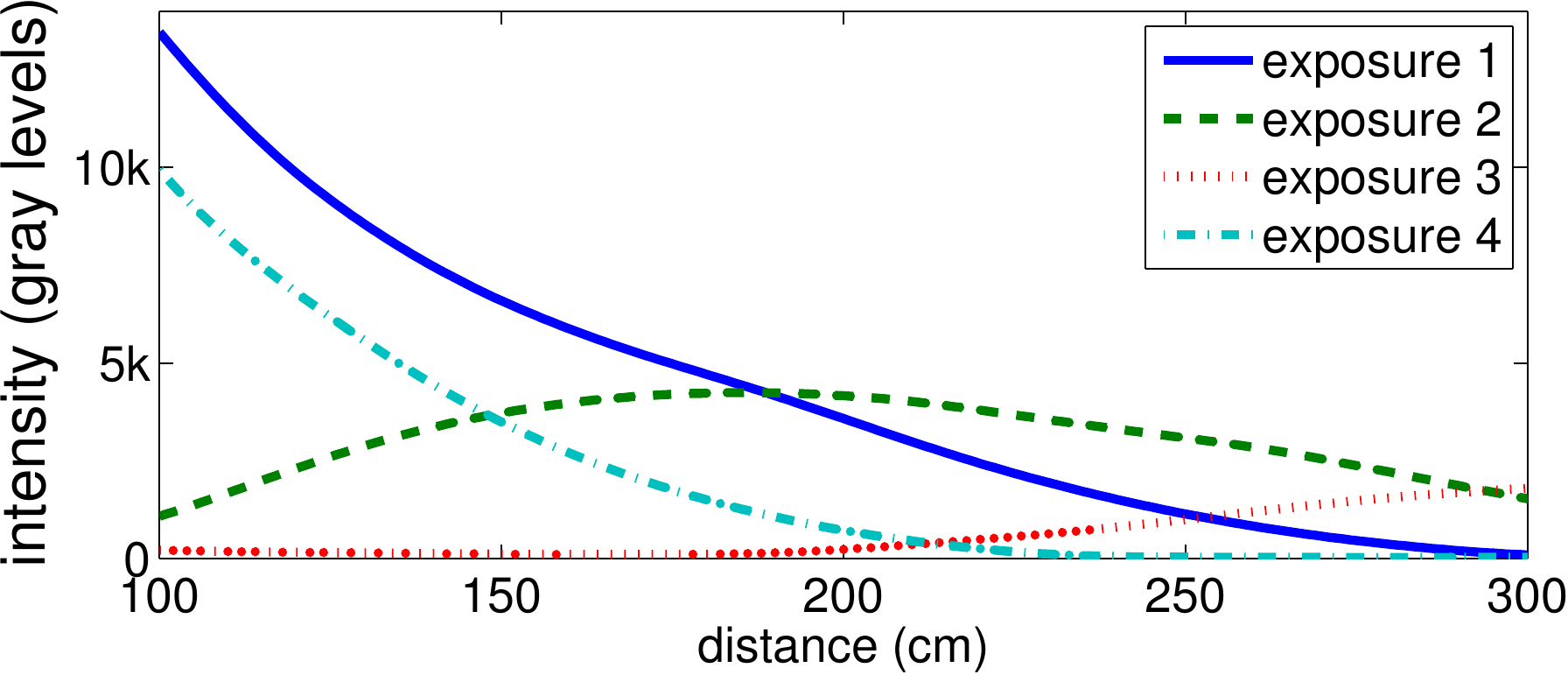}%
		\label{fig:wp-C}%
	}\hfill%
	\subfigure[$\vec{C}/d$, without decay]{%
		\includegraphics[width=0.47\linewidth]{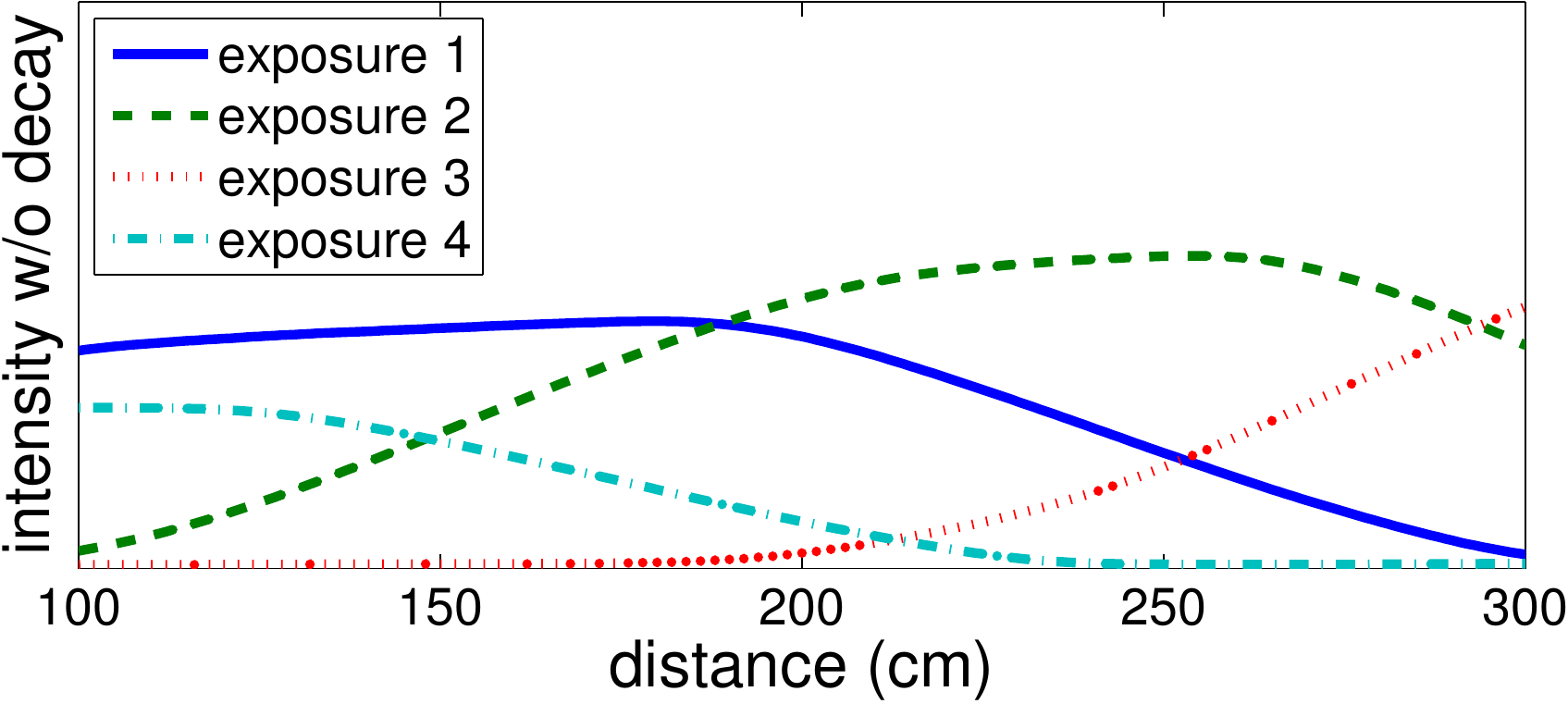}%
		\label{fig:wp-Cdc}%
	}%
\caption{\footnotesize A typical response curve.
\subref{fig:wp-C}
The actual curve $\vec{C}(\cdot)$.
As distance grows the response decays, as per equation~(\ref{c_of_t}).
\subref{fig:wp-Cdc}
Decay-compensated response where we plot $\vec{C}(t)/d(t) = t^2 \vec{C}(t)$
for more details (from now on we use decay-compensated curves for
visualization).
}
\label{fig:wp_example}%
\end{figure}

Figure~\ref{fig:wp_example} shows the curve
$\vec{C}(\cdot)$ as a function of depth $t$, for a typical exposure profile
design.
The four colored curves denote the  specific response curves of four exposure
profiles $S_1(\cdot),\ldots,S_4(\cdot)$, namely
\be
C_i(t) = \int S_i(u) \: P(u-t) \: d(t)\,\textrm{d}u. \label{c_of_t}
\ee
Looking at Figure~\ref{fig:wp_example}, consider the response vector $\vec{R}$ we may expect
from depth $t=150cm$. We see that the first (blue) coordinate should be high,
the second and fourth coordinates should be approximately equal and the third coordinate (red) should be the lowest.
In contrast, at depth $t=190cm$ the first and second entries of $\vec{R}$ should be approximately equal (blue and green).
Thus we see that by suitable design of the curve $\vec{C}(\cdot)$, we may expect to be
able to infer depth accurately using the responses we observe.

Since the response we observe is scaled by the albedo $\rho$, it
may be tempting to normalize the response vector.
However, as we discuss in the next section, the noise \emph{does} depend on the
magnitude and therefore the unnormalized response contains relevant
information for depth inference and for predicting depth uncertainty that
would be lost upon normalization.

\section{A Probabilistic Model}
\label{prob_model}
We now rephrase~(\ref{vec_response_model}) as a probabilistic model,
relating the imaging conditions $(t,\rho,\lambda)$ to a distribution over
responses $\vec{R}$.
Specifically we model $\vec{R}$ given $t$, $\rho$, $\lambda$ as
{
\be
\vec{R} \sim P(\vec{R} \mid t,\rho,\lambda),
	\label{eqn:model1}
\ee
}
where we assume that $P(\vec{R} \mid t,\rho,\lambda)$ is a multivariate
Gaussian distribution with mean vector as in~(\ref{vec_response_model}),
{
\be
\E[\vec{R} \mid t,\rho,\lambda ] = \vec{\mu}(t,\rho,\lambda)
	= \rho \: \vec{C}(t) + \rho \: \lambda \: \vec{A}, \label{generative_mean}
\ee
}
and with a diagonal covariance matrix
{
\be
\Sigma(\vec{\mu}) = \left( \begin{array}{ccc}
\alpha\mu_1 + K & & \\
 & \ddots & \\
 & & \alpha\mu_n + K
\end{array} \right).
\label{generative_cov}
\ee
}
Here $K$ is related to \emph{read  noise}---noise that is part of the system even
when no light is present.  The affine relationship between the magnitude of the
response and its variance is due to \emph{shot noise} and is well known
\cite{healey1994ccdnoise,DBLP:journals/tip/FoiTKE08}.
We validate this noise model experimentally in the supplementaries.

The generative model~(\ref{eqn:model1}) is the distribution of the
observed $\vec{R}$ at a pixel given the imaging conditions.
We would like to \emph{infer} the imaging conditions depth $t$, reflectivity
$\rho$ and ambient light level $\lambda$ given the observation $\vec{R}$.
There are three mainstream approaches for doing so, namely Bayesian posterior
inference, maximum likelihood estimation (MLE), and maximum aposteriori (MAP)
estimation.

\subsection{Bayesian Inference}
We assume certain priors on depth, reflectivity and ambient light level,
denoted by $p(t)$, $p(\rho)$, and $p(\lambda)$.
In addition we assume independence between these factors.
Let us focus on inferring depth $t$, the most relevant unknown for
depth cameras.
Bayes rule gives
\begin{eqnarray}
\!\!\!\!\!\!P(t \mid \vec{R}) & \!\!\propto\!\! & P(\vec{R} \mid t) \: p(t) \nonumber\\
\!\!	& \!\!=\!\! & p(t) \iint P(\vec{R} | t, \rho, \lambda)
		\: p(\rho) \: p(\lambda) \,\textrm{d}\rho \,\textrm{d}\lambda.~~\label{Bayes}
\end{eqnarray}
Equation~(\ref{Bayes}) gives the posterior distribution over the true unknown depth.
We get the posterior density up to a normalization factor which may be
extracted by integrating over every possible $t$.
The posterior density is the ideal input to higher level applications which use
probabilistic models~\cite{DBLP:journals/ijcv/IsardB98,thrun2005probabilistic}.
For other applications, it may be sufficient to summarize this posterior
distribution by a point estimate, for example the posterior mean
$\hat{t}_{\textrm{Bayes}}(\vec{R}) = \E[t\mid\vec{R}]$ or the MAP 
depth $\hat{t}_{\textrm{map}}(\vec{R}) = \argmax_t \: P(t|\vec{R})$, together
with a measure of the dispersion such as the posterior variance.

Computationally, we have to solve the
integration problem~(\ref{Bayes}) at every pixel.
Doing this at frame rate under low compute resources is currently not feasible.

A second issue with~(\ref{Bayes}) is that it requires the specification of
priors $p(t)$, $p(\rho)$, and $p(\lambda)$.
While using uniform priors on depth and reflectivity is physically plausible,
specifying the prior on ambient light level is harder.
For example, operating the camera in a dark room versus a sunlit room, would
require very different priors.
If the used prior deviates too much from the actual situation our estimates of
depth could be biased, that is, suffer from systematic errors.

\subsection{Maximum Likelihood Inference (MLE)}
Alternatively we use maximum likelihood estimation for the imaging conditions,
{
\be
(\hat{t}_{\textrm{mle}},\hat{\rho}_{\textrm{mle}},\hat{\lambda}_{\textrm{mle}}) =
	\underset{t,\rho,\lambda}{\argmax} \: P(\vec{R} \mid t, \rho, \lambda). \label{max_like}
\ee
}
Instead of considering the depth that accumulates the most probability over
all reflectivity and ambient light explanations, we consider the single combined
imaging conditions
$(\hat{t}_{\textrm{mle}},\hat{\rho}_{\textrm{mle}},\hat{\lambda}_{\textrm{mle}})$
which have the highest probability of producing the observed response
$\vec{R}$.

\subsection{Maximum Aposteriori Inference (MAP)}
This method is the most likely point estimate taking into account prior
preferences.  We obtain it similar to the MLE estimate as
{
\be
(\hat{t}_{\textrm{map}},\hat{\rho}_{\textrm{map}},\hat{\lambda}_{\textrm{map}}) =
	\underset{t,\rho,\lambda}{\argmax} \:\: p(t) \: p(\rho) \: p(\lambda) \: P(\vec{R} \mid t, \rho, \lambda).
	\label{map}
\ee
}

The optimization problems~(\ref{max_like}) and~(\ref{map}) are non-linear
because $\vec{\mu}(\cdot)$ is non-linear and because our
noise model~(\ref{generative_cov}) has a signal-dependent variance.
An iterative numerical optimization is required and a frame rate solution at
every pixel is infeasible.
We discuss further details of the inference procedures for MLE, MAP, and
Bayesian inference in the supplementary materials.

\section{A Regression Tree Approach}\label{sec:rf}
All inference methods we propose,
MLE $\hat{t}_{\textrm{mle}}$,
MAP $\hat{t}_{\textrm{map}}$, and
Bayesian inference $\hat{t}_{\textrm{Bayes}}$ produce reliable depth estimates.
However the computation of these estimates is expensive and impractical for a
realtime camera system.
To perform realtime inference we use a regression approach to approximate the
model as follows.
\begin{enumerate}
\item \textbf{Offline:}
Sample imaging conditions $(t_i,\rho_i,\lambda_i)$ from the prior and
responses $\vec{R}_i$ from the model~(\ref{eqn:model1}). Then use one
of the slow inference methods to generate labeled training data
$(\vec{R}_i, \hat{t}(\vec{R}_i))$.
\item \textbf{Offline:}
Train a regression tree/forest using the training data set, to obtain a
predictor $\hat{t}_{\textrm{RF}}$.
\item \textbf{Online:}
Given an observed response $\vec{R}$ predict the inferred depth
$\hat{t}_{\textrm{RF}}(\vec{R})$.
\end{enumerate}

\noindent \textbf{Why would this procedure be a good idea?}

\vspace{0.15cm}

-- First, $\hat{t}_{\textrm{mle}}$, $\hat{t}_{\textrm{map}}$, and
$\hat{t}_{\textrm{Bayes}}$ are smooth functions from the response space to
depth and are simple to learn.

-- Second, the regression tree $\hat{t}_{\textrm{RF}}$ has small
performance requirements in terms of memory and computation.

-- Third, it decouples the runtime from future changes to
the probabilistic model and inference procedures.

In principle it would be desirable to train directly on a large and diverse
corpus of ground truth data captured from the real world; however, capturing
ground truth depth data is challenging~\cite{scharstein2014newmiddlebury},
expensive, and ensuring the diversity in imaging conditions is difficult.
Training on our forward model instead allows us to represent a wide variety of
imaging conditions.
Likewise, while we could train directly on samples $(\vec{R}_i,t_i)$ from the
model this would incur additional variance because the noise makes $\vec{R}$
stochastic even for a fixed depth value.  By training on the estimator
$(\vec{R},\hat{t}_i)$ instead we effectively remove this variance from the
regression task.

For learning the regression tree we use the standard CART sum-of-variances
criterion in a greedy depth first manner~\cite{breiman1984cart}.
%
For the interior nodes of the trees we perform binary comparisons on
the individual responses, $R_i \leq a$.
At each leaf node $b$ we store a linear regression model,
\begin{equation}
	\hat{t}_b(\vec{R}) = \theta^T_b \cdot \left[ 1, R_1, \hdots, R_n,
		R_1^2, R_1 R_2, \hdots, R_n^2 \right]^T,
	\label{eqn:leaflinear}
\end{equation}
where we use a quadratic expansion of the responses.
We estimate the parameters $\theta_b$ of each leaf model using least squares
on all training samples that reach this
leaf~\cite{DBLP:journals/ftcgv/CriminisiSK12}.

We cannot over emphasize the practical importance of a flexible and
decoupled-from-model regression scheme, in handling unexpected
or new phenomena. An example is detailed in the supplementaries.

\textbf{Approximation Tradeoffs.}
Because of its non-parametric nature the regression tree or forest can approach the
quality of the exact inference output if given sufficient training data and expressive power.
However, the key limiting factor in our actual implementation are specific
constraints on available memory and compute. Basically the depth of the tree
and the structure of the leaf predictor, determine the memory requirements.
In Section~\ref{results} we example the accuracy vs memory tradeoffs
experimentally.

%
%
%
%
%

\subsection{Additional Regression Outputs}
In addition to the estimated depth we output several other quantities per
pixel. These outputs too are produced using trained regression trees.
%
Specifically, we produce the following additional outputs:

-- Reflectivity, $\hat{\rho}$, using $\E[\rho|\vec{R}]$ or via
(\ref{max_like}), (\ref{map}).

-- Ambient light level, $\hat{\lambda}$, using $\E[\lambda|\vec{R}]$ or
via (\ref{max_like}), (\ref{map}).

-- Depth uncertainty, as described below.

-- Fit-to-model invalidation score $\gamma$, for detection of irregular imaging
conditions, described in Section~\ref{sec:gamma}.

\subsection{Computing Depth Uncertainty}
\label{sec:depthuncertainty22}
In many applications of depth cameras to computer vision problems the
estimated depth is used as part of a larger system; in these applications it
is useful to know the uncertainty of the depth estimate.
One example would be surface
reconstruction~\cite{DBLP:conf/ismar/NewcombeIHMKDKSHF11}, where uncertainty
can be used to weight individual estimates and to average them over time.

We use the variance of the depth, in the form of a \emph{standard deviation}
$\hat{\sigma}(\vec{R})$, as a measure of uncertainty.
Depending on whether we use $\hat{t}_{\textrm{Bayes}}$ or
$\hat{t}_{\textrm{mle}}$, we compute the standard deviation as follows.

For $\hat{t}_{\textrm{Bayes}}$ we use the posterior
distribution~(\ref{Bayes}), and directly compute
$\hat{\sigma}_{\textrm{Bayes}}(\vec{R})=\sqrt{\V_{t \sim P(t|\vec{R})}[t]}$.

For $\hat{t}_{\textrm{mle}}$ in~(\ref{max_like}), we employ the approach
described in~\cite{DBLP:journals/ijprai/Haralick96}. A first order
Taylor expansion of the gradient (wrt imaging conditions) of the likelihood function in~(\ref{max_like}) is used to relate a perturbation
$\Delta\vec{R}$ in the response to the resulting perturbation of the estimator
$\hat{t}_{\textrm{mle}}(\vec{R}+\Delta\vec{R})$.
This analysis leads to the covariance matrix of the maximum likelihood
estimator and an approximation to the standard deviation,
$\hat{\sigma}_{\textrm{mle}}(\vec{R}) = \sqrt{\V[\hat{t}_{\textrm{mle}}]}$.

In Section~\ref{results} and Figure~\ref{fig:depthuncertainty} we demonstrate
the accuracy of our uncertainty estimates by comparing them with the actual
observed uncertainty.
In the context of phase-based TOF, previous
work~\cite{reynolds2011tofconfidence-randomforest} used random forests
to output depth confidence scores for measured phase
signals; their regressor was trained using laser scans.
Here we instead obtain uncertainty directly from our probabilistic model.


\section{Two-Path Model for Simple Multipath}
\label{sec:twopath}

The generative model~(\ref{vec_response_model}) we used so far assumed a single direct response from the point being imaged. In order to account for multipath, this model needs to be extended
as to describe the additional multipath light being integrated at the pixel.
We demonstrate a simple extension of the model and inference as follows.

Consider a two-path model as proposed
in~\cite{dorrington2011twopath,godbaz2012twopath,DBLP:conf/eccv/FreedmanSKLS14}.
In addition to the three unknowns $t$, $\rho$, and $\lambda$, we now also
assume a second contribution having travelled depth $t_2 > t$, from a patch with reflectivity
$\rho_2$.
We extend the generative model~(\ref{generative_mean}) to
\begin{equation}
\vec{\mu}_2(t,\rho,\lambda,t_2,\rho_2)
	= \rho \left(\vec{C}(t) + \lambda\vec{A}
		+ \rho_2 \vec{C}(t_2)\right),\label{generative_mean2}
\end{equation}
where $\rho$ scales both the direct and indirect response, and $\rho_2$ scales
only the indirect response.
The model is exact for a second specular surface, but becomes an approximation
in case the second surface is diffuse.
For inference we extend the inference procedures to
this model in a straightforward manner (details in the supplementaries).

\begin{figure}[t!] \centering
	\subfigure{%
		\includegraphics[width=0.20\linewidth]{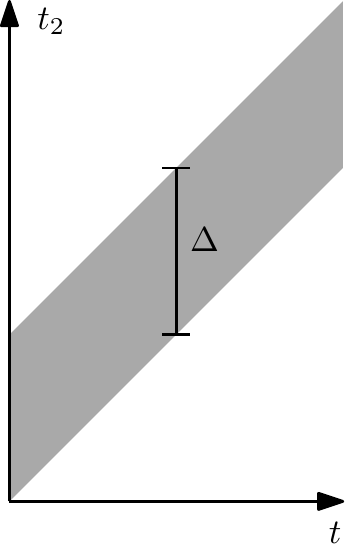}%
		\label{fig:tp-prior-t2}%
	}\hfill%
	\subfigure{%
		\includegraphics[width=0.76\linewidth]{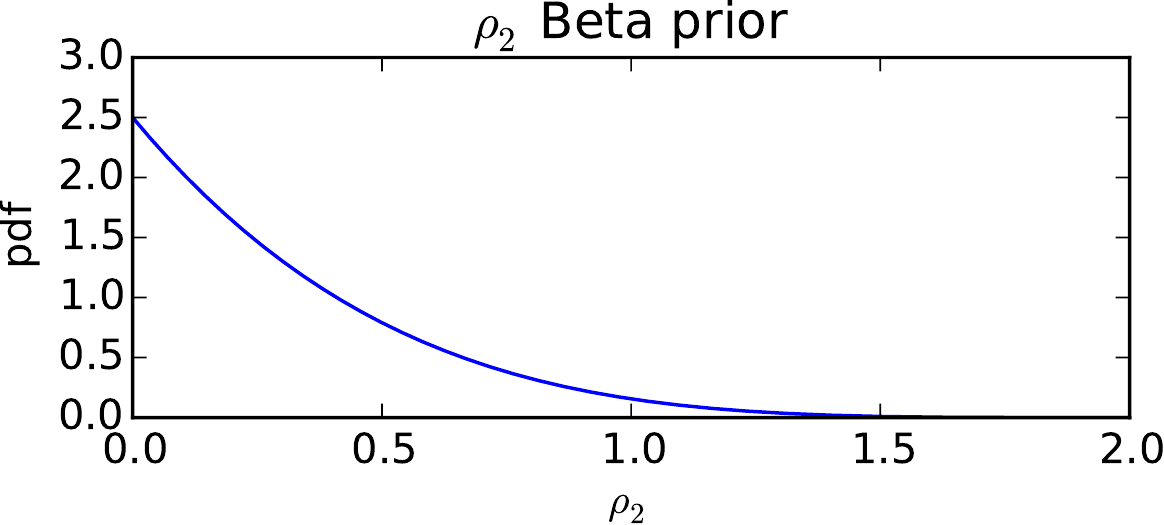}%
		\label{fig:tp-prior-rho2}%
	}%
\caption{\footnotesize Two-path model priors for the additional latent
variables $t_2$ and $\rho_2$.  Left: The prior for the second bounce depth $t_2$ is
uniform over the shaded polygon.  Right: the prior for $\rho_2$ is defined
over $[0;2]$ in order to handle large diffuse reflectors.}
\label{fig:tp-priors}
\end{figure}

For the prior of $t_2$ we select a uniform prior relative to $t$, such that
$t_2-t$ is uniform between $0cm$ and $\Delta$ (typically $\Delta=150cm$), that is,
\begin{equation}
	p(t_2|t) = \mathcal{U}(t_2; t, t+\Delta).
	\label{eqn:prior-t2}
\end{equation}
For the second reflectivity we allow $\rho_2 > 1$ to be able to handle diffuse
surfaces, and after studies of simulation data select a Beta
distribution on the interval $[0;2]$.
\begin{equation}
	p(\rho_2) = \mathcal{B}(\rho_2 / 2; \alpha=1,\beta=5).
	\label{eqn:prior-rho2}
\end{equation}
This prior specifies that values up to $\rho_2=2$ are possible, but that low
values of $\rho_2$ are more likely.
Both priors are visualized in Figure~\ref{fig:tp-priors} and we will evaluate
this model on real and simulated data in the experiments section.
%

It is important to emphasize that our regression-decoupled-from-model approach allows
us to seamlessly use this extended model in the camera, just by plugging it in the offline
step 1 of the procedure outlined at the beginning of Section \ref{sec:rf}. The runtime process
and its computational cost do not change at all.

\section{Bayesian Model Invalidation Score $\gamma$}\label{sec:gamma}
Our imaging model is an idealization of the real world and in each frame a certain
number of pixels will have measured responses $\vec{R}$ which do not conform
to this model.
The main reasons for this are systematic errors such as
multipath~\cite{fuchs2010multipath,DBLP:conf/eccv/FreedmanSKLS14},
pixels of mixed depth, sensor saturation, as well as statistical extremes in
imaging noise.  In Section~\ref{sec:twopath} we extended our model to explain
some multipath effects, but even this extended model may fail to explain
some of the responses.

When our model fails to accurately explain the observed response
vector $\vec{R}$ we would like to detect such a deviation from the model
assumptions and exclude affected observations from further processing.
The strict Bayesian paradigm cannot detect deviation from model assumptions
because it only provides the calculus to go from assumptions and observations
to conclusions and no mechanism to falsify the assumptions
themselves~\cite{gelman2013philosophybayes}.
However, in Bayesian modelling practice~\cite{lunn2012bugs} a common method to
assess deviations from model assumptions is to perform so called \emph{posterior
predictive checks}.

We use the posterior predictive
p-value~\cite{meng1994posteriorpredictivepvalue,bayarri2000compositenull,robins2000asymptoticpvalue}
for our purposes.
Intuitively our particular p-value will measure the total probability mass
of all observations which have a smaller likelihood than the likelihood of the
observed response.  Therefore the score is always between zero and one and a
value close to zero indicates that the observation is unlikely under the
assumed model.
This intuition is helpful but the controversy around p-values and model
checking more generally is deep and we give a brief discussion in the
supplementary materials.

To formalize this problem, let us first unify notation by writing
$\theta=(t,\rho,\lambda)$ or $\theta=(t,\rho,\lambda,t_2,\rho_2)$, depending
on whether we use the single path model~(\ref{generative_mean}) or the two path
model~(\ref{generative_mean2}), so that $\theta$ are all the unknown
imaging conditions to be inferred.
Given an observed response vector $\vec{R}$ and using the model
$P(\vec{R}|\theta)$ and the prior $P(\theta)$ we can use Bayesian inference to
infer the posterior distribution $P(\theta|\vec{R})$.
Following the above intuition the invalidation score $\gamma$ is defined as
\begin{equation}
	\gamma(\vec{R}) = \E_{\theta \sim P(\theta|\vec{R})} \left[
			\E_{\vec{R}' \sim P(\vec{R}'|\theta)} \left[
				1_{\{P(\vec{R}'|\theta) \leq P(\vec{R}|\theta)\}}
			\right]
		\right],\label{eqn:gamma-exp}
\end{equation}
Here we used the notation $1_{\{\textrm{predicate}\}}$ which evaluates to one
if the predicate is true and to zero otherwise.
The above equation integrates all probability mass of less likely
observations, weighted by the posterior $P(\theta|\vec{R})$.
If we have many repetitions of the experiment $\theta \sim P(\theta)$ and
$\vec{R} \sim P(\vec{R}|\theta)$ the scores $\gamma(\vec{R})$ would be
uniformly distributed.
The computation of~(\ref{eqn:gamma-exp}) is essentially free during
our approximate Bayesian inference procedure.

The value $\gamma(\vec{R})$ can be used to reject the null hypothesis of the
assumed model: if $\gamma(\vec{R}) \leq \tau$ for some threshold $\tau$ we
reject the assumed model for this observation.
The score~(\ref{eqn:gamma-exp}) is also applicable to MLE and MAP inference if
we replace the outer expectation by a unit point mass at the inferred imaging
conditions $\hat{\theta}_{\textrm{mle}}$ or $\hat{\theta}_{\textrm{map}}$,
respectively.
We evaluate the invalidation score experimentally in
Section~\ref{sec:exp:invalidation}.

\begin{figure}[t!] \centering
	\centering%
	\subfigure{\includegraphics[width=0.9\linewidth]{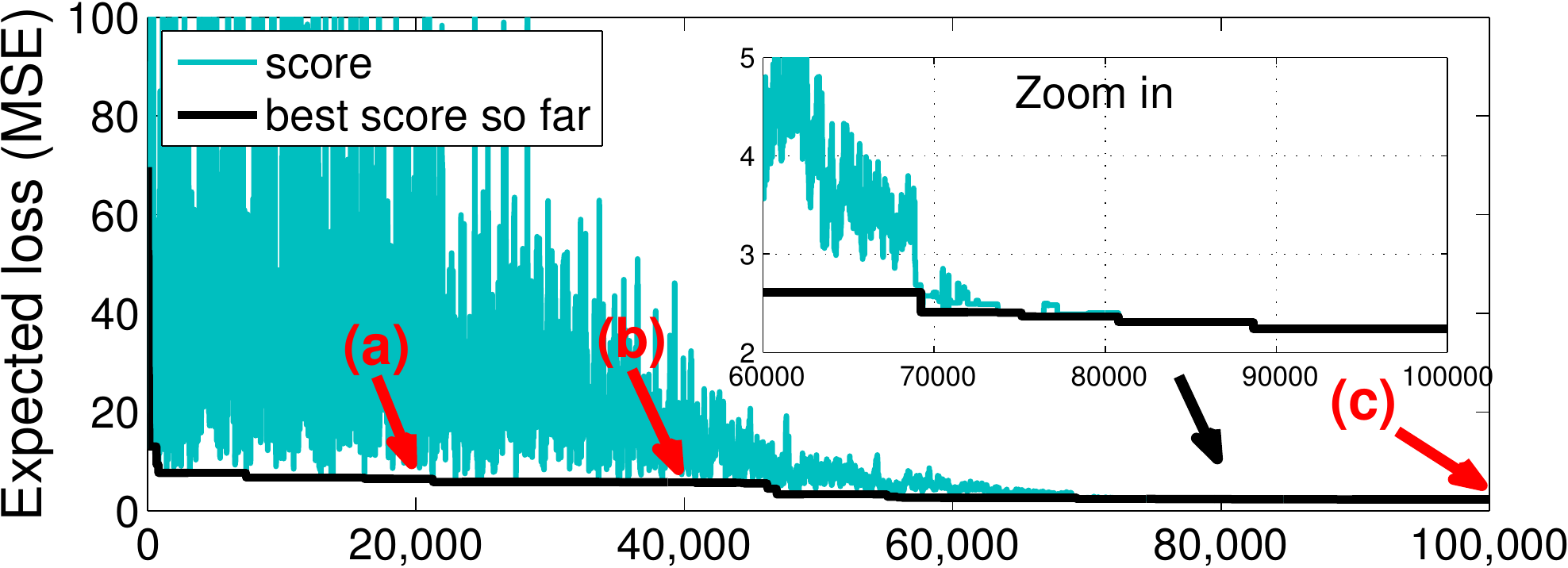}}\\%
	\subfigure[20,000]{
		\includegraphics[width=0.3\linewidth]{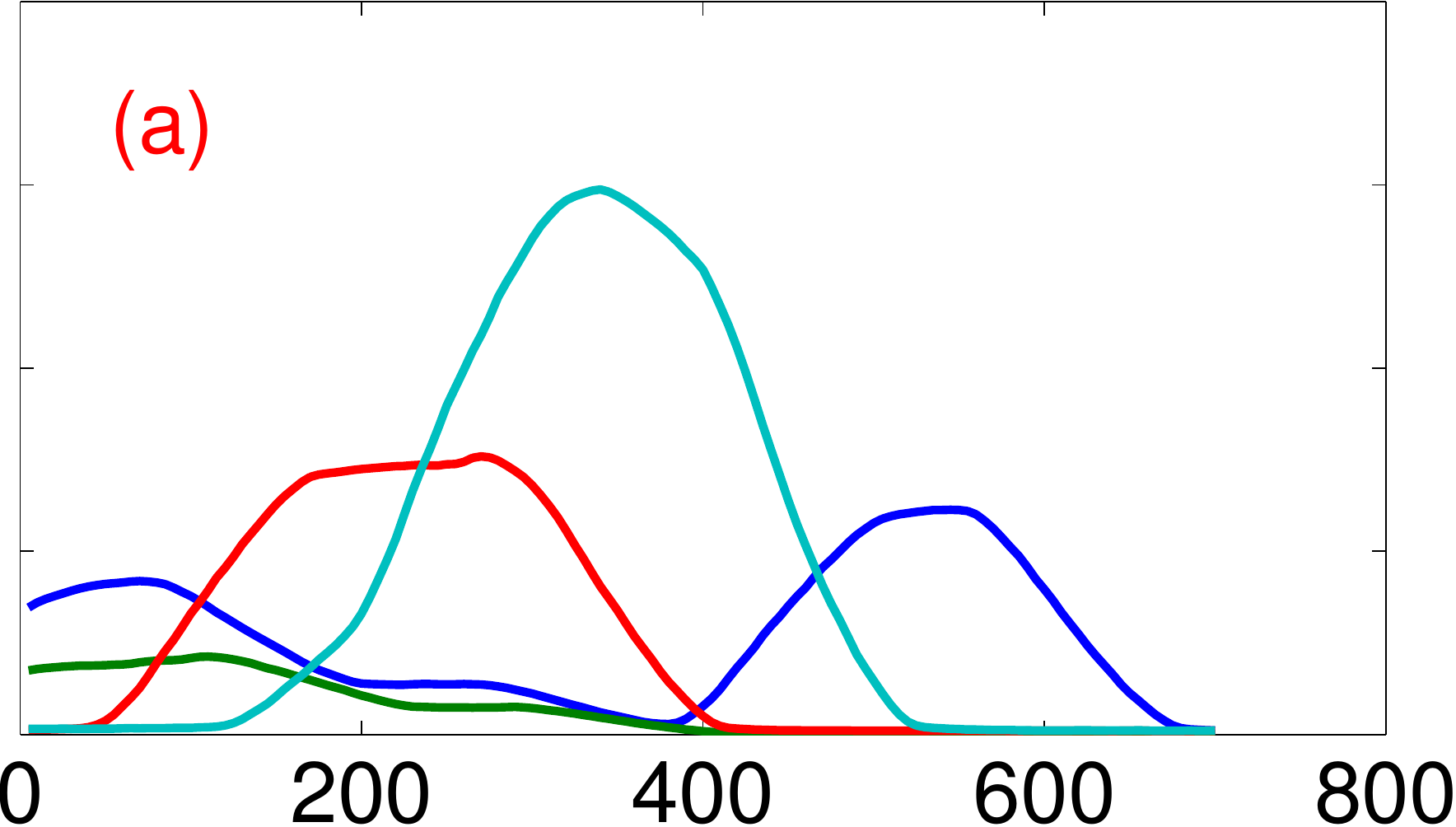}%
		\label{fig:annealing-20k}%
	}\hfill%
	\subfigure[40,000]{
		\includegraphics[width=0.3\linewidth]{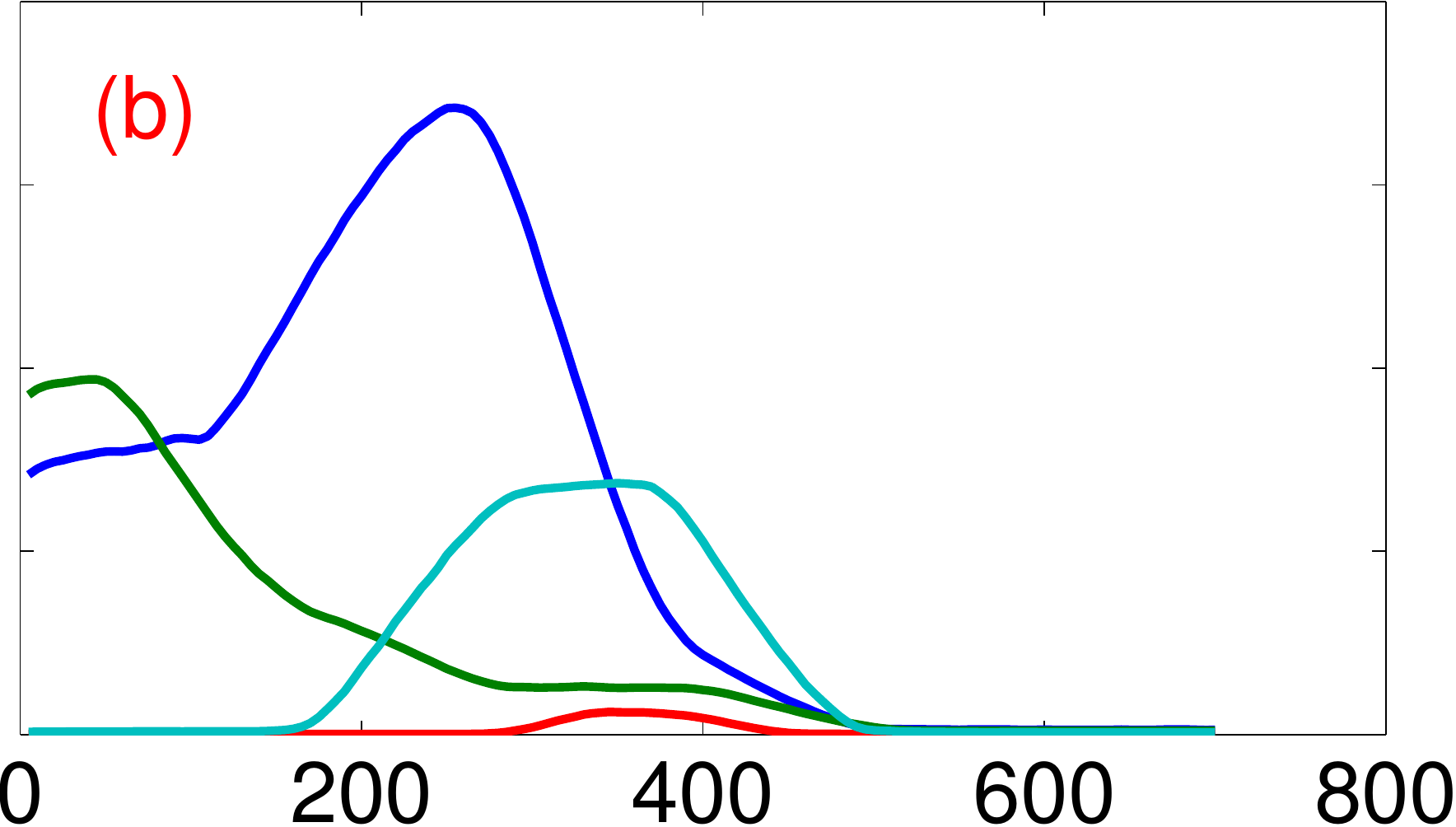}%
		\label{fig:annealing-40k}%
	}\hfill%
	\subfigure[100,000]{
		\includegraphics[width=0.3\linewidth]{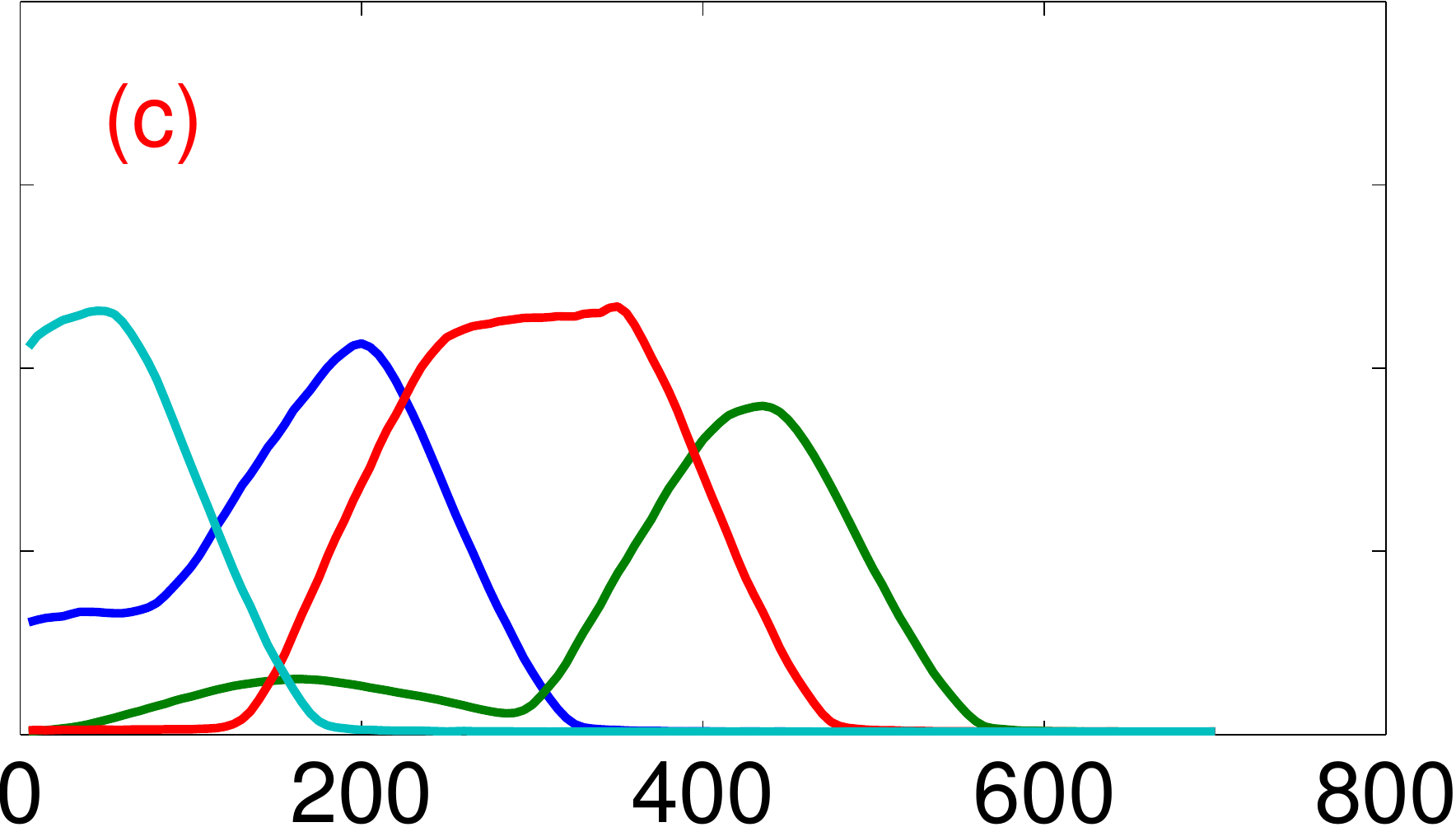}%
		\label{fig:annealing-100k}%
	}
\caption{\footnotesize Exposure profile optimization.
\textbf{Top}:
Simulated annealing over 100k iterations, finding response curves to
minimize~(\ref{prob:wpdesign}), the expected error (MSE) in depth
estimation.
\textbf{Bottom}:
snapshots of the response curves after 20k, 40k, and after all 100k
iterations. The x-axis is depth (cm).}
\label{fig:annealing}%
\end{figure}


\section{Exposure Profiles Design}\label{wp_design}
%
\setlength{\columnsep}{0.5cm}%
\begin{wrapfigure}[8]{r}{0.45\linewidth}
	\vspace{-0.25cm}%
	\begin{center}%
	\includegraphics[width=0.7\linewidth]{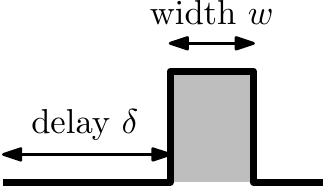}%
	\end{center}%
	\caption{\footnotesize A basis element $j=(\delta,w)$ defining $B_j(\cdot)$.}%
	\label{basis_element}%
\end{wrapfigure}%

So far we covered our imaging model, our realtime regression approach and how
we can invalidate responses unlikely to have been generated by our model.
We now turn to an orthogonal question of designing a suitable response curve $\vec{C}$ for
use in~(\ref{vec_response_model}) ($\vec{A}$ is closely related to $\vec{C}$ and
are both derived from $\Z$ which will immediately be defined).
Recall from~(\ref{c_of_t}) that $\vec{C}$ is the integral of the illumination
pulse $P$ with the exposure profile $S$.
In the camera, a laser diode and driver produce the illumination pulse
$P$, and its design is fixed.
The exposure gain $S$, however, has a flexible design space
parameterized by linear basis functions.
We would like to design response curves $\vec{C}$ that will produce observations from
which low-error estimates of the imaging conditions could be inferred.

%
\begin{figure}[t]\centering%
	\subfigure[]{%
		\includegraphics[width=0.51\linewidth]{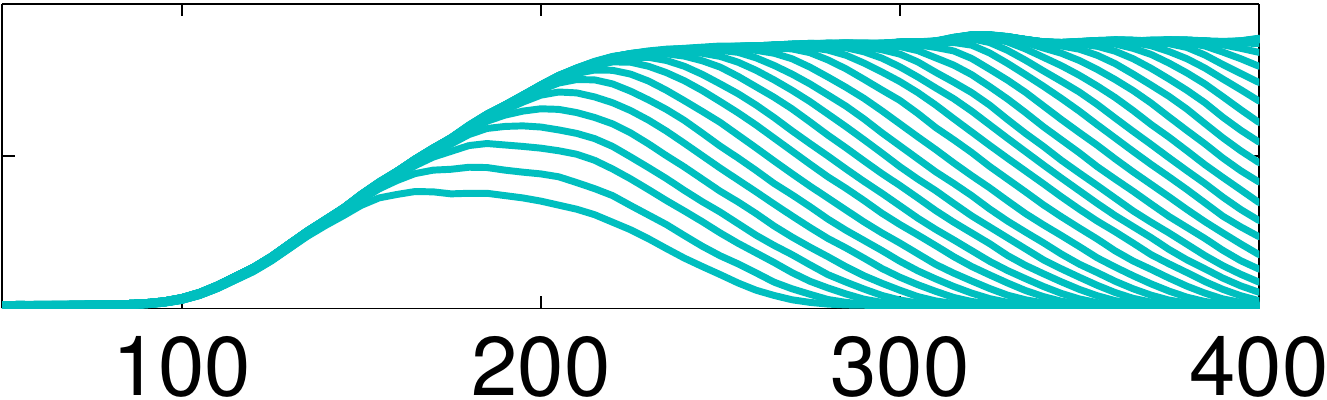}%
		\label{fig:wp-basis1}%
	}\quad%
	\subfigure[]{%
		\includegraphics[width=0.44\linewidth]{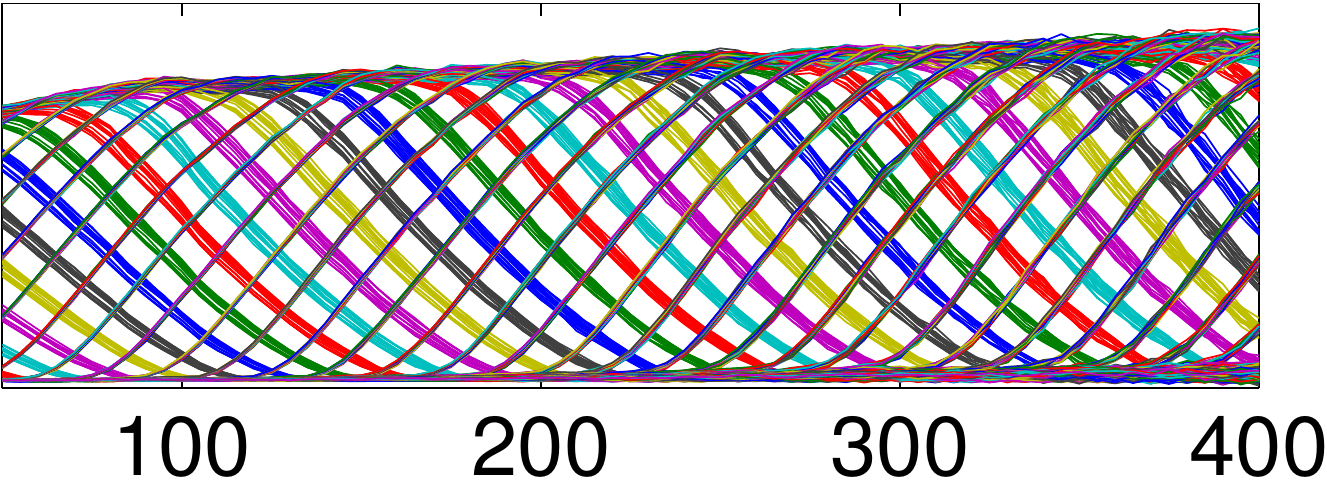}%
		\label{fig:wp-basis-all}%
	}%
\vspace{-0.3cm}%
\caption{\footnotesize %
\textbf{Left}, \subref{fig:wp-basis1}: basis functions $\{Q_j/d\}$ for a fixed
delay $\delta$ and varying width $w$.
\textbf{Right}, \subref{fig:wp-basis-all}:
all basis functions $\{Q_j/d\}_{j \in J}$, defined by Eq.~(\ref{profiling_basis}).}
\label{profiling_plot}%
\end{figure}

The camera is able to use basic exposure gain profiles in the form of a boxcar
function, as shown in Fig.~\ref{basis_element}.
Each basic gain profile has two parameters: a delay $\delta$, and a width $w$.
Each possible pair $j=(\delta,w)$ specifies one possible gain profile $B_j$ from a fixed
discrete set of choices $J$.  Typically the set $J$ contains several hundred
possible combinations.
%
%
\begin{figure*}[t!] \centering
	\subfigure[Scene]{%
		\includegraphics[width=0.145\linewidth]{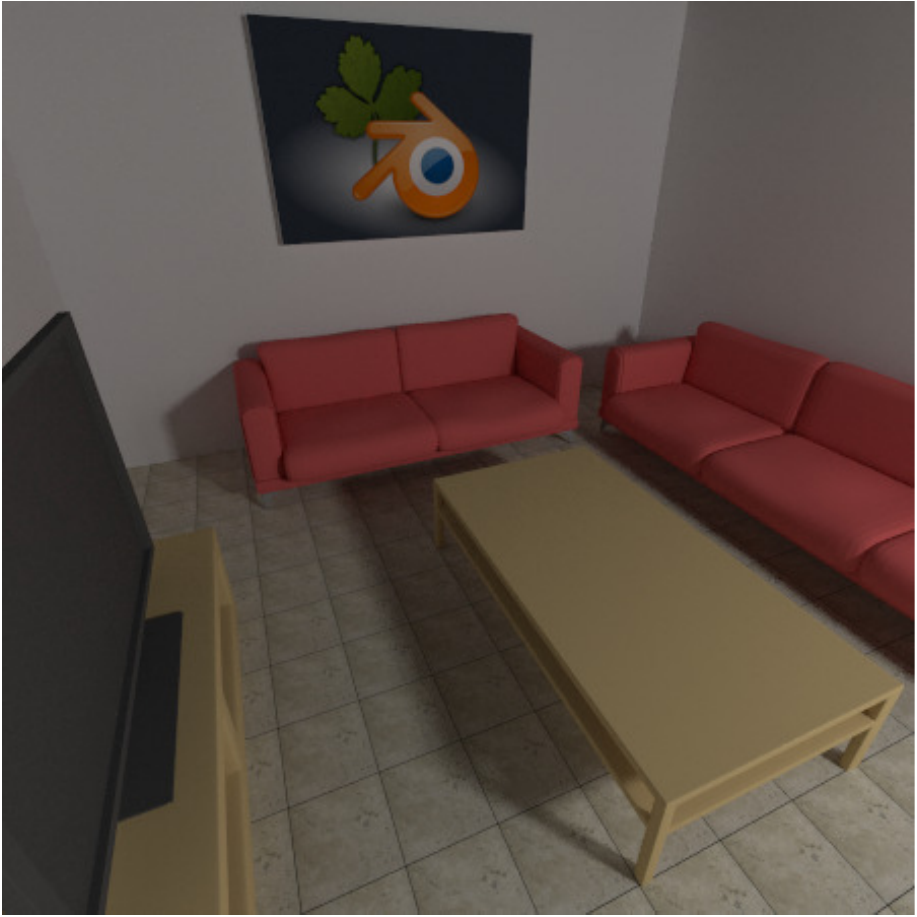}%
		\label{fig:maslulim-visible}%
	}\hfill%
	\subfigure[Ground truth depth]{%
		\includegraphics[width=0.19\linewidth]{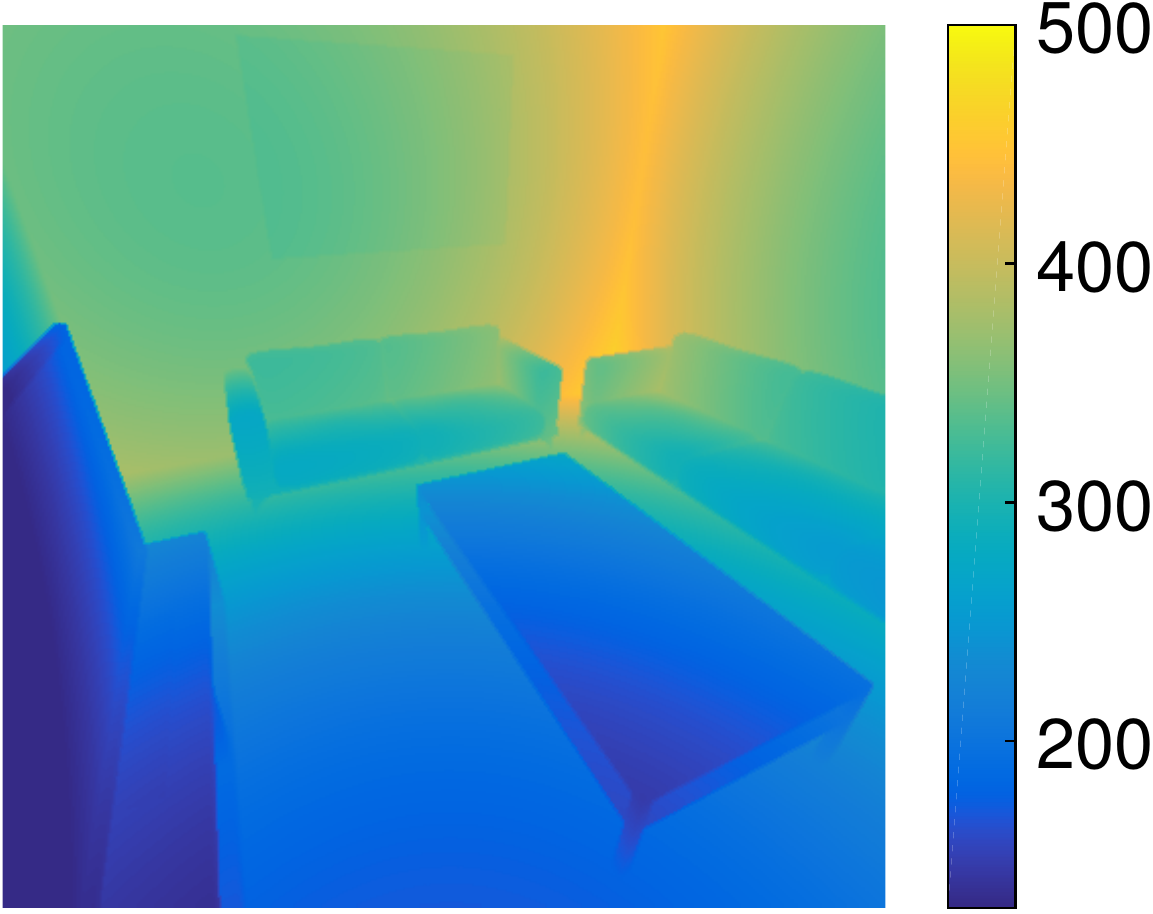}%
		\label{fig:maslulim-depth}%
	}\hfill%
	\subfigure[Multipath ratio]{%
		\includegraphics[width=0.18\linewidth]{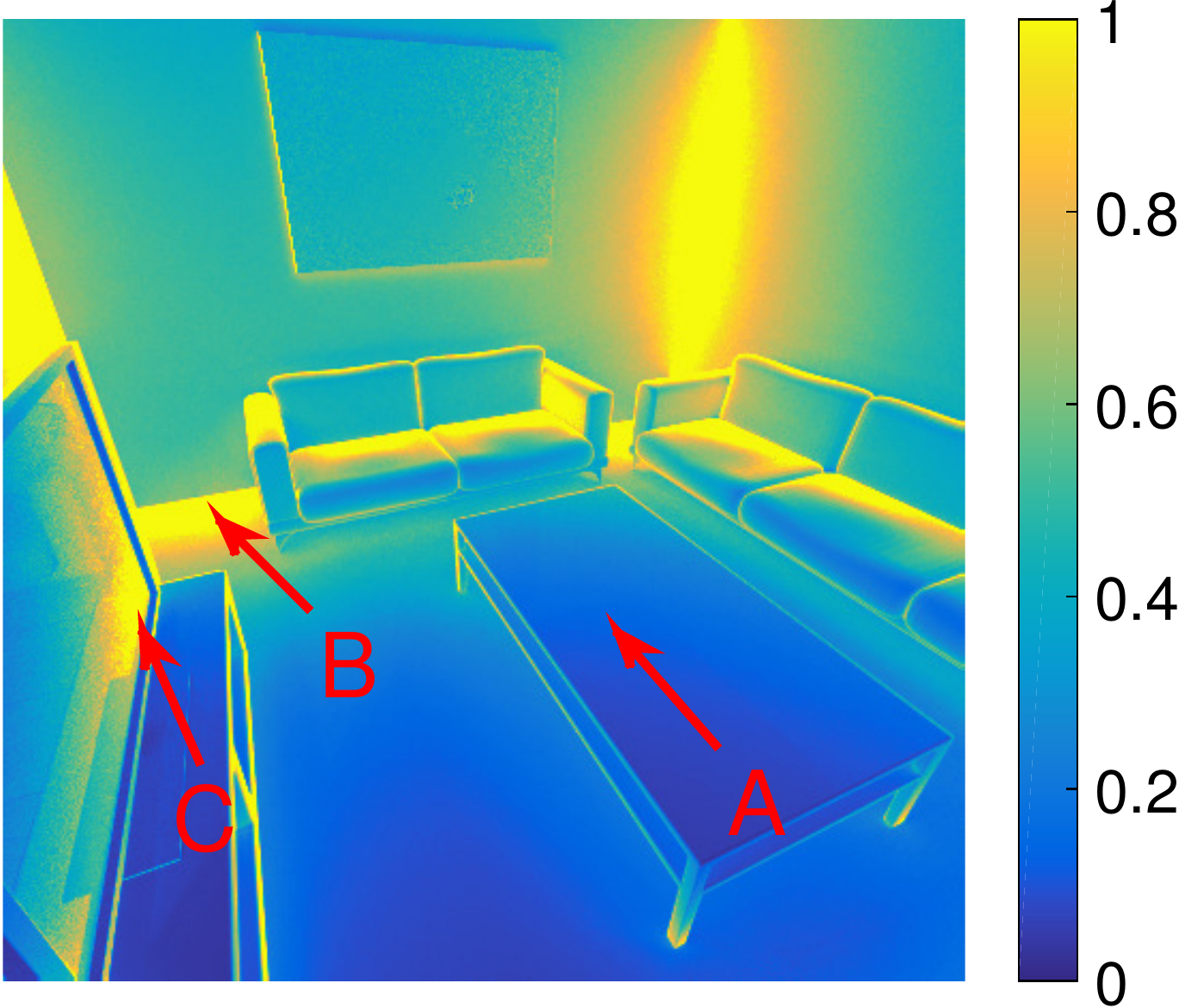}%
		\label{fig:maslulim-multipathratio}%
	}\hfill%
	\subfigure[Pixel A]{%
		\includegraphics[width=0.145\linewidth]{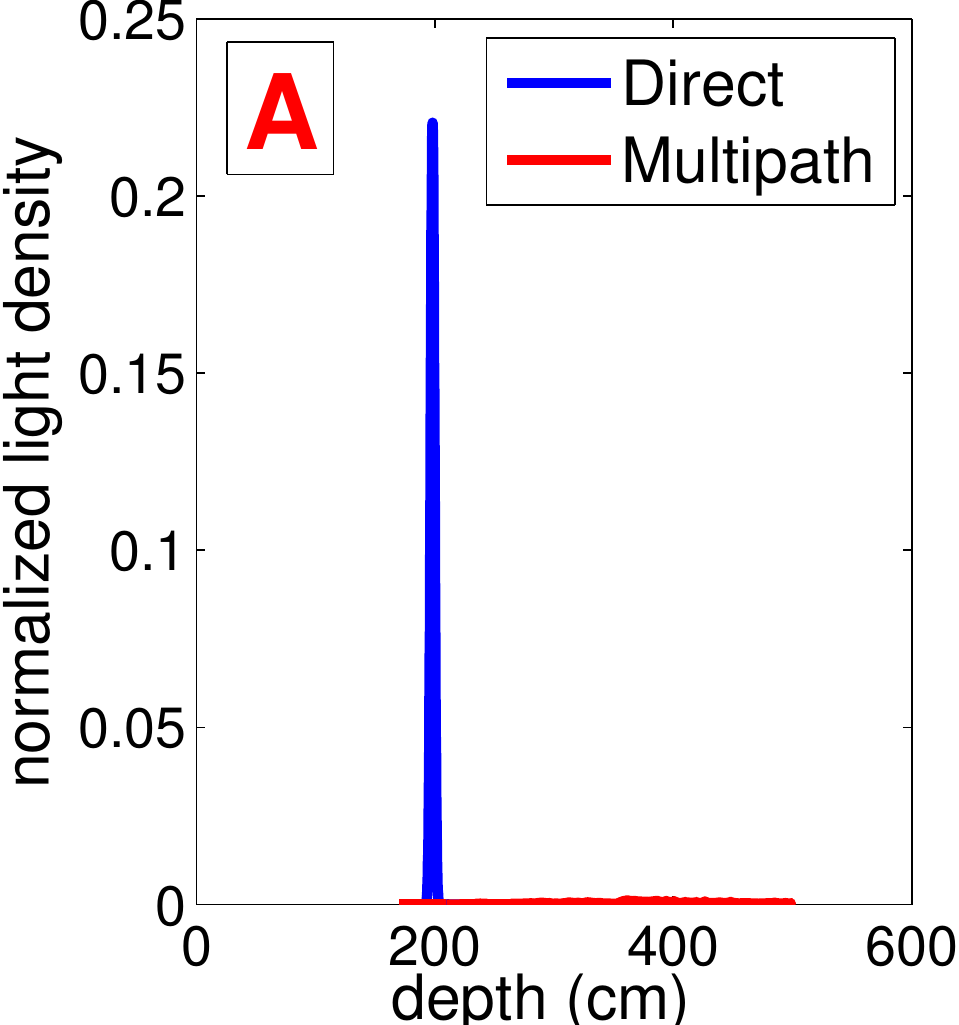}%
		\label{fig:maslulim-pixel-A}%
	}\hfill%
	\subfigure[Pixel B]{%
		\includegraphics[width=0.145\linewidth]{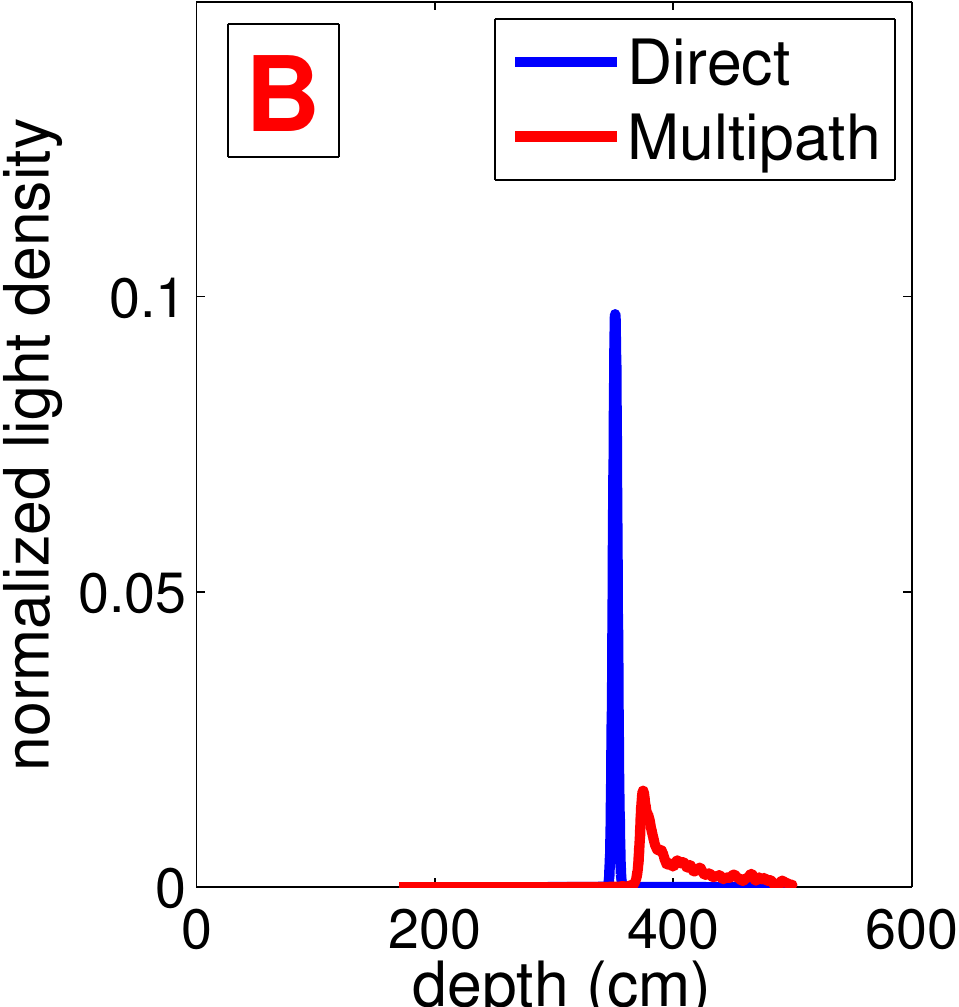}%
		\label{fig:maslulim-pixel-B}%
	}\hfill%
	\subfigure[Pixel C]{%
		\includegraphics[width=0.145\linewidth]{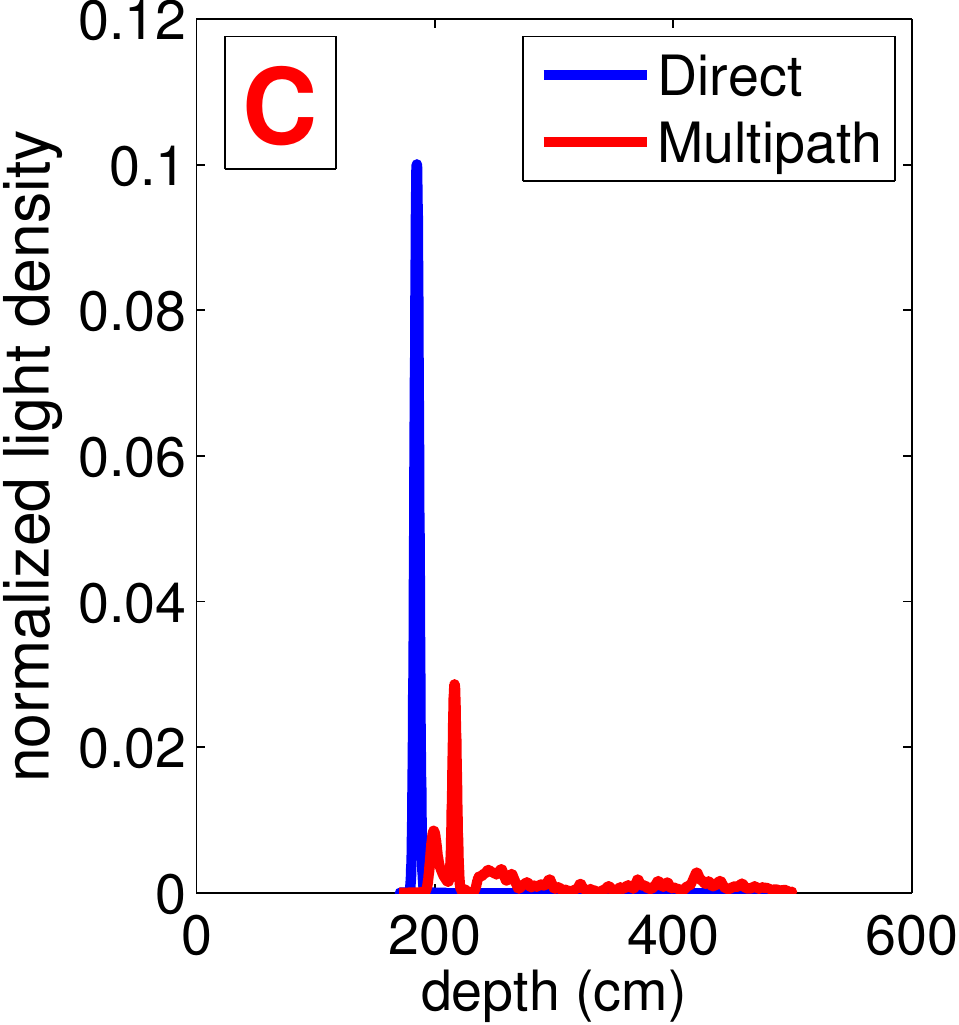}%
		\label{fig:maslulim-pixel-C}%
	}%
\vspace{-0.3cm}%
\caption{\footnotesize Insights into multipath using physically accurate light transport simulation.
\subref{fig:maslulim-visible}
A scene created in Blender.
\subref{fig:maslulim-depth}
Ground truth depth.
\subref{fig:maslulim-multipathratio}
Normalized measure of multipath intensity compared to direct contributions.
(please see the main text).
\subref{fig:maslulim-pixel-A}-\subref{fig:maslulim-pixel-C}
Normalized light densities for three selected pixels;
pixel A has no multipath component,
pixel B has one multipath component from 30--50cm further away, and
pixel C, where a specular component gives rise to a narrow multipath response.
}
\label{maslulim_fig}%
\end{figure*}
%
%
With~(\ref{c_of_t}) we now get the basis function $Q_j$ associated to $B_j$ as
convolution with the pulse,
{\small
\be
	Q_j(t) = \int B_j(u) \: P(u-t) \: d(t) \,\textrm{d}u. \label{profiling_basis}
\ee
}
Fig.~\ref{profiling_plot} shows a set of basis functions for all possible $j \in J$.
We represent the basis response curves as vectors $Q_j \in \R^T$, for a time
discretization with $T$ values.  By stacking the $m=|J|$ vectors horizontally
we obtain a matrix $\Q \in \R^{T\times m}$ containing all possible basis
response curves.
%
The possible design space represents each curve as linear
combination of basis curves, that is $S(\cdot) = \sum z_j B_j(\cdot)$.
With~(\ref{profiling_basis}) we then obtain the combined response curve
$C(\cdot) = \sum z_j Q_j(\cdot)$.
To design not just one but $n$ response curves $S_i(\cdot)$ for $i=1,\ldots,n$,
we represent the design space using a matrix
$\Z \in \R^{m\times n}$ as
{\small
\be
	\C = \Q \: \Z,
\ee
}
where in $\C \in \R^{T \times n}$ the $k$'th column contains the response for
the $k$'th exposure sequence.

For the design objective we utilize \emph{statistical decision
theory}~\cite{berger1985statisticaldecisiontheory}
to select $\Z$ to optimize the expected quality of depth inference.
There are two components to this idea:
the quality measure, and the expectation.
The quality of depth inference is measured by means of a \emph{loss function}
which compares an estimated depth $\hat{t}$ with a known ground truth depth
$t$ to yield a quality score $\ell(\hat{t},t)$.
One possible loss function which we use is the squared error, $\ell(\hat{t},t) = (\hat{t}-t)^2$,
but we can also use other functions, for example
$\ell_t(\hat{t},t)=\ell(\hat{t},t) / t$.
For the expectation, as for the Bayesian depth inference, we devise priors,
typically uniform, $p(t)$, $p(\rho)$, and $p(\lambda)$ over the unknowns.
Then the design problem is
{\small
\begin{eqnarray}
\!\!\!\!\!\!\!\!& \underset{\Z}{\argmin} &
	\E_{t,\rho,\lambda} \: \E_{\vec{R} \sim P(\vec{R} | t,\rho,\lambda,\Z)}
		\left[\ell(\hat{t}(\vec{R}), t)\right]\label{prob:wpdesign}\\
\!\!\!\!& \textrm{sb.t.} &
	\sum_{j=1}^m \sum_{i=1}^n Z_{ji} \leq K_{\textrm{shutter}},\label{con:shuttertotal}\\
\!\!\!\!	& & \sum_{j=1}^m 1_{\{Z_{ji} > 0\}} \leq K_{\textrm{sparsity}},\quad i=1,\dots,n,\quad\label{con:sparsity}\\
\!\!\!\!	& & Z_{ji} \in \N, \quad j=1,\dots,m, \: i=1,\dots,n,\nonumber
\end{eqnarray}
}
where the notation $1_{\{\textrm{pred}\}}$ evaluates to one if the predicate
is true and to zero otherwise.

The constraints~(\ref{con:shuttertotal}) and~(\ref{con:sparsity}) deserve some
comments.
Each captured frame contains a fixed number $K_{\textrm{shutter}}$ of light
pulses, each of which is associated with a basic exposure signal $B_j$.
These are assigned in integer units.
The total number of basis functions that can be used is constrained by
$K_{\textrm{sparsity}}$ due to various shutter driver restrictions.
Because in each pulse a single basis function is selected, this makes the
effective response curve $\C$ a non-negative linear combination of the basis
functions.

Solving~(\ref{prob:wpdesign}) is a challenging combinatorial problem on three levels:
first, computing $\hat{t}(\vec{R})$ is the inference problem, which has no closed form solution.
Second, as a result, computing the expectations also has no closed form solution.
Third, more than just merely evaluating it, we would like to optimize the
objective function over $\Z$.

The approximate solution which we adopt is as follows (more details in the
supplementary materials).
We approximate the objective function by a Monte Carlo evaluation for both
expectations (imaging conditions, and responses):
for $i=1,\dots,K$ we draw $t_i$, $\rho_i$, $\lambda_i$, then draw $\vec{R}_i$, then
perform inference to obtain $\hat{t}_i=\hat{t}(\vec{R}_i)$ and evaluate
$\ell_i = \ell(\hat{t}_i,t_i)$.
Finally we approximate the objective~(\ref{prob:wpdesign}) as empirical mean
$\frac{1}{K} \sum_{i=1}^K \ell_i$.  For $K=8192$ samples this computation
takes around one second.
For optimization of~(\ref{prob:wpdesign}) we use simulated
annealing~\cite{kirkpatrick1983simulatedannealing} on a
custom-designed Markov chain which respects the structure induced
by~(\ref{con:shuttertotal}) and~(\ref{con:sparsity}).

Figure~\ref{fig:annealing} shows the progress of the optimization process.
We start the optimization at a completely closed exposure profile with zero
gain, that is $Z_{ji} = 0$ for all $j$, $i$.

We remark that the optimization scheme just described
outperforms all our previous attempts to manually design the exposure
profiles.


\section{Multipath Modeling and Design}
\label{mp_simulator}
In this section we describe our method for simulating realistic
multipath images together with ground truth. Having a realistic simulation enables
several important goals:

-- exposure design for reduced multipath artifacts

-- learning/obtaining realistic priors for multipath effects

-- benchmarking

We show results for the first and last goal in section \ref{results}.

\subsection{Time of Flight Simulation}
In computer graphics physically-accurate renderers are mature technology that
are readily available.
We adapt the open source \emph{Mitsuba renderer}~\cite{wenzel2010mitsuba}.
Mitsuba supports, based on physical modeling of light scattering, light
transport simulation, integrating paths of light at every pixel, thereby
producing a highly realistic rendered image.
We adapt the code so that we obtain the total light path length and the number
of segments of the light trajectory.
%

In more detail,
we modify two rendering algorithms, the bidirectional path tracer
algorithm~\cite{veach1995bidirectional} and the
Metropolis light transport (MLT)~\cite{veach1997mlt} algorithm;
normally both algorithms are used to
render the intensity of a pixel by means of approximating an integral over
light pathes connecting light sources to surfaces to camera
pixels~\cite{pharr2010physicallybasedrendering}.

Our modification is to record for each pixel a weighted set of light path
samples $\{(w_i,L_i,t_i)\}_{i=1,\dots,N}$, typically a few thousand, say
$N=4096$.
For each light path we store the intensity weight $w_i \geq 0$, the number of
straight path segments $L_i \in \N$, and the total length of the path $t_i$.
The segment count allows us to distinguish direct responses ($L_i=2$,
emitter-to-surface and surface-to-camera) from indirect responses
($L_i > 2$, multipath).
Together with a fixed ambient value $\tau$ measuring light intensity without
active illumination, for example from a regular rendering pass, the path
lengths and weights now permit us to simulate a realistic mean response vector
$\vec{\mu}$ as
{\small
\begin{equation}
	\vec{\mu} = \tau \vec{A} + \sum_{i=1}^N \frac{w_i}{d(t_i)} \vec{C}(t_i).
	\label{eqn:simforward}
\end{equation}
}
The sum in the second term approximates the time-of-flight integral
$\int_{\R_+} \vec{C}(t) \,\textrm{d}\nu(t)$, where $\nu$
is an intensity measure over time.
The division by $d(t_i)$ is due to both $w_i$ and $\vec{C}$ containing the
distance decay function $d(t)$; see~(\ref{c_of_t}).
Once we have $\vec{\mu}$ we can optionally simulate sensor noise as specified
in~(\ref{generative_cov}).
We provide details in the supplementaries.

We remark that additional relevant work
on light transport is considered in \cite{Pitts2014,JaraboSIGA14},
published independently and concurrently with our work.

\subsection{Simulation Results}
In part (a) of Figure \ref{maslulim_fig} we show a synthetic scene. Part (b) shows the ground truth depth map corresponding to the scene.
We marked three points (A, B and C shown in part \ref{fig:maslulim-multipathratio})
at which we have different amounts of multipath.
In parts \ref{fig:maslulim-pixel-A}, \ref{fig:maslulim-pixel-B}, and~\ref{fig:maslulim-pixel-C}
we show the depth histograms we obtain from our modified Mitsuba renderer.
For every point, the histogram shows the distribution of distances
travelled by the photons integrated at this pixel.
This distribution is properly weighted to account for both distances and
reflectivity of materials along the pathes. Furthermore we
show the distribution of distances travelled over a direct path in blue (this essentially corresponds to a delta function), and distances travelled over multiple pathes in red. We see that at point A (part \ref{fig:maslulim-pixel-A})  there is no multipath, while at point B (part \ref{fig:maslulim-pixel-B}) there is multipath due to the wall. We may see from the histogram the dominant additional path lengths - 30 to 50 cm in this case. Finally, in part \ref{fig:maslulim-multipathratio} we show a normalized measure of the percentage of intensity integrated from multipath (as opposed to intensity integrated from a single direct light path), for every pixel in the image. We see that corners and just in front of the wall or other vertical surfaces actually return more multipath signals than direct path signals.

\begin{figure}[t!]%
\centering%
	\subfigure[$R_1$ response and 500 random pixels.]{%
		\includegraphics[width=0.47\linewidth]{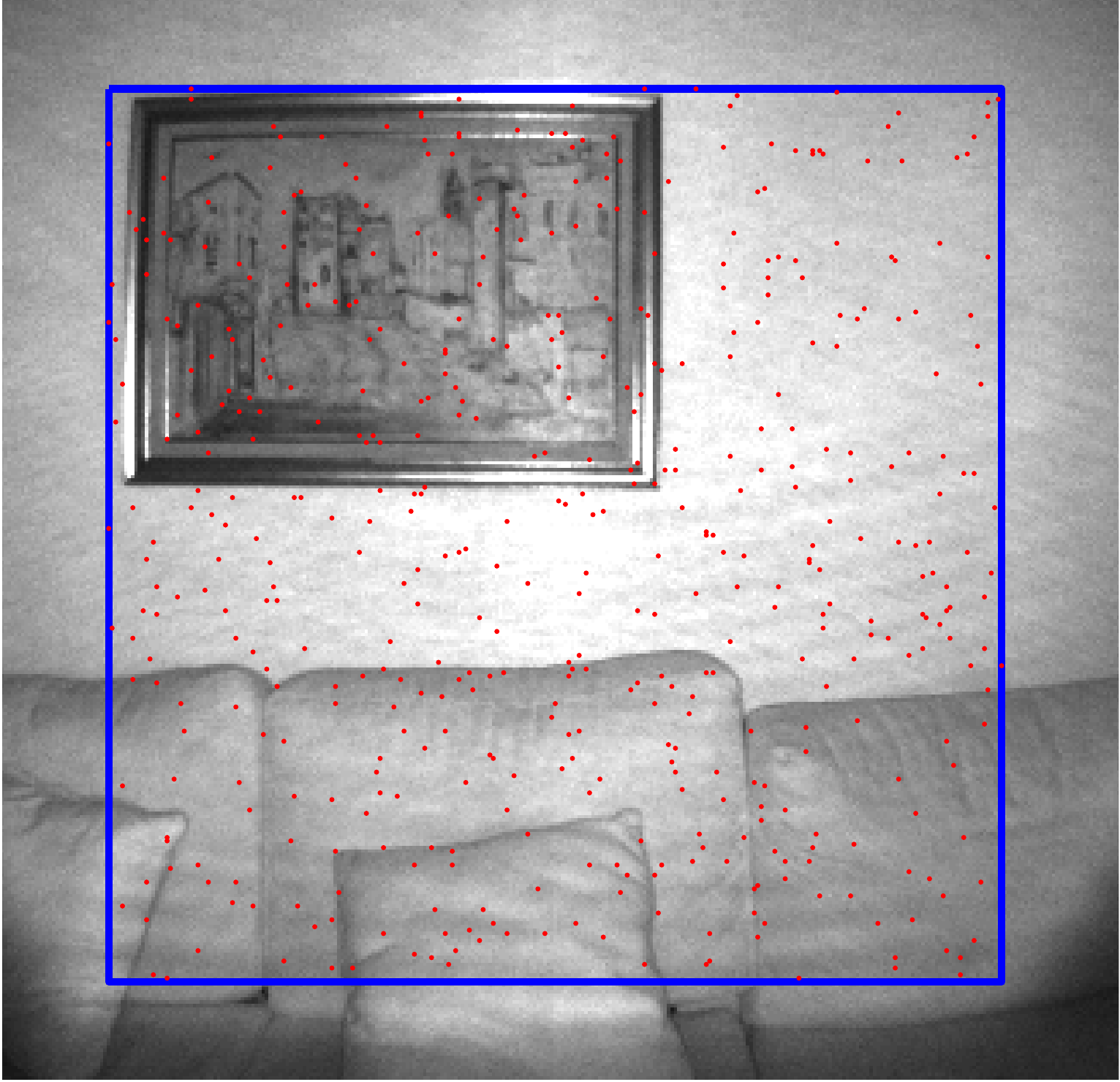}%
		\label{fig:sceneresponse}%
	}\hfill%
	\subfigure[Predicted vs actual depth uncertainty.]{%
		\includegraphics[width=0.45\linewidth]{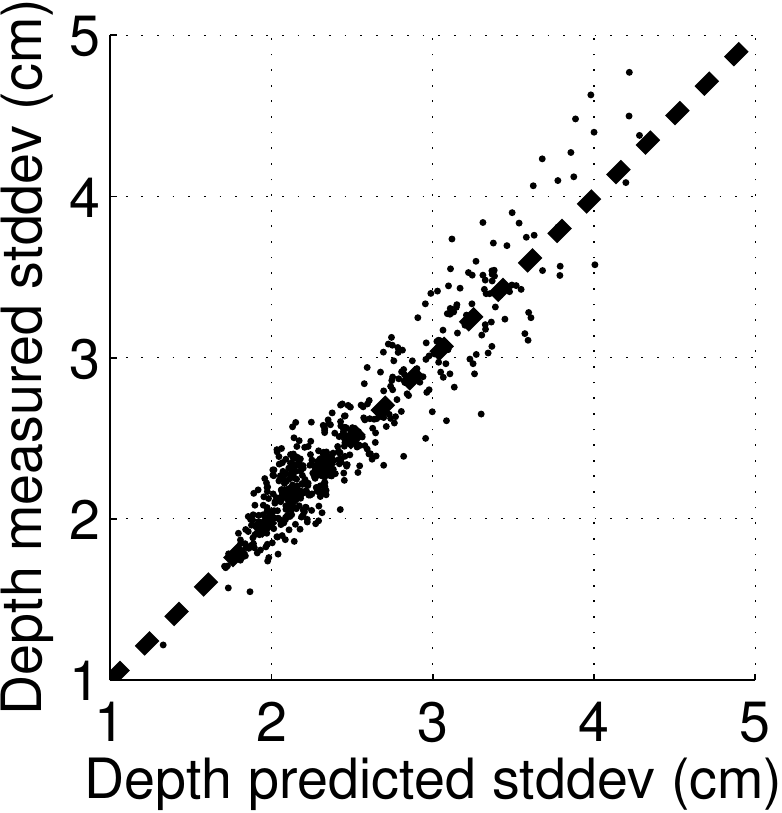}%
		\label{fig:depthuncertainty}%
	}%
	\caption{Predicted uncertainty versus actual uncertainty; the model is
well-calibrated in that it accurately predicts depth uncertainty.}
\end{figure}

%
%
%
\begin{figure}[t!]
%
	\subfigure{\includegraphics[width=0.33\linewidth]{figures-c/workpoints/far_wp_dc2.pdf}}\hfill%
	\subfigure{\includegraphics[width=0.33\linewidth]{figures-c/workpoints/far_wp_dc2.pdf}}\hfill%
	\subfigure{\includegraphics[width=0.33\linewidth]{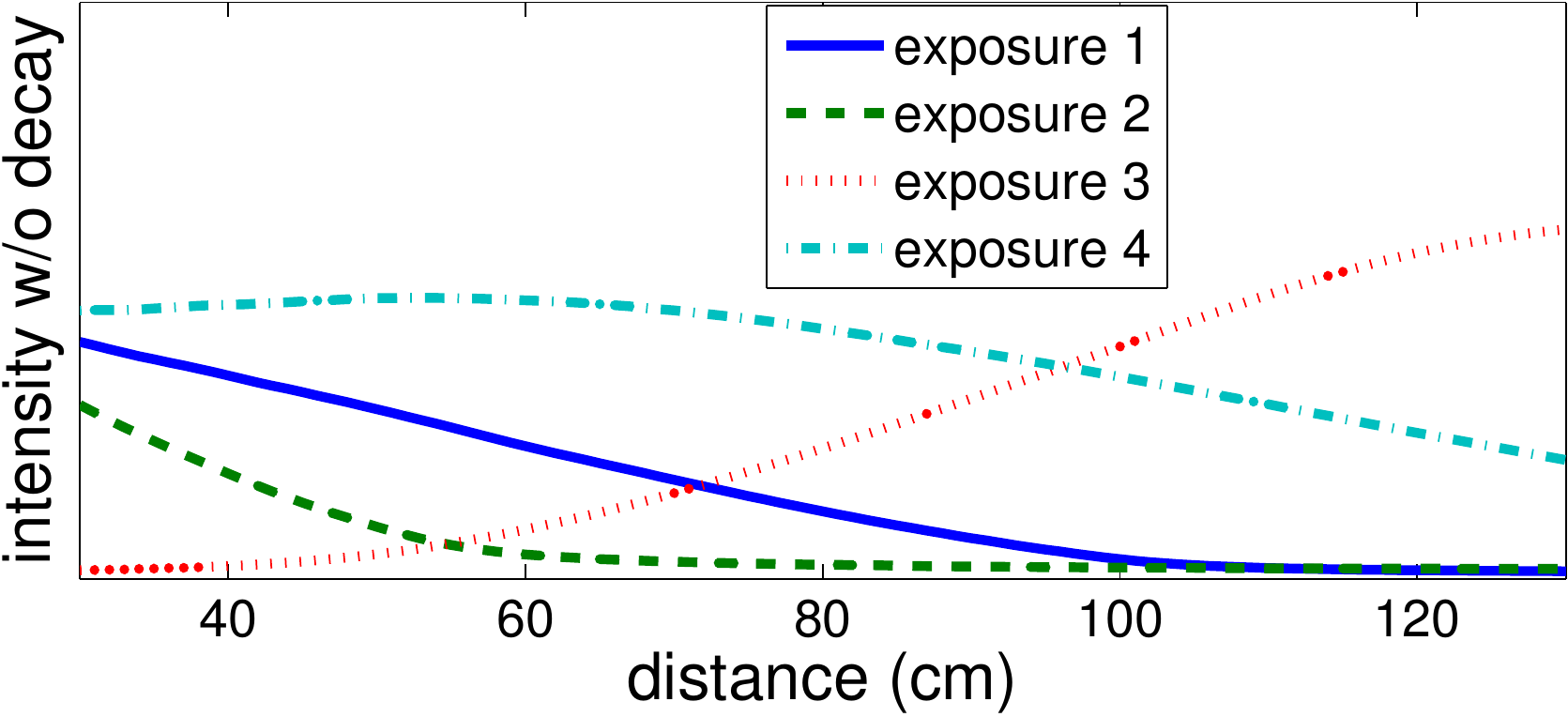}}%
\\%
%
	\subfigure{\includegraphics[width=0.33\linewidth]{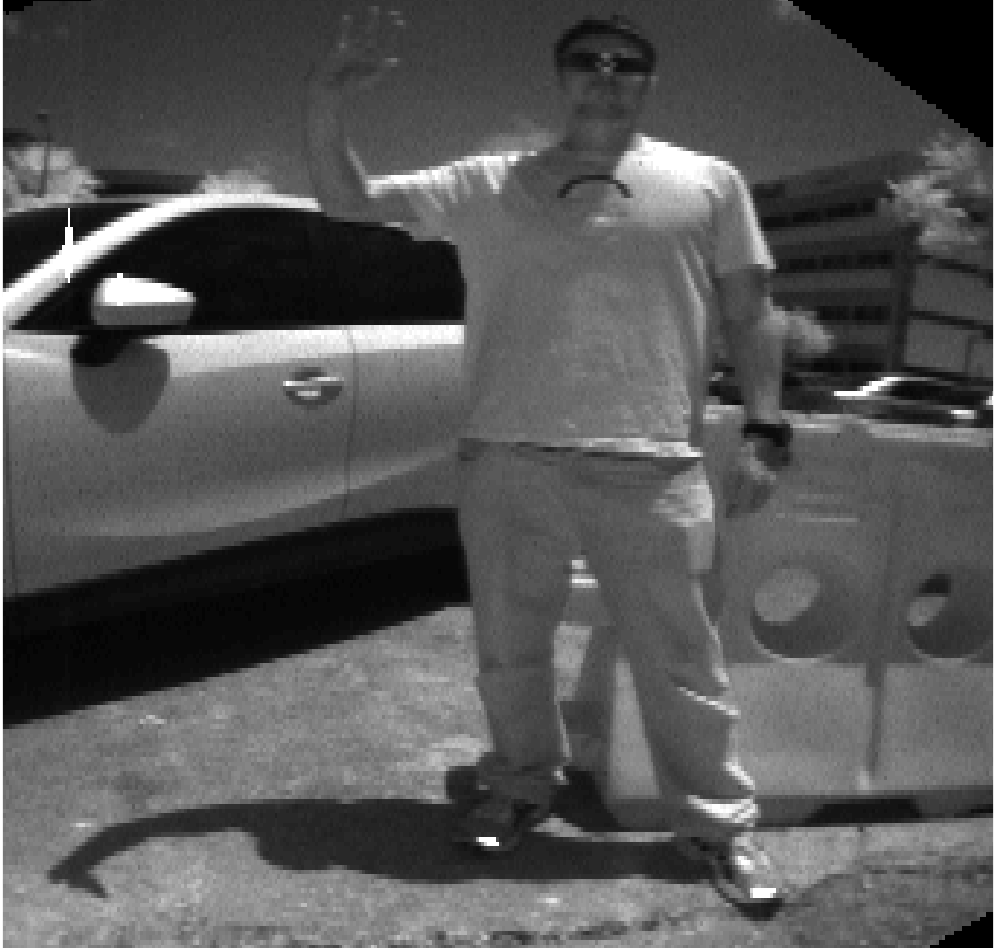}}\hfill%
	\subfigure{\includegraphics[width=0.33\linewidth]{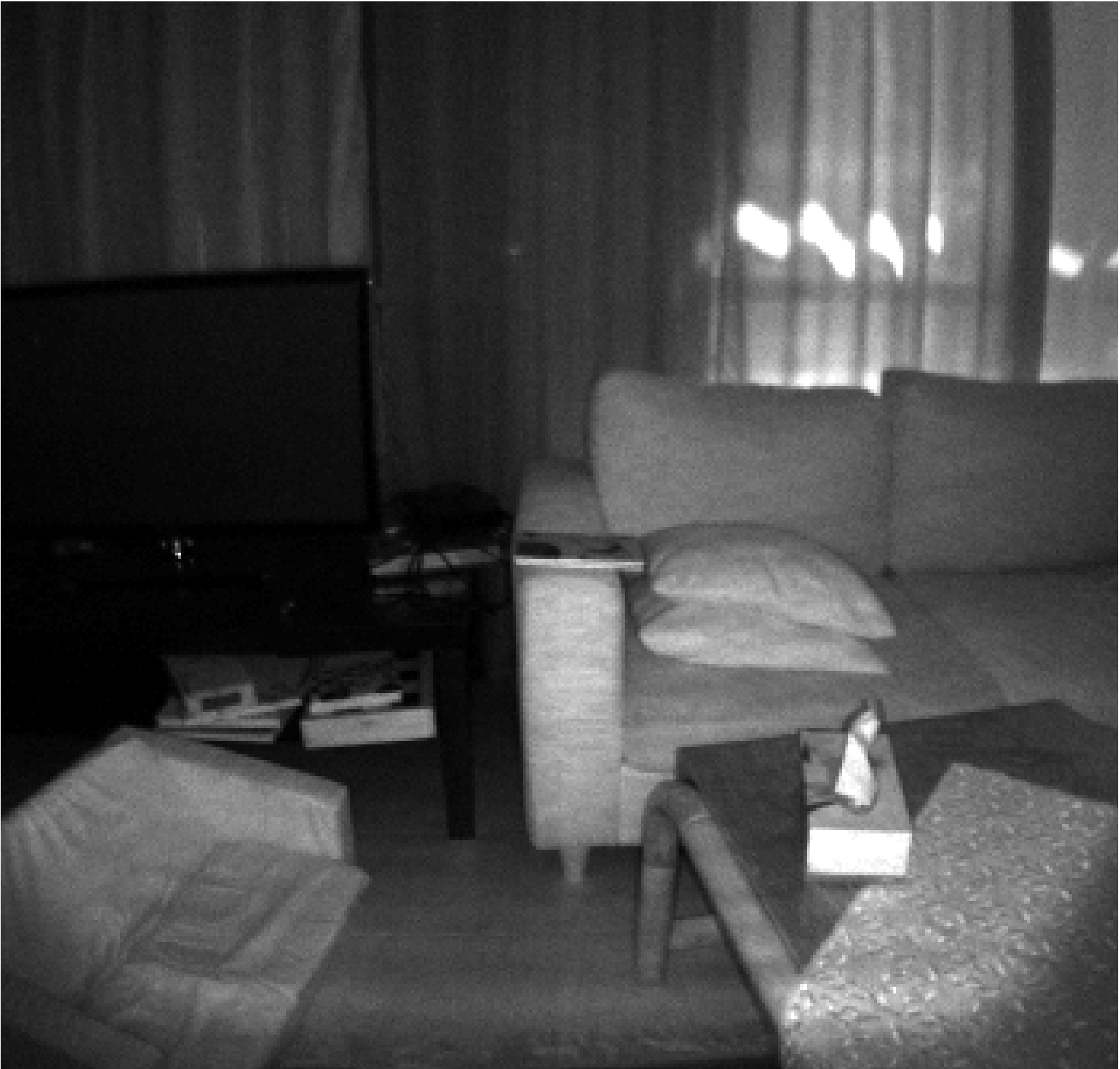}}\hfill%
	\subfigure{\includegraphics[width=0.33\linewidth]{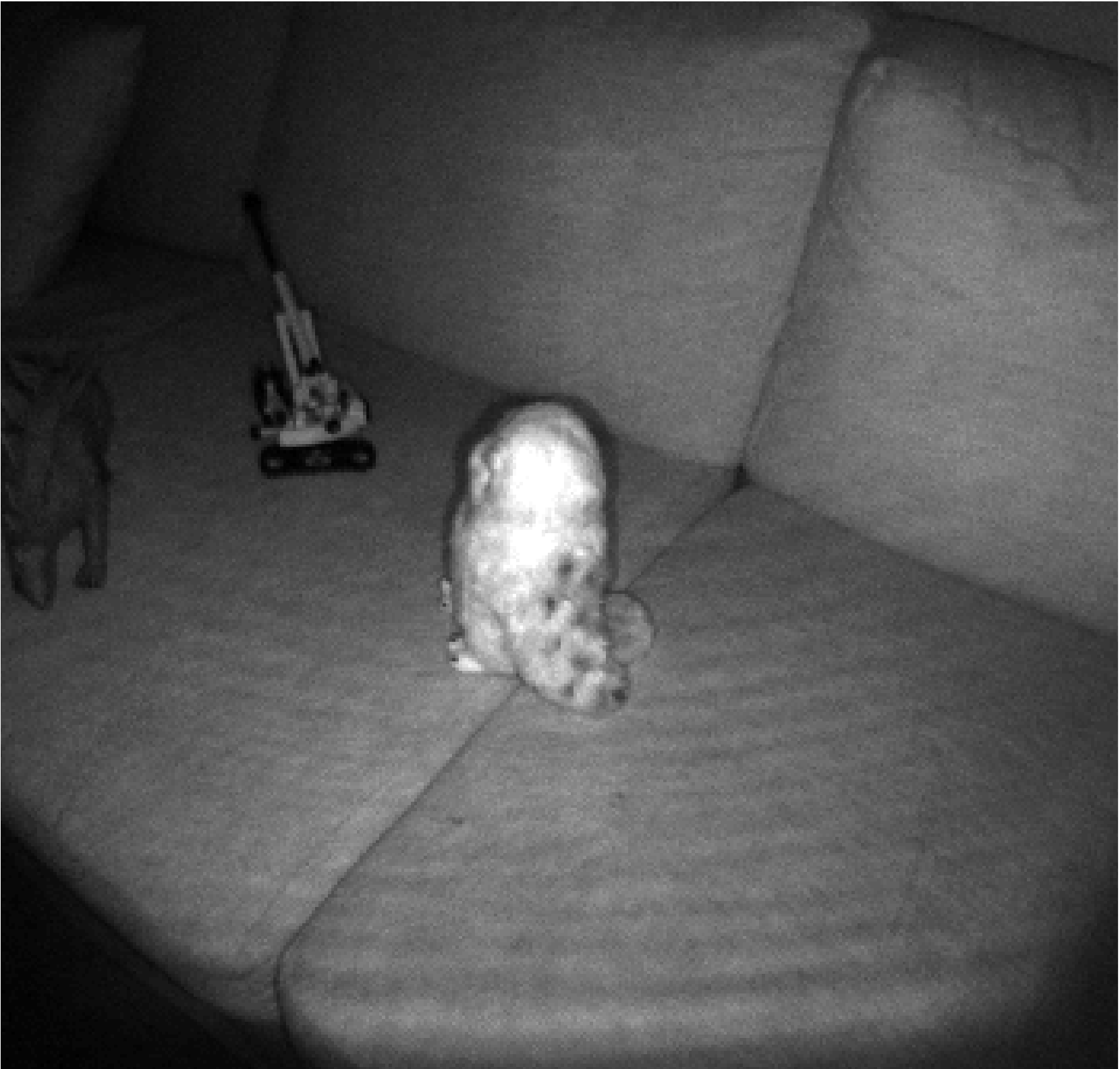}}%
\\%
%
	\subfigure{\includegraphics[width=0.33\linewidth]{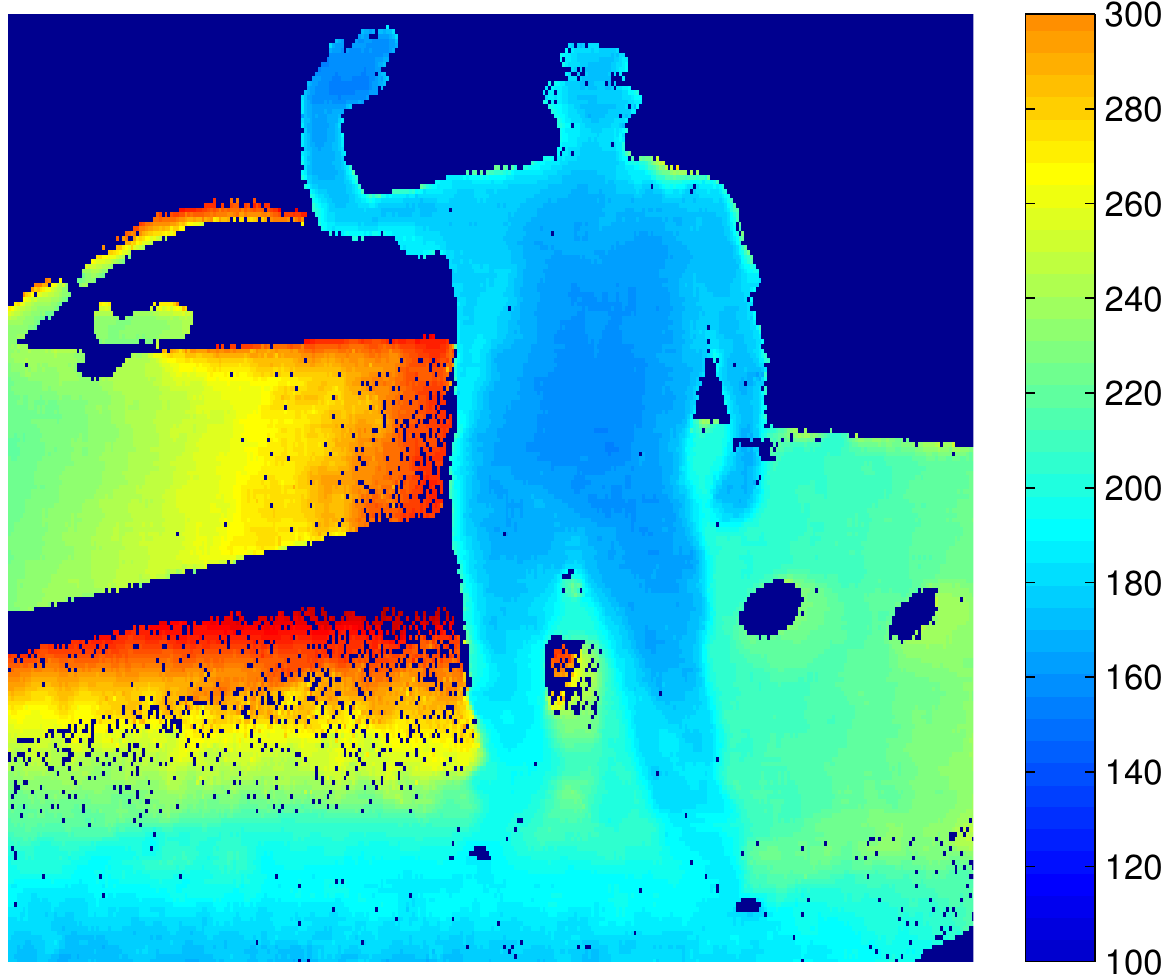}}\hfill%
	\subfigure{\includegraphics[width=0.33\linewidth]{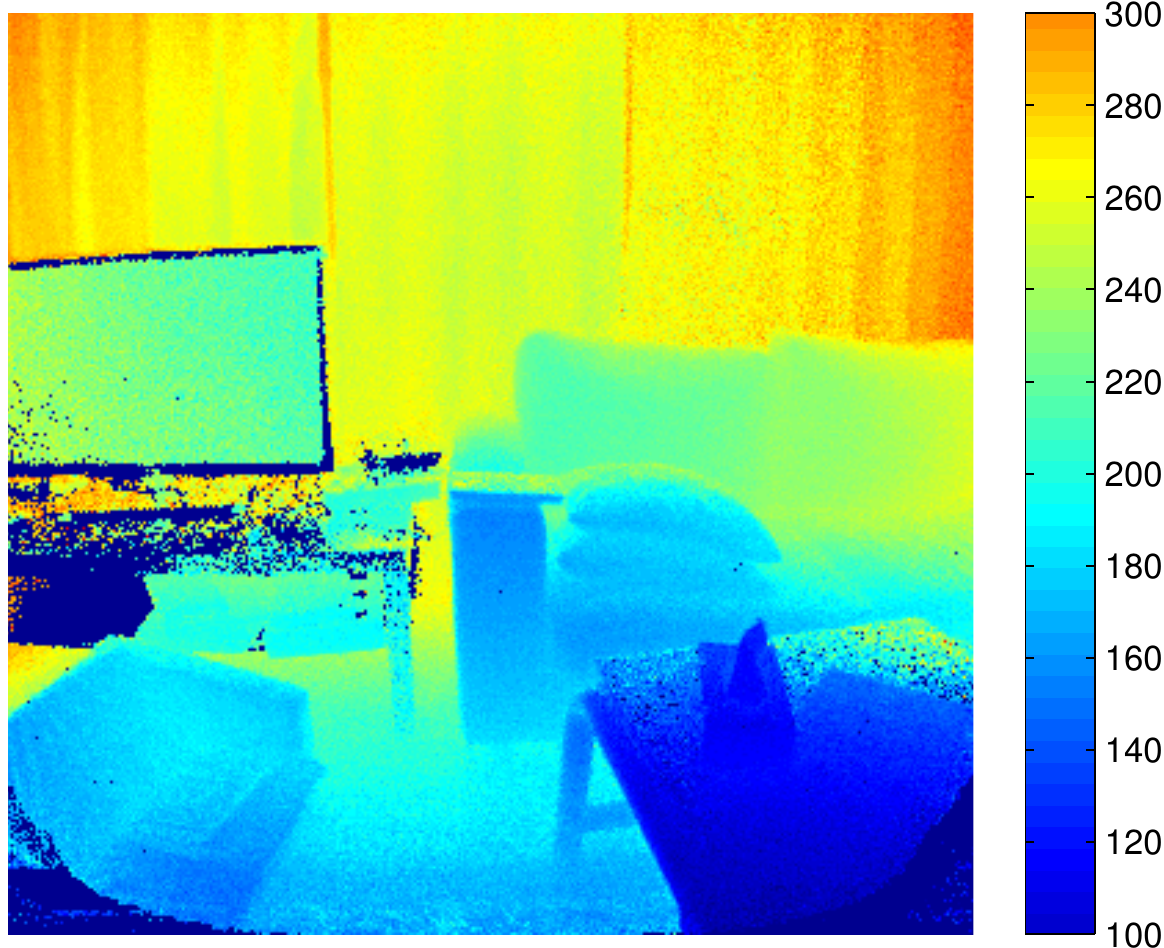}}\hfill%
	\subfigure{\includegraphics[width=0.33\linewidth]{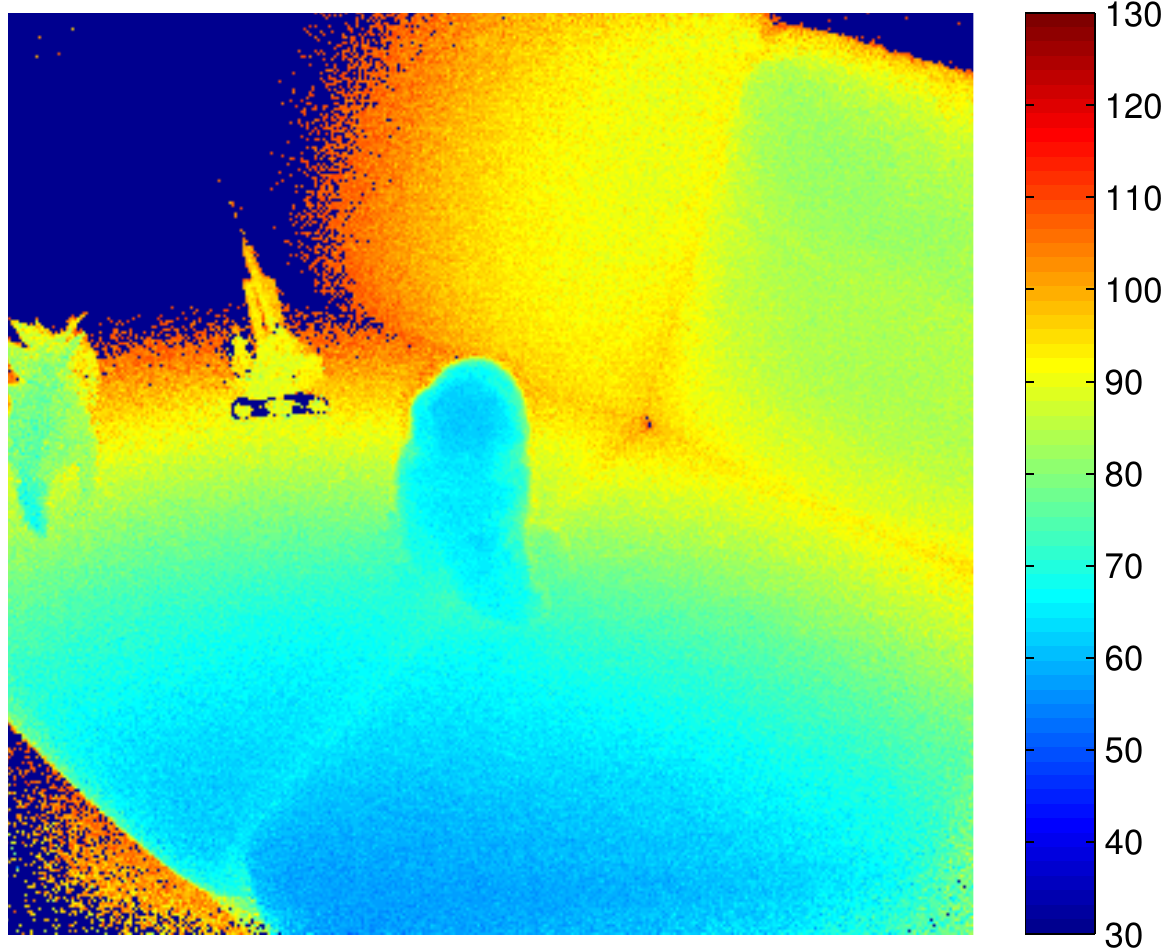}}%
\caption{\footnotesize Sample scenes.
\textbf{Top}: exposure profile used.
\textbf{Middle}: first response image $R_1$.
\textbf{Bottom}: inferred depth image using the SP-MLE model.
The left and middle column are scenes with a far-range design, the right
column is a scene with a near-range design.
The designs were obtained using different priors $p(t)$ in~(\ref{prob:wpdesign}).
}
\label{sample_scenes}%
\end{figure}

\subsection{Multipath-Robust Exposure Profile Design}
\label{sec:mp-robust}

In the exposure profile design objective~(\ref{prob:wpdesign}) we take two
expectations: the first over prior imaging conditions (prior $p$) and the second over the
assumed forward model (forward model $P$, equation~(\ref{eqn:model1})).
This indeed is the way to minimize the loss when responses come from our basic generative model, which does not include multipath.

We now want to design an exposure profile that will be more resistant to
multipath. Therefore we should measure the loss over responses that also include multipath.
We use our realistic simulator for that as follows.
Given one or multiple 3D scenes and their realistic light transport pathes, we
sample responses from these scenes. Formally, the scenes and the simulator are
a more complex generative model $G$.
We denote the sampling from this complex model by $(\vec{R},t) \sim G$, but do
keep in mind that the model $G$ uses multiple reflectivity values and ambient
lighting along the pathes to generate the response $\vec{R}$.
%
%
Both $P$ and $G$ depend on the design $\Z$ through~(\ref{generative_mean})
and~(\ref{eqn:simforward}), respectively.

We combine both generative models in a mixture:
a fraction $\beta \in [0;1]$ are samples from our assumed model
prior $p$ and $P$, and a fraction $1-\beta$ are samples from the physical
simulation prior $Q$.
Then the expectation~(\ref{prob:wpdesign}) becomes
{\small
\begin{equation}
\beta \: \E_{t,\rho,\lambda \sim p}
	\E_{\vec{R} \sim P}[\ell(\hat{t}(\vec{R}),t)]
	+ (1-\beta) \: \E_{(\vec{R},t) \sim G}[\ell(\hat{t}(\vec{R}),t)].
	\label{eqn:mpath-prior}
\end{equation}
}
%
We see that the design objective~(\ref{prob:wpdesign}) can, by a simple change
as in~(\ref{eqn:mpath-prior}), accommodate richer priors over scenes and
effects such as multipath.
We demonstrate this in section~\ref{sec:wpdesign-experiments}.


\section{Experimental Results}
\label{results}
We use a prototype camera as shown in Figure~\ref{fig:page1}.
In our experiments we avoid reference and comparison with other depth cameras
in terms of noise characteristics and variance of depth estimates because the
validity of such comparison is affected by hardware configurations such as
power used, field of illumination, resolution, thermal design constraints, and
sensor sensitivity.
Instead we focus on demonstrating the validity of our model, inference
procedures, and regression approximations.

Throughout the experiments we will use the abbreviations SP and TP to refer to
the single-path model~(\ref{eqn:model1}) and the two-path
model~(\ref{generative_mean2}), respectively.  Depending on the inference
method we use MAP, MLE, and Bayes, so that TP-Bayes for example means the
two-path model with full Bayesian inference.

\subsection{Sample scenes}
We start with a few sample scenes shown in Figure~\ref{sample_scenes}.
We designed two exposure profiles using two different uniform priors on depth
$p(t)$ in~(\ref{prob:wpdesign}).
The first prior focused on larger depths while the second prior focused on
smaller ranges.
The two left columns show outdoor and indoor scenes using the far range
exposure profile, and the right column shows a scene captured with the short
range profile.
The middle row shows the first response image, and the bottom row shows the
inferred depth (obtained using the regression tree).

\begin{figure}[t!] \centering
	\subfigure[$R_1$]{%
		\includegraphics[width=0.24\linewidth]{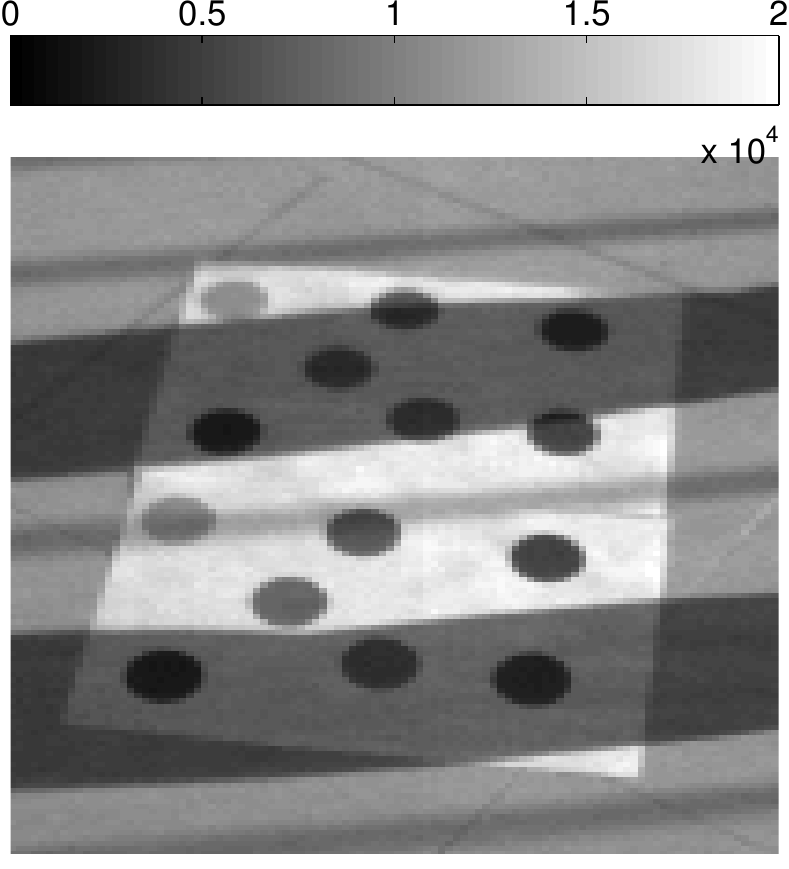}%
		\label{fig:circles-a}%
	}\hfill%
	\subfigure[$\hat{\sigma}_{\textrm{mle}}(\vec{R})$]{%
		\includegraphics[width=0.24\linewidth]{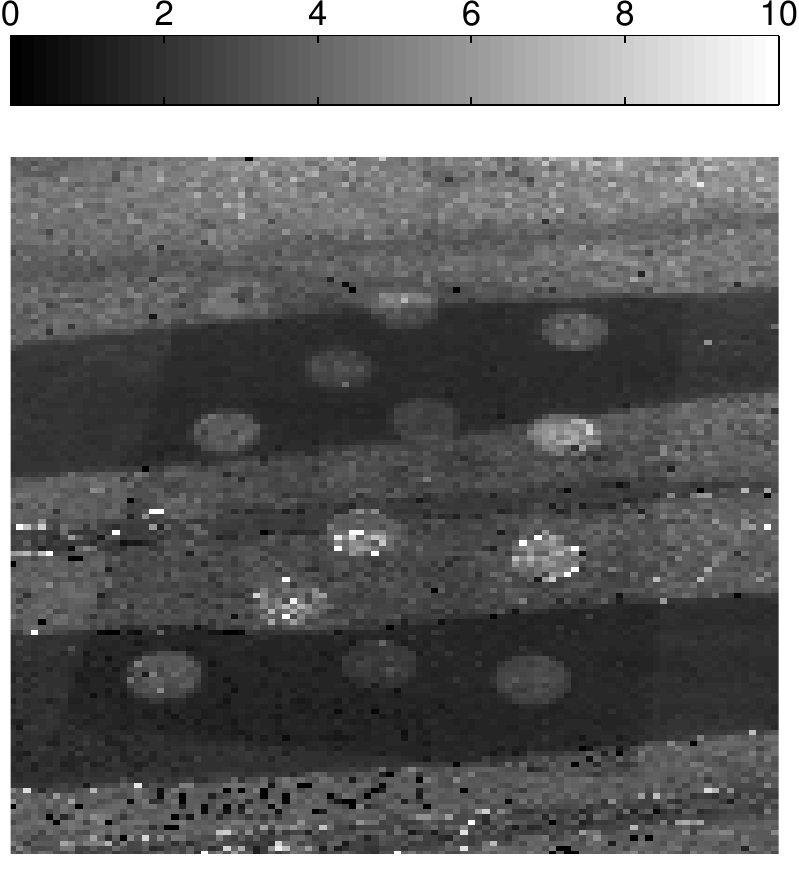}%
		\label{fig:circles-b}%
	}\hfill%
	\subfigure[$\hat{\rho}_{\textrm{mle}}(\vec{R})$]{%
		\includegraphics[width=0.24\linewidth]{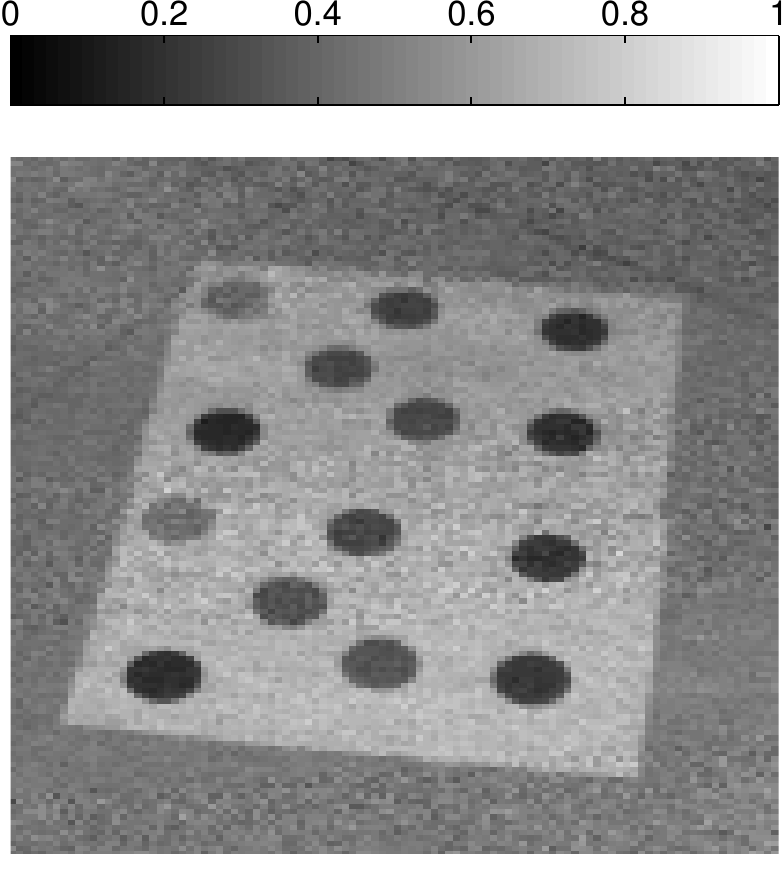}%
		\label{fig:circles-c}%
	}\hfill%
	\subfigure[$\hat{\lambda}_{\textrm{mle}}(\vec{R})$]{%
		\includegraphics[width=0.24\linewidth]{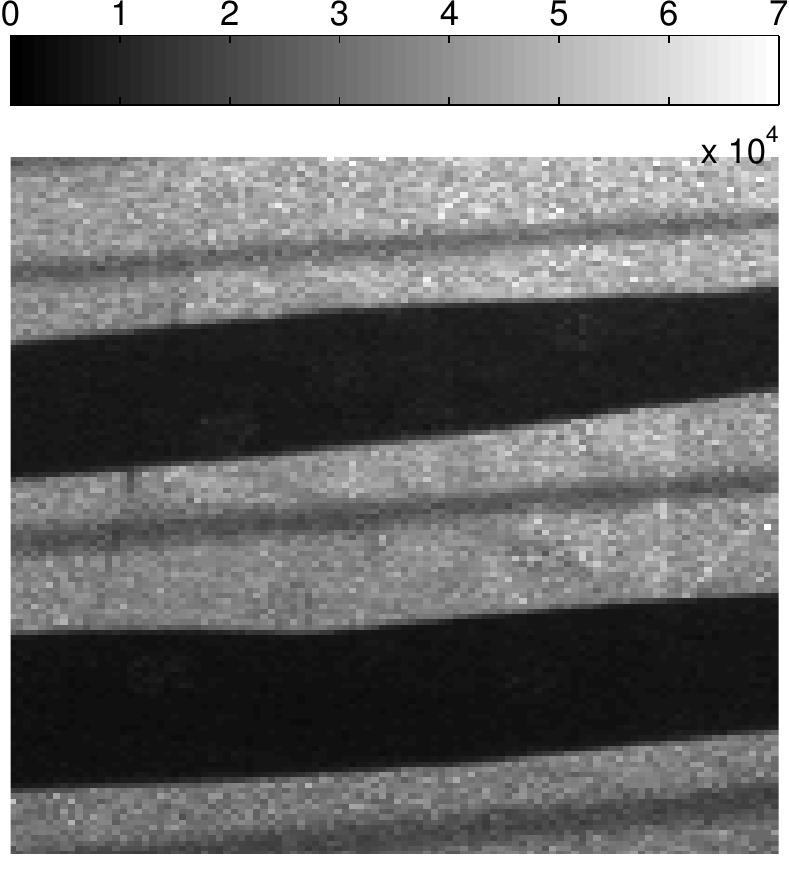}%
		\label{fig:circles-d}%
	}%
\caption{\footnotesize %
Posterior inference results under different illuminations and albedos.
\subref{fig:circles-a}
First response image, exhibiting varying ambient light levels and albedos,
including strong shadows.
\subref{fig:circles-b}
Posterior depth uncertainty (cm), higher under either stronger ambient light
or lower albedo.
\subref{fig:circles-c}
Inferred albedo map, in $[0;1]$.
\subref{fig:circles-d}
Inferred ambient illumination map.
}
\label{circles_std}
\end{figure}

\subsection{Accurate Depth Uncertainty}
\label{sec:exp-depthuncertainty}

Next we show that by accurately modeling the noise present in the observed response our model is
able to assess its own uncertainty in the inferred depth.
%
To demonstrate this
we capture 200 frames of a static scene as shown in
Figure~\ref{fig:sceneresponse} and sample 500 pixel locations in the shown
box.
%


%
Since the camera is static, we can obtain the empirical standard deviation of
the depth estimators for each of the 500 points. We plot this empirical depth
uncertainty, against the predicted uncertainty obtained in the first frame as
described in Section~\ref{sec:depthuncertainty22} (the predicted uncertainty
is nearly identical over all 200 frames).
Figure~\ref{fig:depthuncertainty} shows the good agreement between the
predicted uncertainty and the actual uncertainty.
This provides empirical data about how well the model is
\emph{calibrated}~\cite{dawid1982calibration}, that is, how accurately it
judges the uncertainty in its own predictions.

To gain some insight on what determines depth standard deviation, we turn to
Figure~\ref{circles_std}, showing a part of the scene shown in Figure~\ref{fig:page1}.
In Fig.~\ref{fig:circles-a}, we see one of the input responses,
showing the combined effect of different albedos and shadows.
In Fig.~\ref{fig:circles-b} we see how imaging conditions affect the
variance of the depth estimates.
In the shadowed regions the ratio between the active illumination and ambient
light is higher, and this generally leads to a tighter posterior.
On materials with higher albedo (the white page vs the dark circles) the amount of
reflected light is higher and this also leads to smaller variances (as
compared with the variances on dark circles which reflect less light).
In addition, the depth itself affects the measure of uncertainty but this is
not illustrated in this zoomed scene.

\subsection{Ambient and Effective Reflectivity}
In our model, the inferred albedo image is illumination-invariant and
therefore does not contain shadows.
Therefore we can perform realtime shadow
removal~\cite{finlayson2002removingshadows,xiao2014shadowremovalrgbd},
providing illumination-invariant inputs to computer vision algorithms.
This is illustrated in Fig.~\ref{fig:circles-c}. In Fig.~\ref{fig:circles-d}
we show the estimated ambient light level at each pixel.

For more results on realtime extraction of illumination, reflectivity and shape
please view the enclosed video.

\begin{figure}[t!] \centering
	\subfigure{%
		\includegraphics[width=0.48\linewidth]{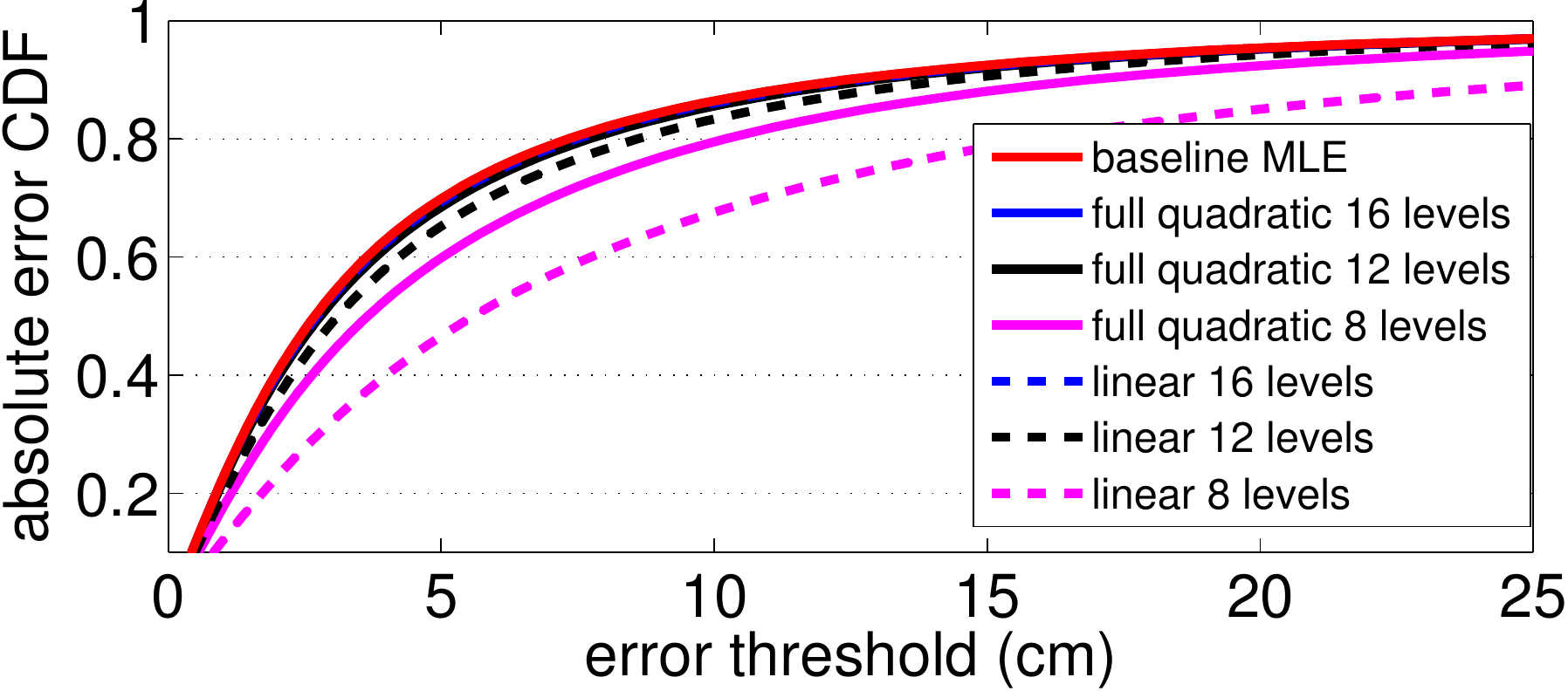}%
		\label{fig:terr-a}%
	}\hfill%
	\subfigure{%
		\includegraphics[width=0.48\linewidth]{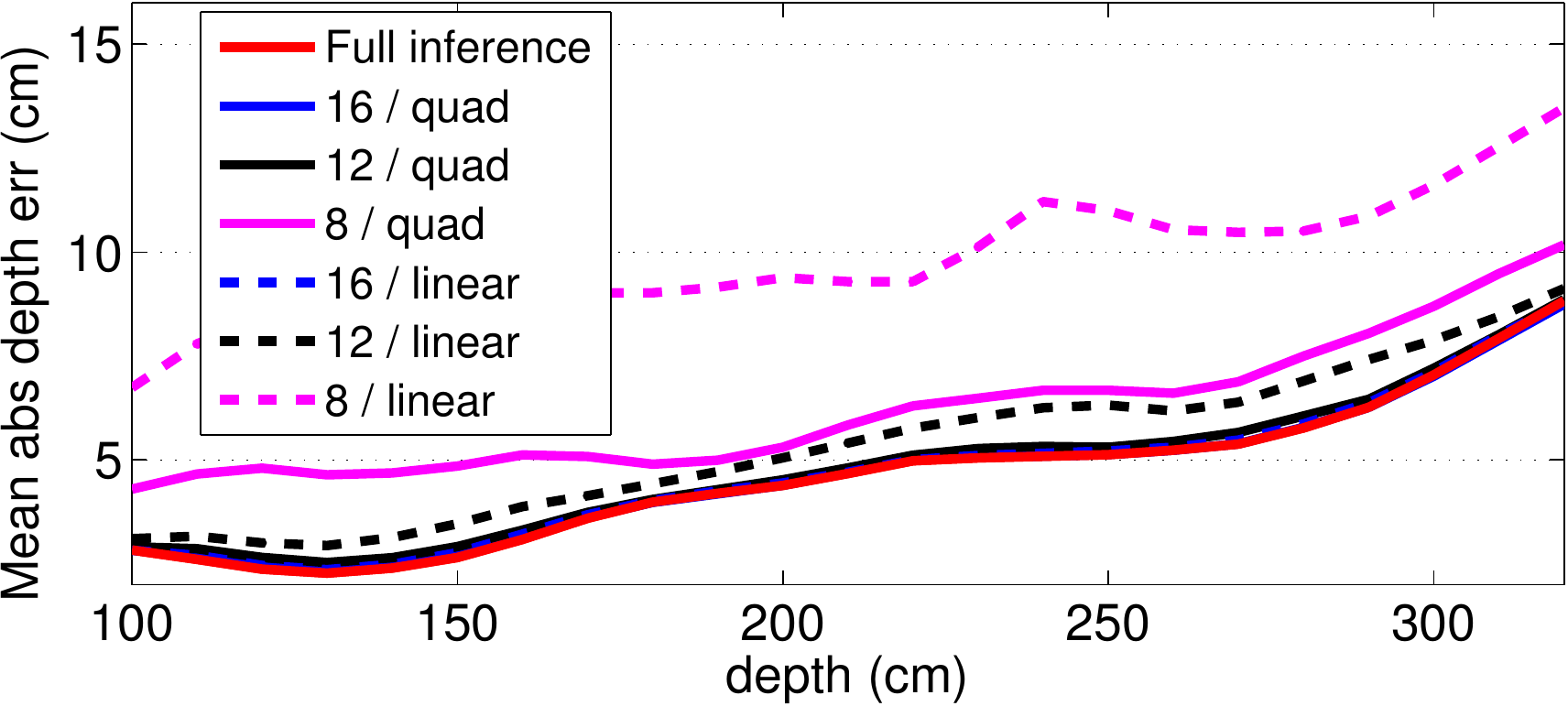}%
		\label{fig:terr-b}%
	}\quad%
\caption{\footnotesize Regression tree errors compared to full inference.
\textbf{Left}:
Cumulative error distribution over test set.
\textbf{Right}:
mean absolute error over prior albedo and ambient levels.}
\label{tree_errors}
\end{figure}

\subsection{Regression Tree Approximation Quality}

As discussed we use regression trees to regress depth, thus approximating full
inference which is infeasible in realtime.
An optimized implementation running on a \emph{Intel HD Graphics 4400 GPU}, evaluates
a regression tree of depth $12$ with full quadratic polynomials on a
200-by-300 pixel frame in $2.5\textrm{ms}$. This means we are able to run four
trees (depth, illumination, albedo and depth std) effectively at
$\approx 100\textrm{fps}$. The enclosed video shows this implementation running
(the std was computed but not shown in the output windows).

We now quantify the additional errors incurred due to the use of regression
trees instead of full inference.
The added error depends on the tree structure, which determines required
memory resources as described in section~\ref{sec:rf}.  We tested two types of
trees at three depths, yielding six possible tree structures. The two types of
trees used either a linear polynomial or a quadratic polynomial on the leafs.
The depths we used were $8, 12$ and $16$ (full binary trees).

After training the trees, we generate test data by sampling from the prior
imaging conditions and our generative model~(\ref{eqn:model1}).
We compute the baseline error by running full inference (in this case MLE, but similar
results hold for Bayes) on the test data, then run the various tree
predictors.
Fig.~\ref{tree_errors} shows the results.
On the left we plot the cumulative distribution of errors over the test set.
On the right we partition the test set by depths, and show the
mean absolute error for each depth, averaged over all albedos and illumination levels.
We see that as the trees get deeper the quality approaches that of
the full inference.  At $16$ levels they essentially match.

\begin{figure}[t!] \centering
	\subfigure{%
		\includegraphics[width=0.47\linewidth]{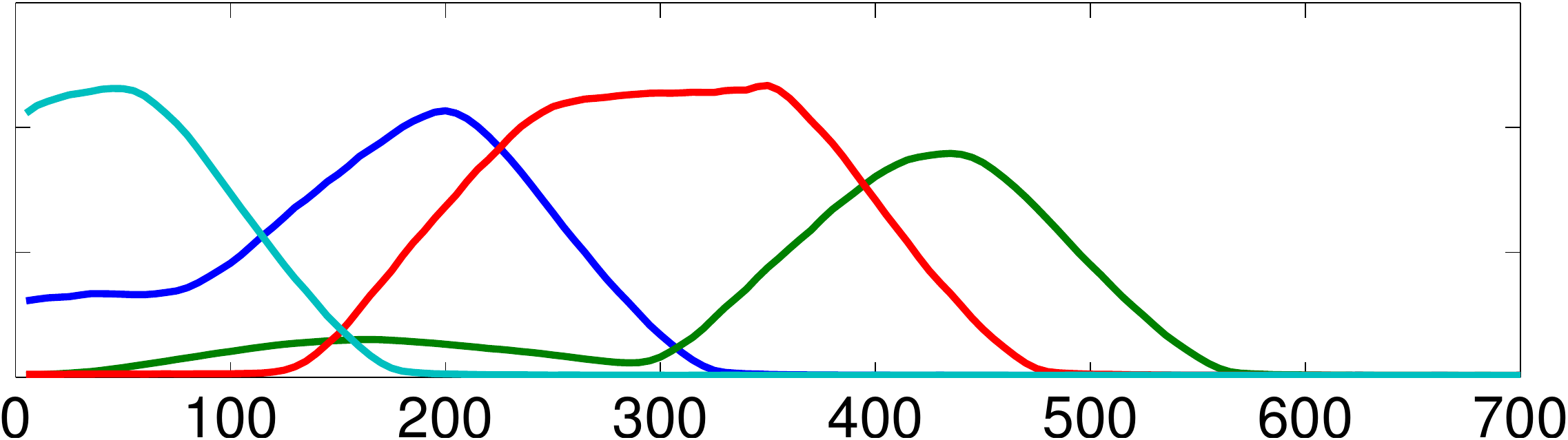}%
	}\quad%
	\subfigure{%
		\includegraphics[width=0.47\linewidth]{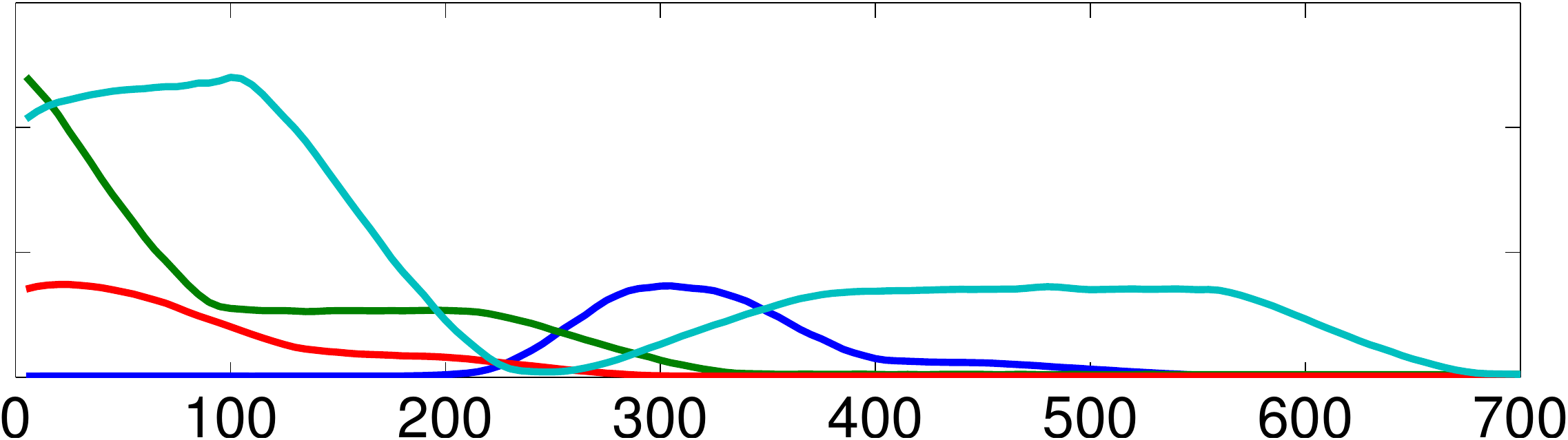}%
	}\\%
	\subfigure{%
		\includegraphics[width=0.47\linewidth]{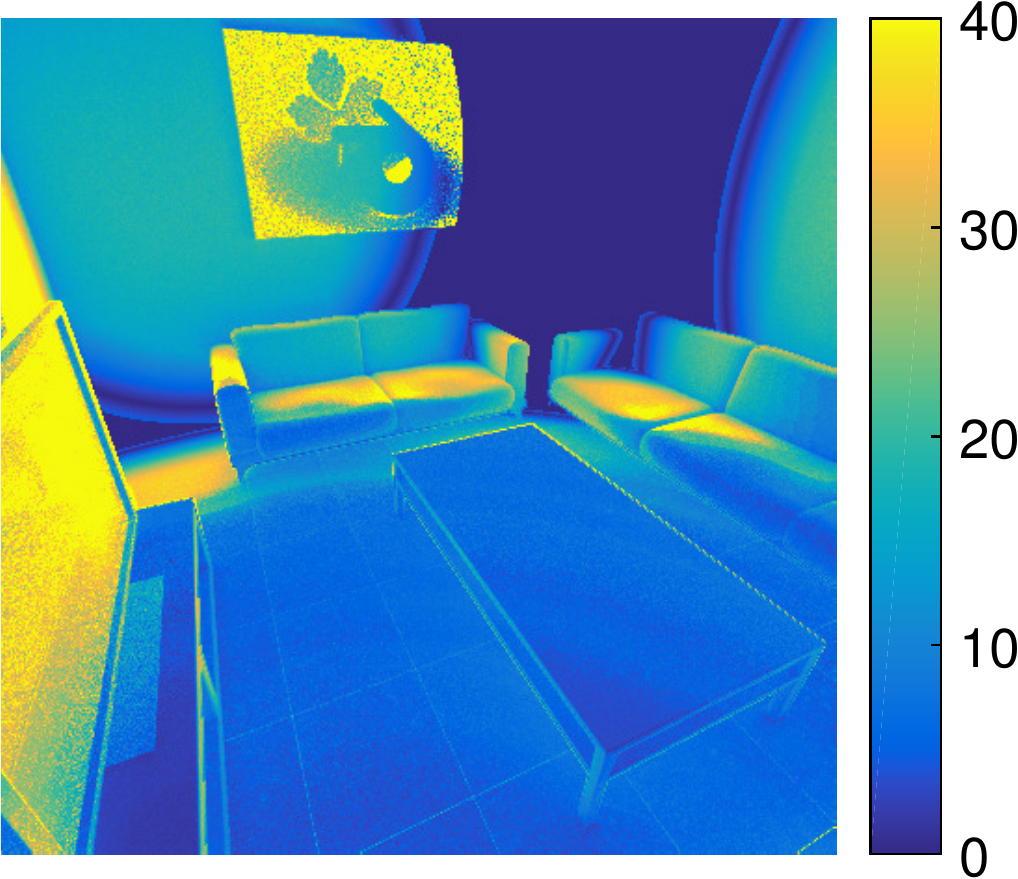}%
	}\quad%
	\subfigure{%
		\includegraphics[width=0.47\linewidth]{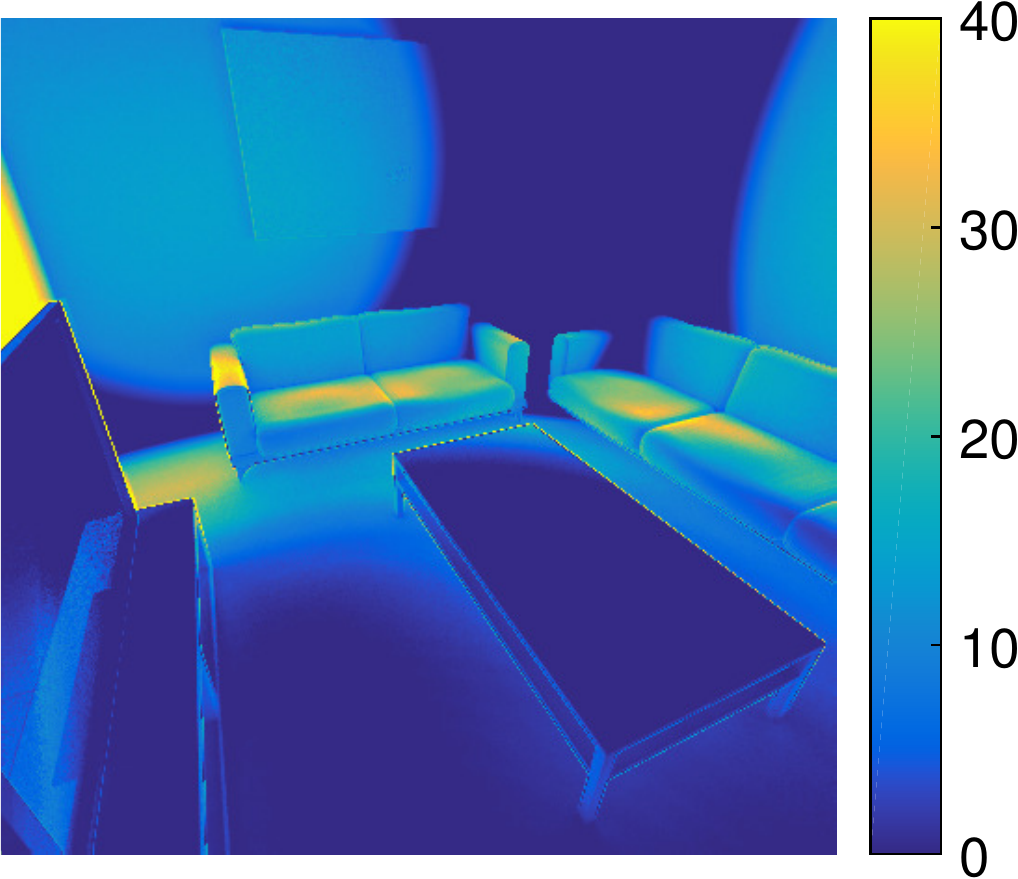}%
	}\\%
	\subfigure{%
		\includegraphics[width=0.47\linewidth]{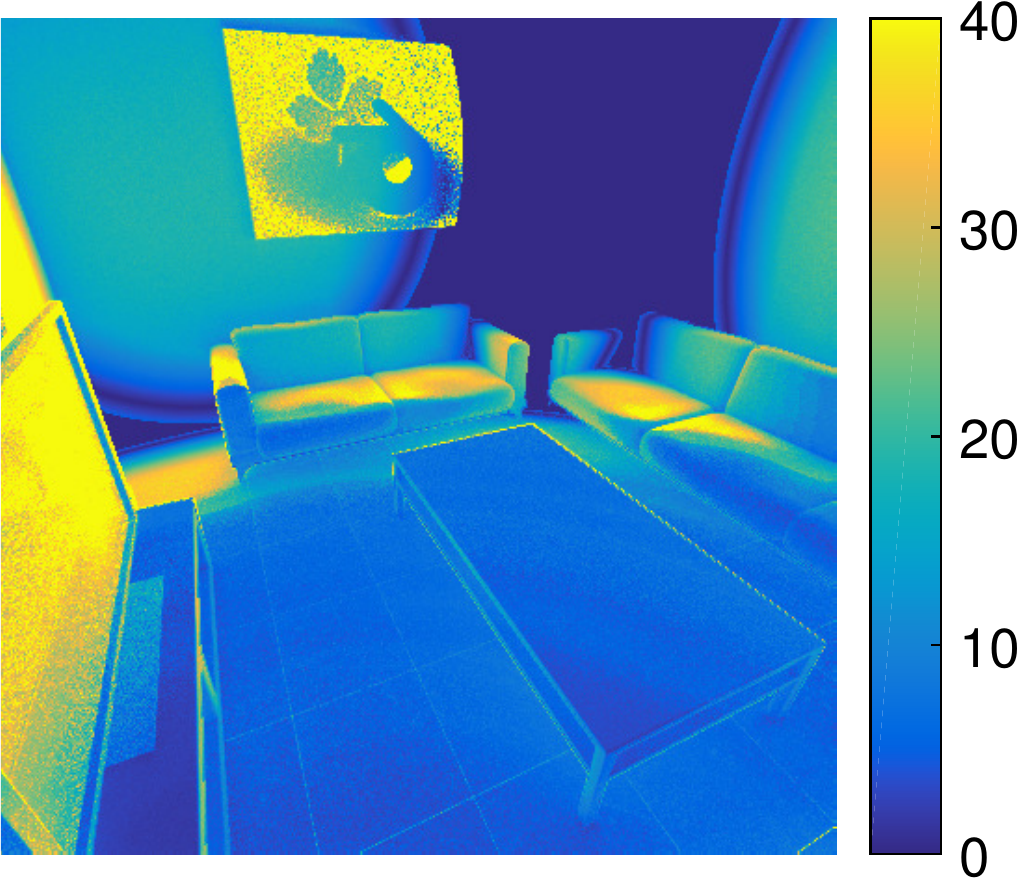}%
	}\quad%
	\subfigure{%
		\includegraphics[width=0.47\linewidth]{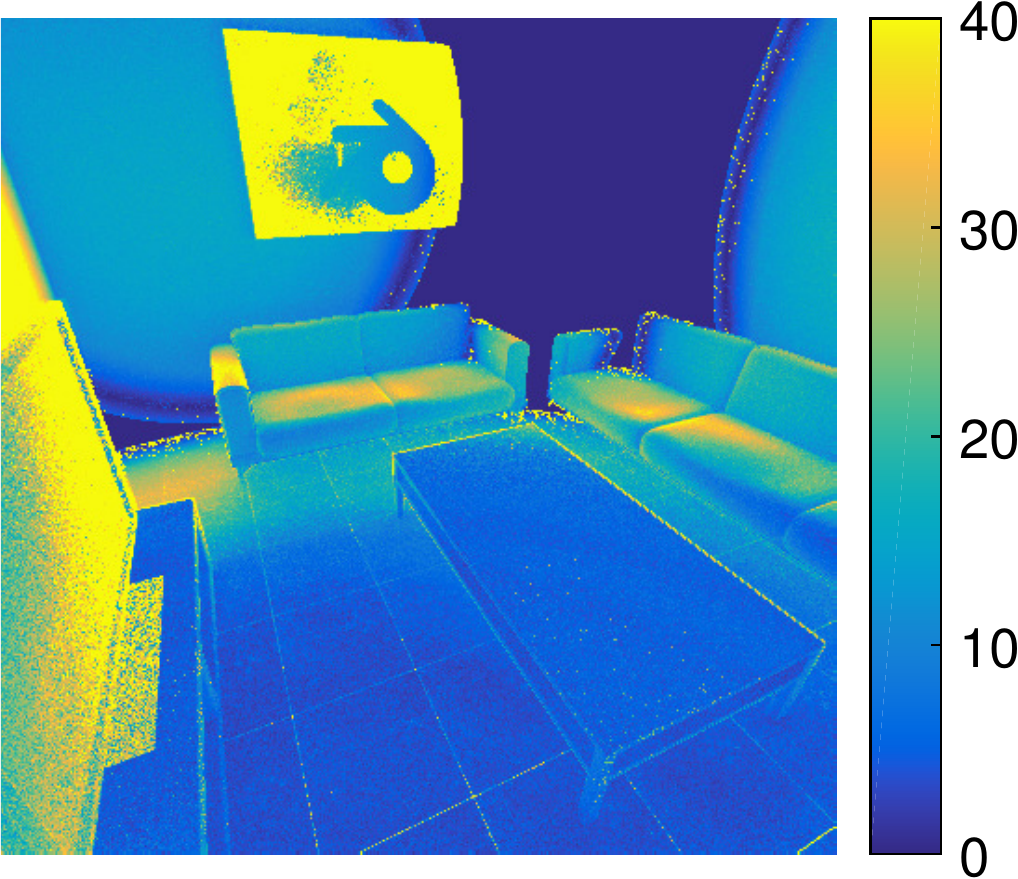}%
	}%
\caption{\footnotesize Multipath-robust exposure profile design.
\textbf{Left column}: original exposure profile.
\textbf{Right column}: multipath-optimized exposure profile.
\textbf{From top-to-bottom}:
\emph{Top};
regular exposure profile and robust-to-multipath profile.
\emph{Middle};
bias magnitude (cm) of resulting depth inference images.
\emph{Bottom};
Simulated RMSE (cm) including noise.
}
\label{fig:maslulim-workpoint}%
\end{figure}

\subsection{Design for Multipath Robustness}\label{sec:wpdesign-experiments}
Now we demonstrate how we may use our simulator
for obtaining exposure profiles designed to be robust to multipath.
In section~\ref{sec:mp-robust} we allowed for more complex
generative models during exposure profile design, one that will generate
responses that are ``contaminated'' by multipath.
We took two simple scenes of an object in front of highly reflective wall
(scene provided in supplementaries) and used them as the model $Q$
as in section~\ref{sec:mp-robust}.
For the mixture model we used $\beta = 0.5$.
We then ran our design optimization scheme to obtain a new exposure profile.
Let us call this exposure profile the MP-resistant design.
We compare it with the standard design obtained using $\beta = 1$.
The two designs are tested on the scene shown in Fig.~\ref{maslulim_fig}.
We emphasize that this test scene is different than the scene used in design.
Fig.~\ref{fig:maslulim-workpoint} compares the results we obtain.
The top row shows the regular design on the left, and the MP-resistant design
on the right (obtained using two different values of $\beta$ in~(\ref{eqn:mpath-prior})).
The second row shows the magnitude of the resulting depth bias.
On the left we see significant biases due to multipath (compare
with the multipath map in Fig.~\ref{fig:maslulim-multipathratio}).
With the MP-resistant design we see a significant reduction in bias.
The bottom row shows the expected RMSE under both designs. Please note that
this RMSE contains the jitter in the estimates due to pixel shot noise.
In temporal integration procedures like \cite{DBLP:conf/ismar/NewcombeIHMKDKSHF11}
this error diminishes, and therefore it is more important to have the bias
reduced than have this RMSE reduced.

\subsection{Experimental Verification of Simulation Accuracy}\label{sec:real-synth}
Our light transport simulation is based on an accurate physical model of
light and the simulation results should agree with the real camera.
However, a real time-of-flight camera is a complex system with many components
and potentially unaccounted for interactions between them.
In this section we verify that our simulation serves as a good proxy for the
real system.

\begin{figure*}[t!]%
	\centering%
	\subfigure[IR0 (measured)]{%
		\includegraphics[width=0.138\linewidth]{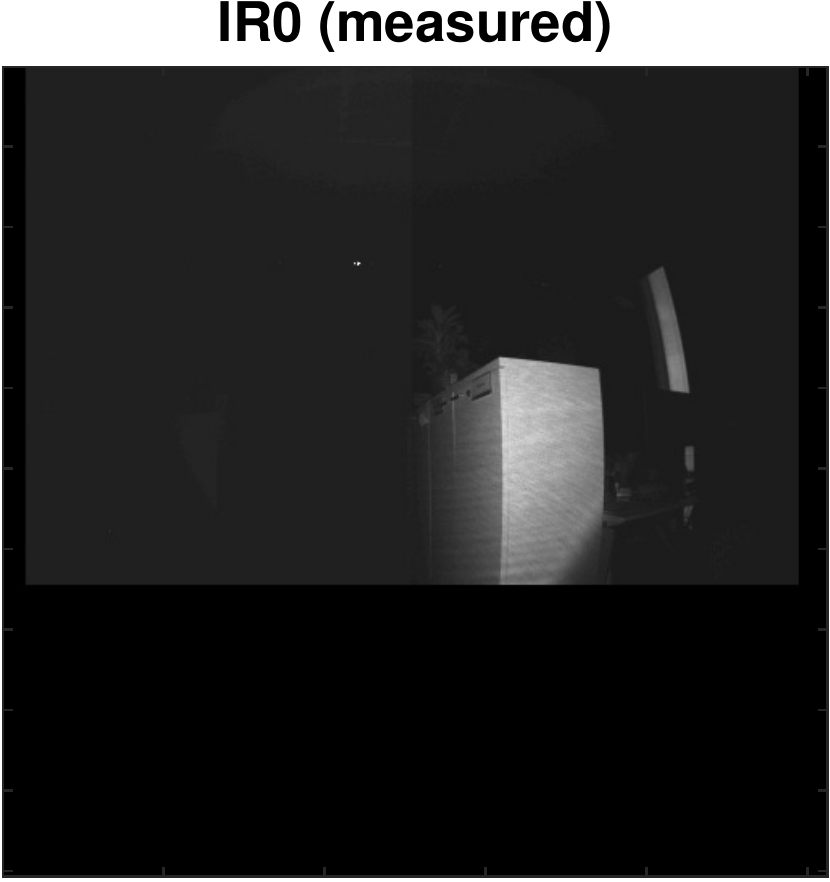}\label{fig:real-nomp-ir0}%
	}\hfill%
	\subfigure[IR0 (syn)]{%
		\includegraphics[width=0.138\linewidth]{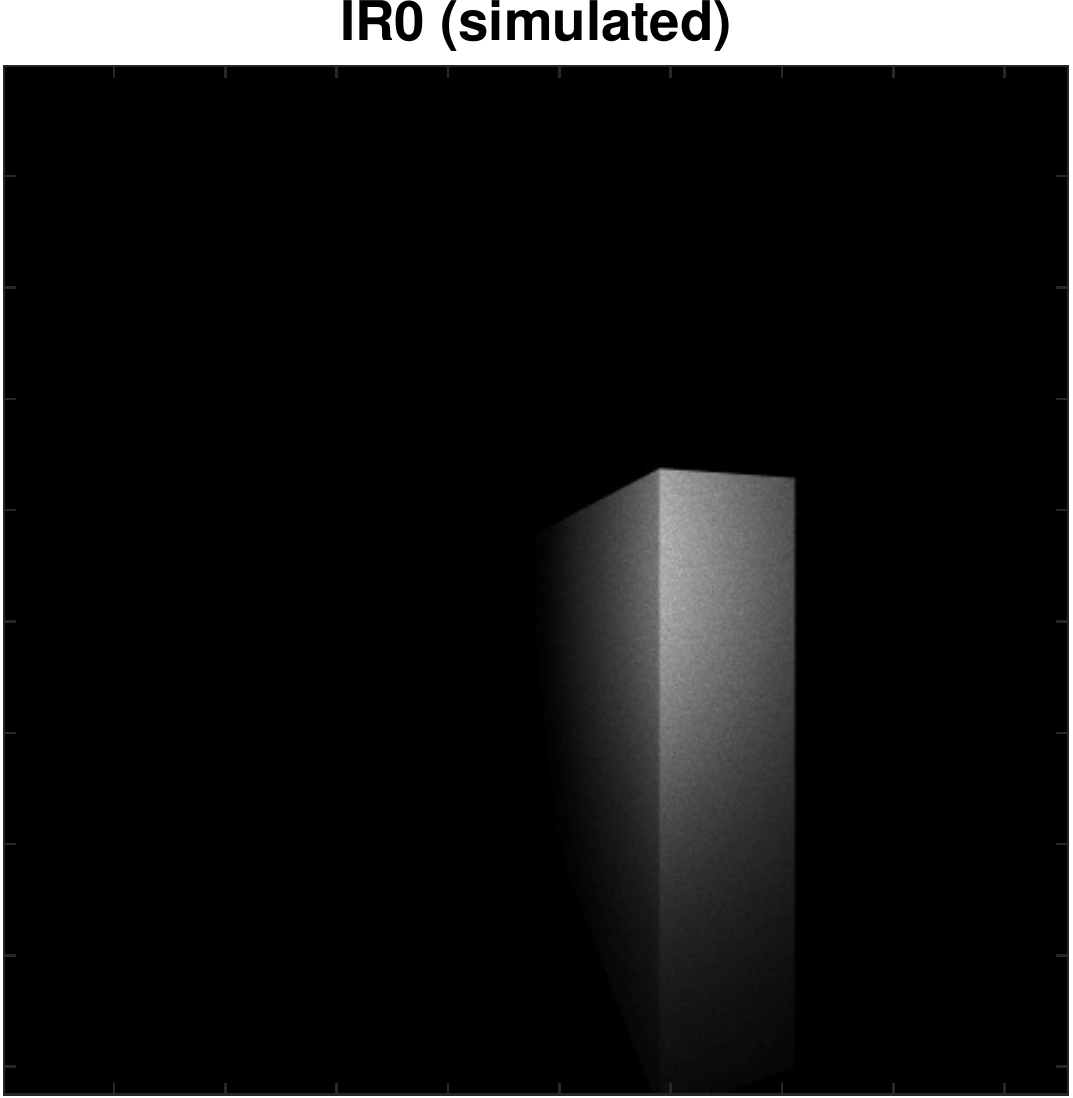}\label{fig:synth-nomp-ir0}%
	}\hfill%
	\subfigure[Scanline (real)]{%
		\includegraphics[width=0.138\linewidth]{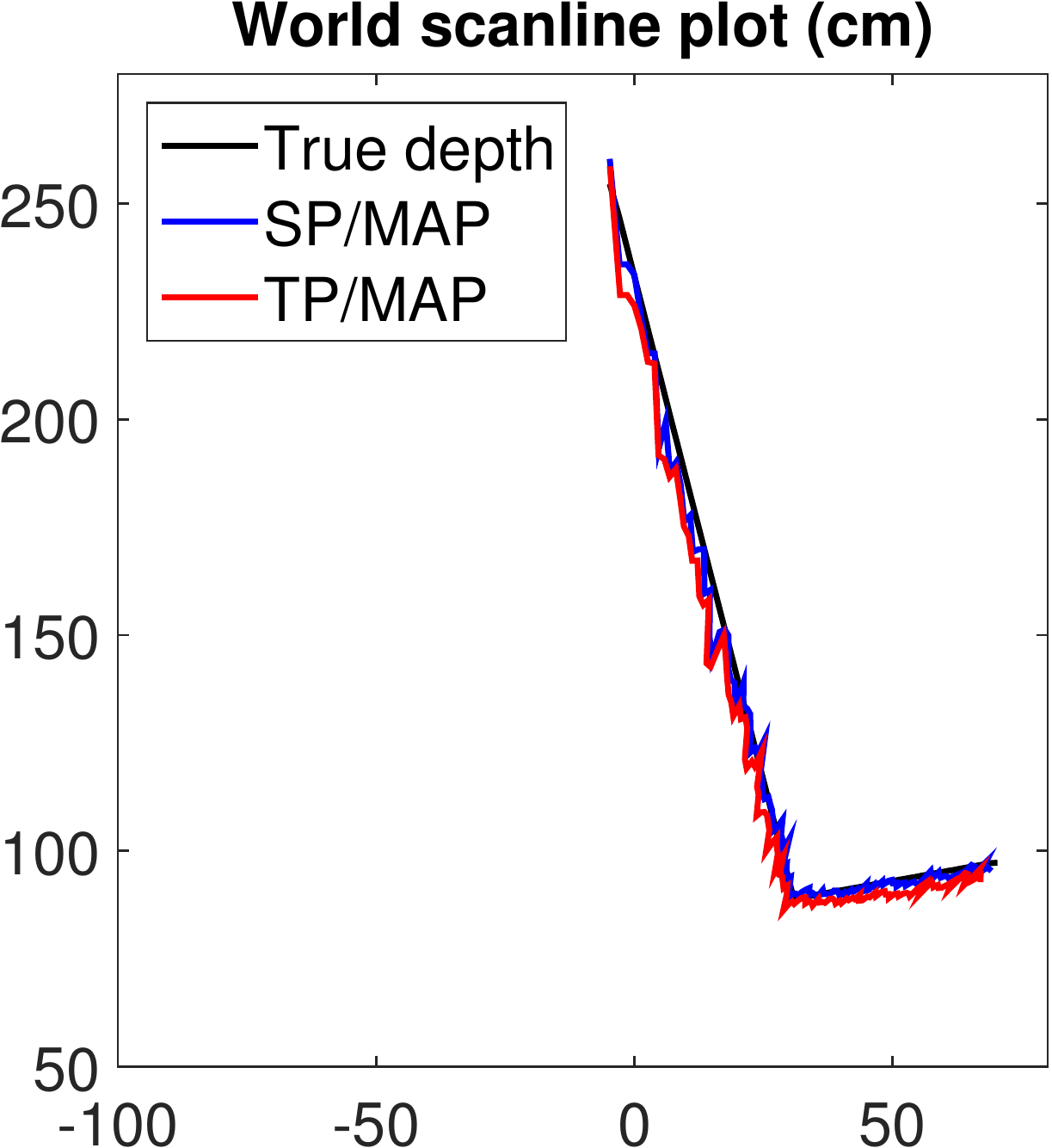}\label{fig:real-nomp-scanline}%
	}\hfill%
	\subfigure[SP err (real)]{%
		\includegraphics[width=0.138\linewidth]{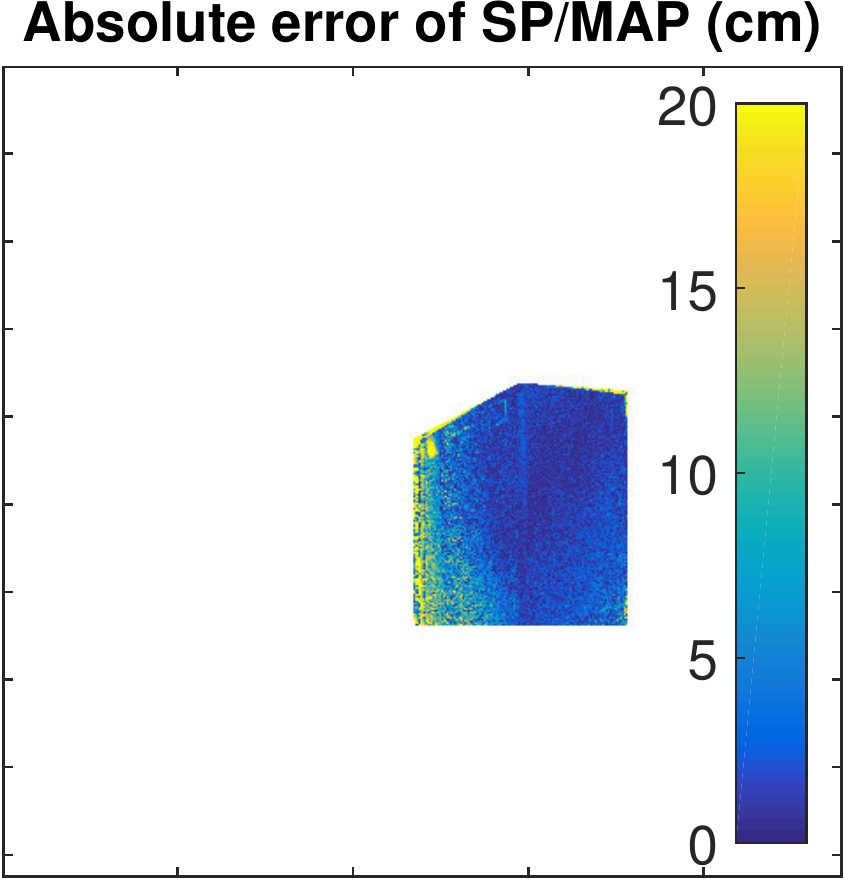}\label{fig:real-nomp-err-spmap}%
	}\hfill%
	\subfigure[SP err (syn)]{%
		\includegraphics[width=0.138\linewidth]{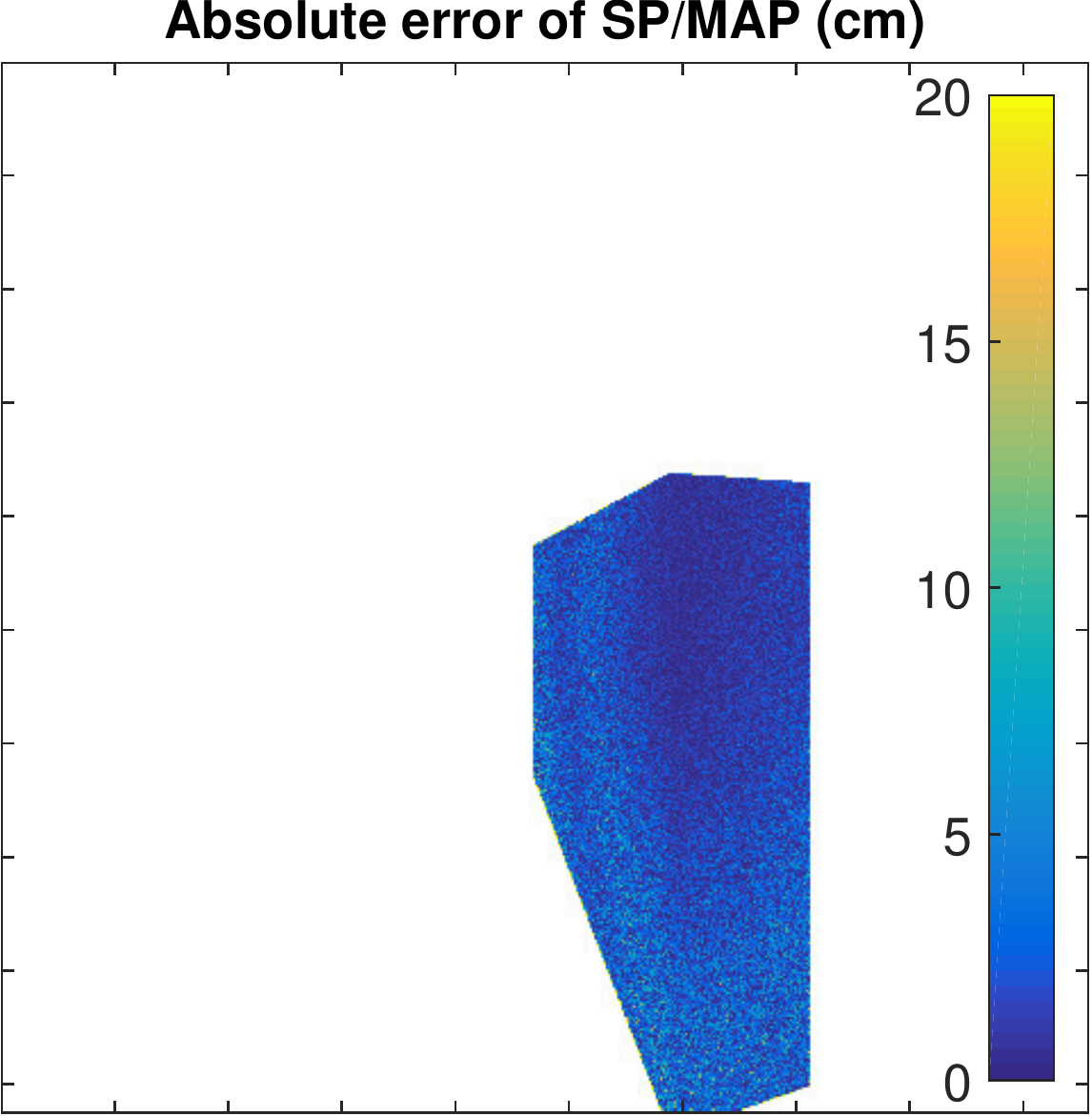}\label{fig:synth-nomp-err-spmap}%
	}\hfill%
	\subfigure[TP err (real)]{%
		\includegraphics[width=0.138\linewidth]{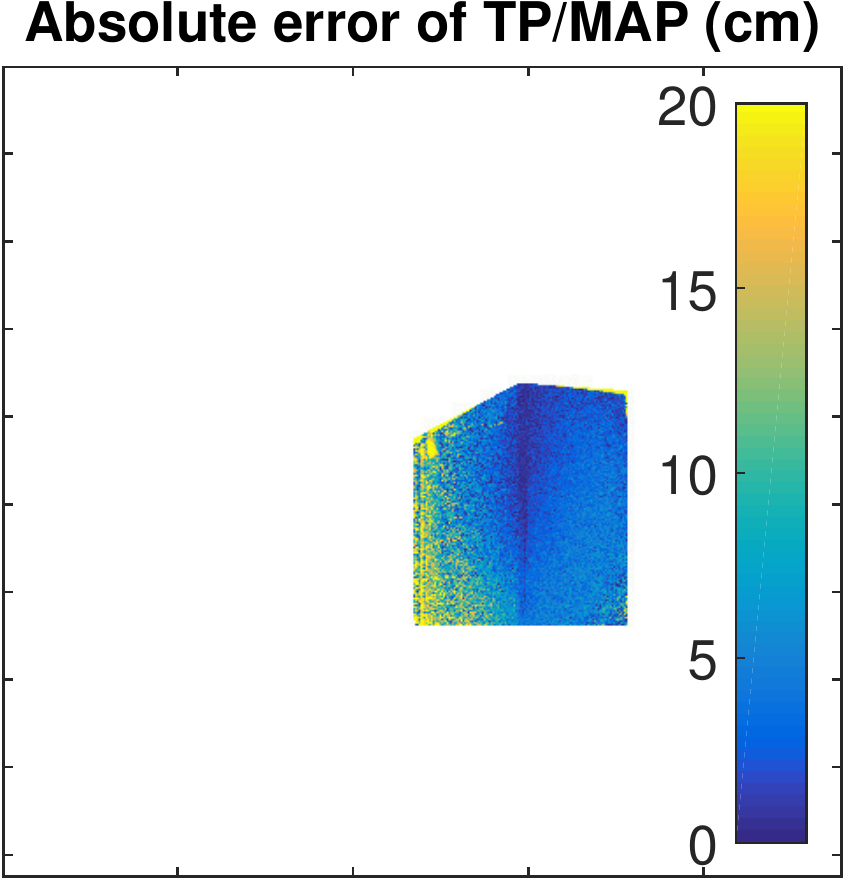}\label{fig:real-nomp-err-tpmap}%
	}\hfill%
	\subfigure[TP err (syn)]{%
		\includegraphics[width=0.138\linewidth]{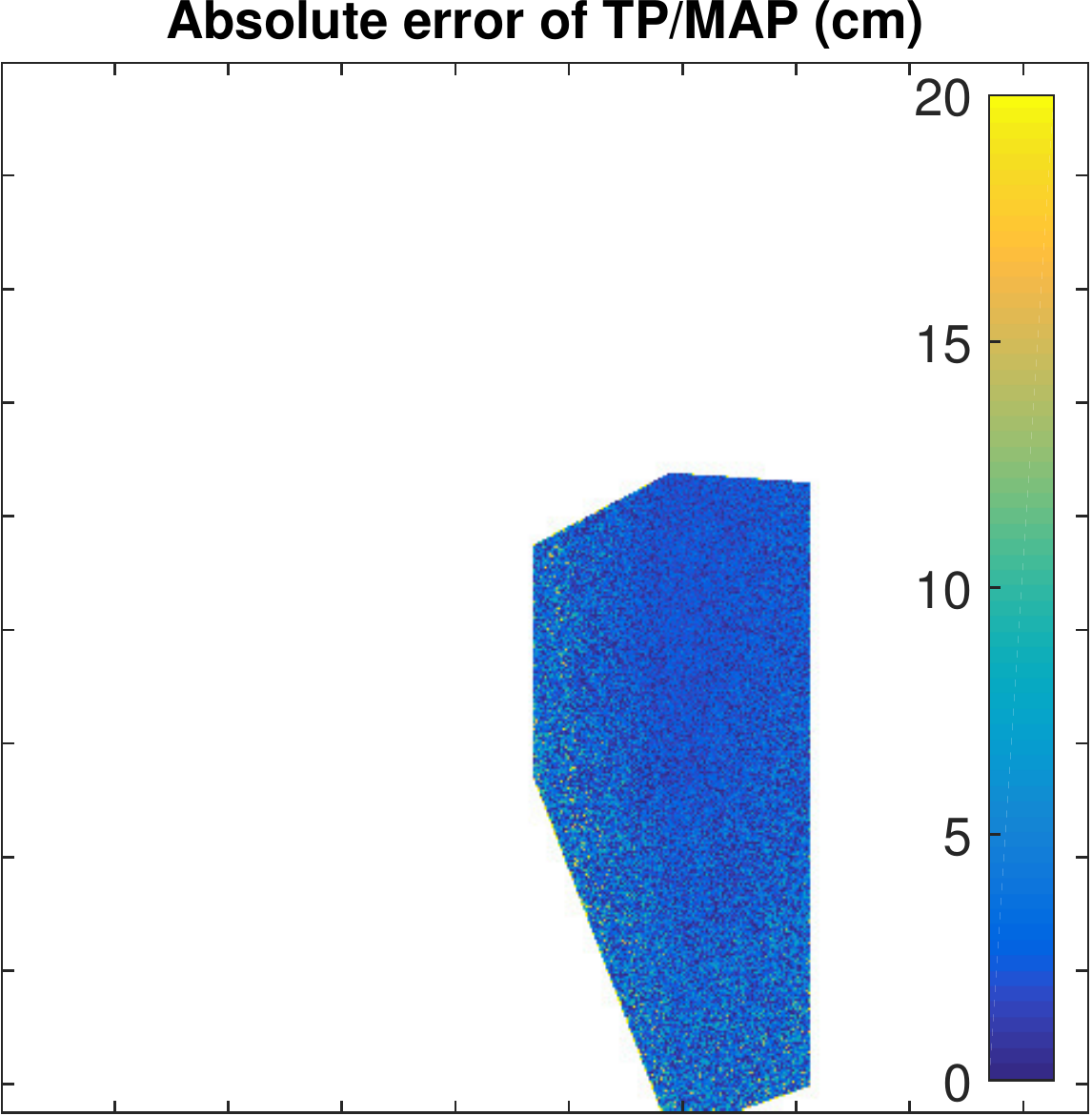}\label{fig:synth-nomp-err-tpmap}%
	}%
\\
	\subfigure[IR0 (measured)]{%
		\includegraphics[width=0.138\linewidth]{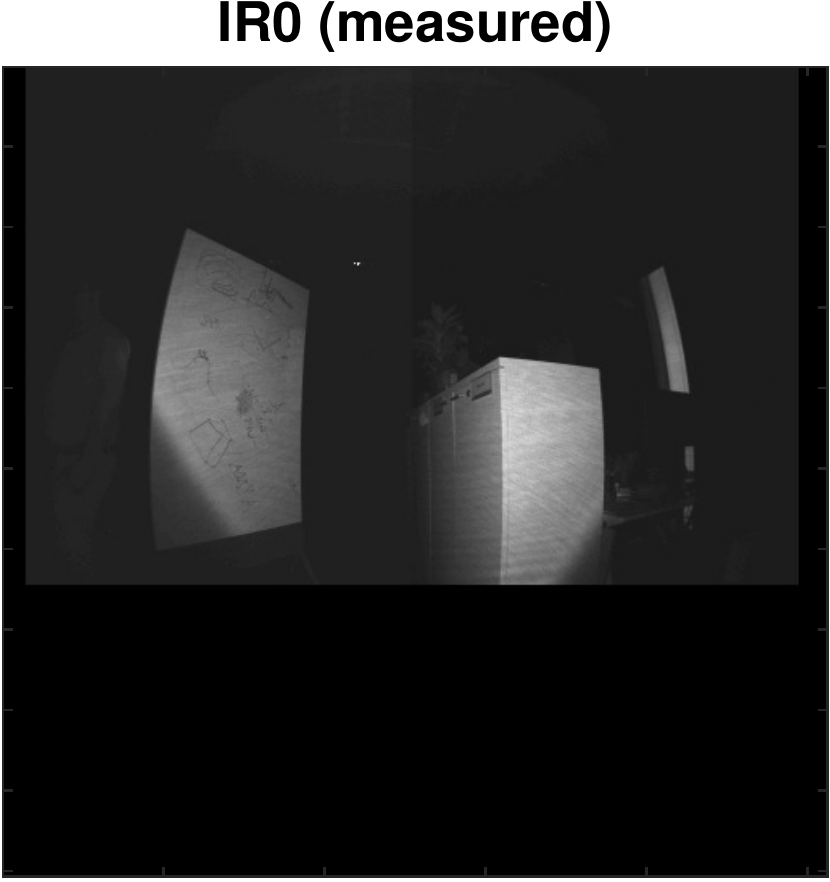}\label{fig:real-withmp-ir0}%
	}\hfill%
	\subfigure[IR0 (syn)]{%
		\includegraphics[width=0.138\linewidth]{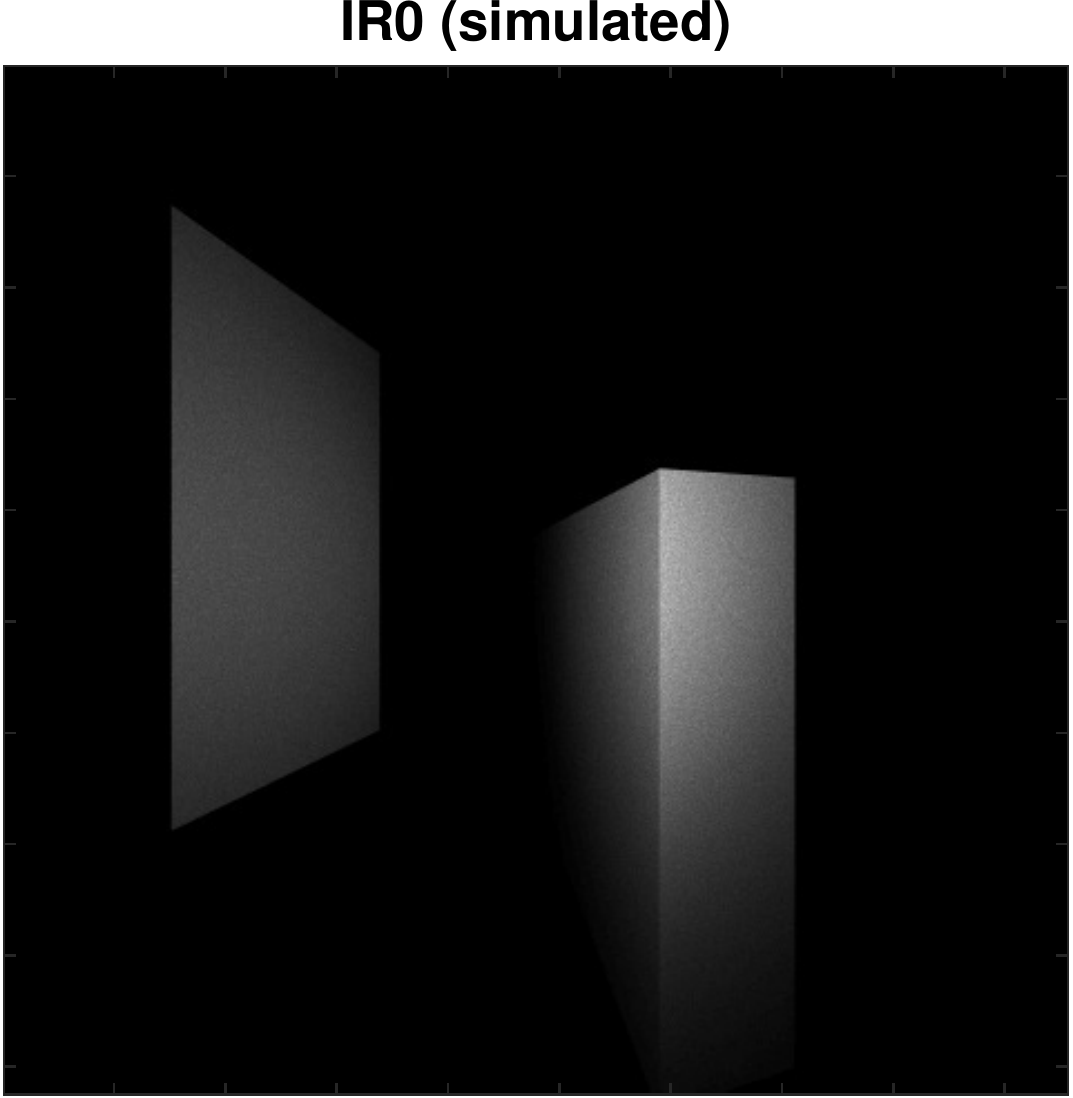}\label{fig:synth-withmp-ir0}%
	}\hfill%
	\subfigure[Scanline (real)]{%
		\includegraphics[width=0.138\linewidth]{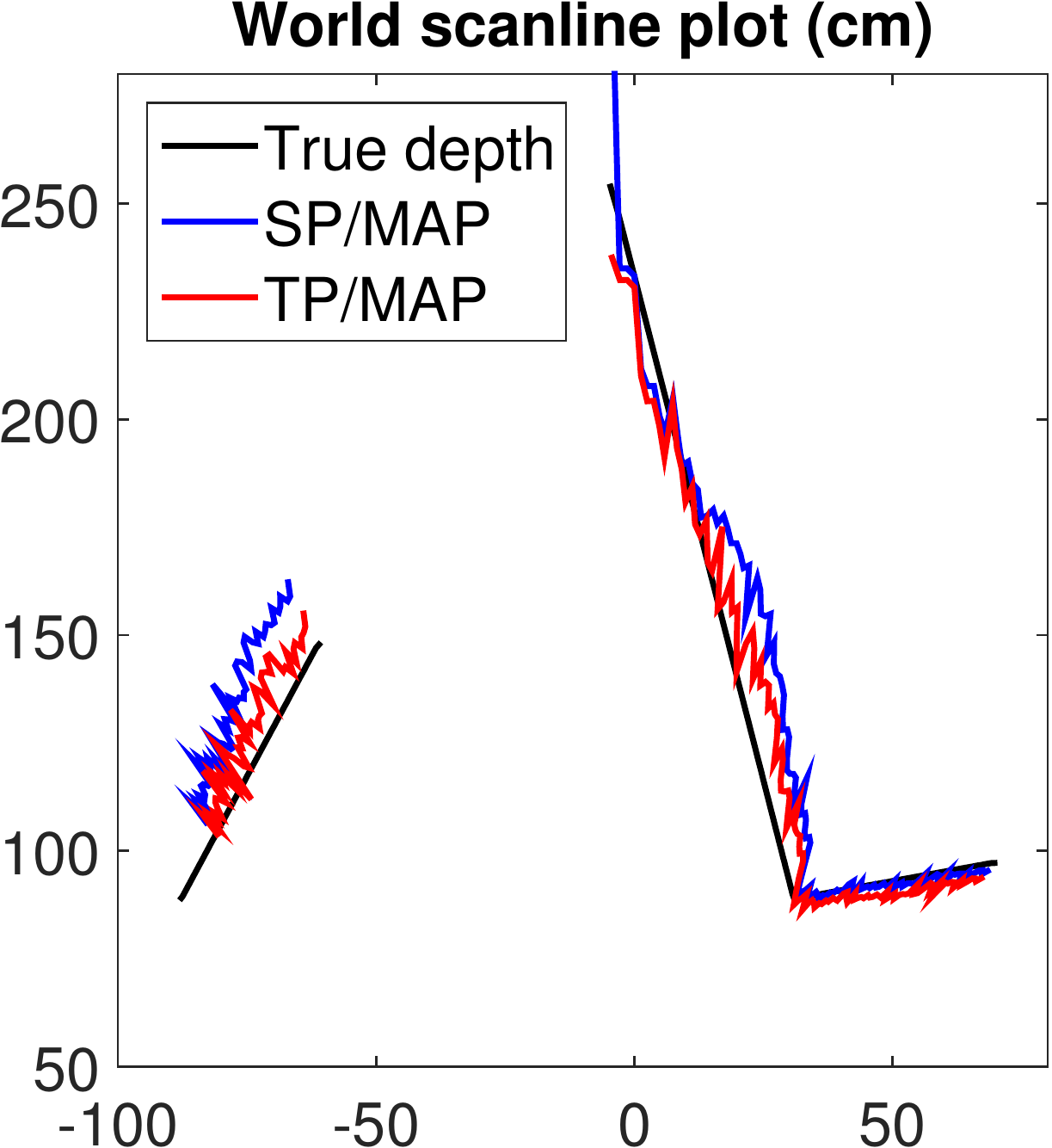}\label{fig:real-withmp-scanline}%
	}\hfill%
	\subfigure[SP err (real)]{%
		\includegraphics[width=0.138\linewidth]{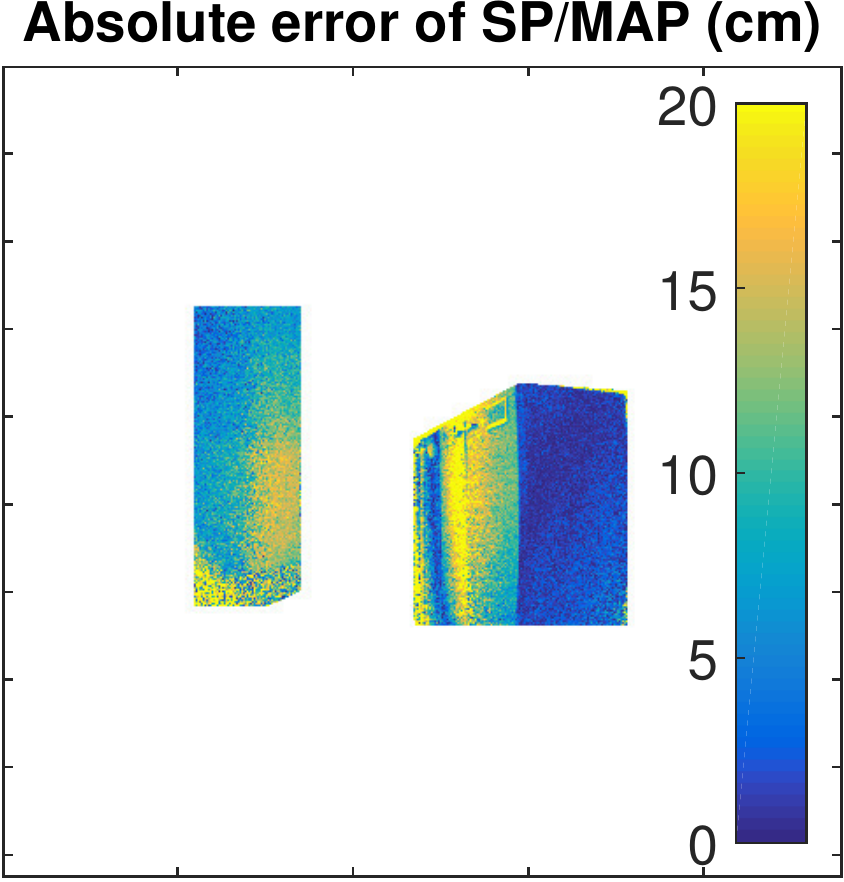}\label{fig:real-withmp-err-spmap}%
	}\hfill%
	\subfigure[SP err (syn)]{%
		\includegraphics[width=0.138\linewidth]{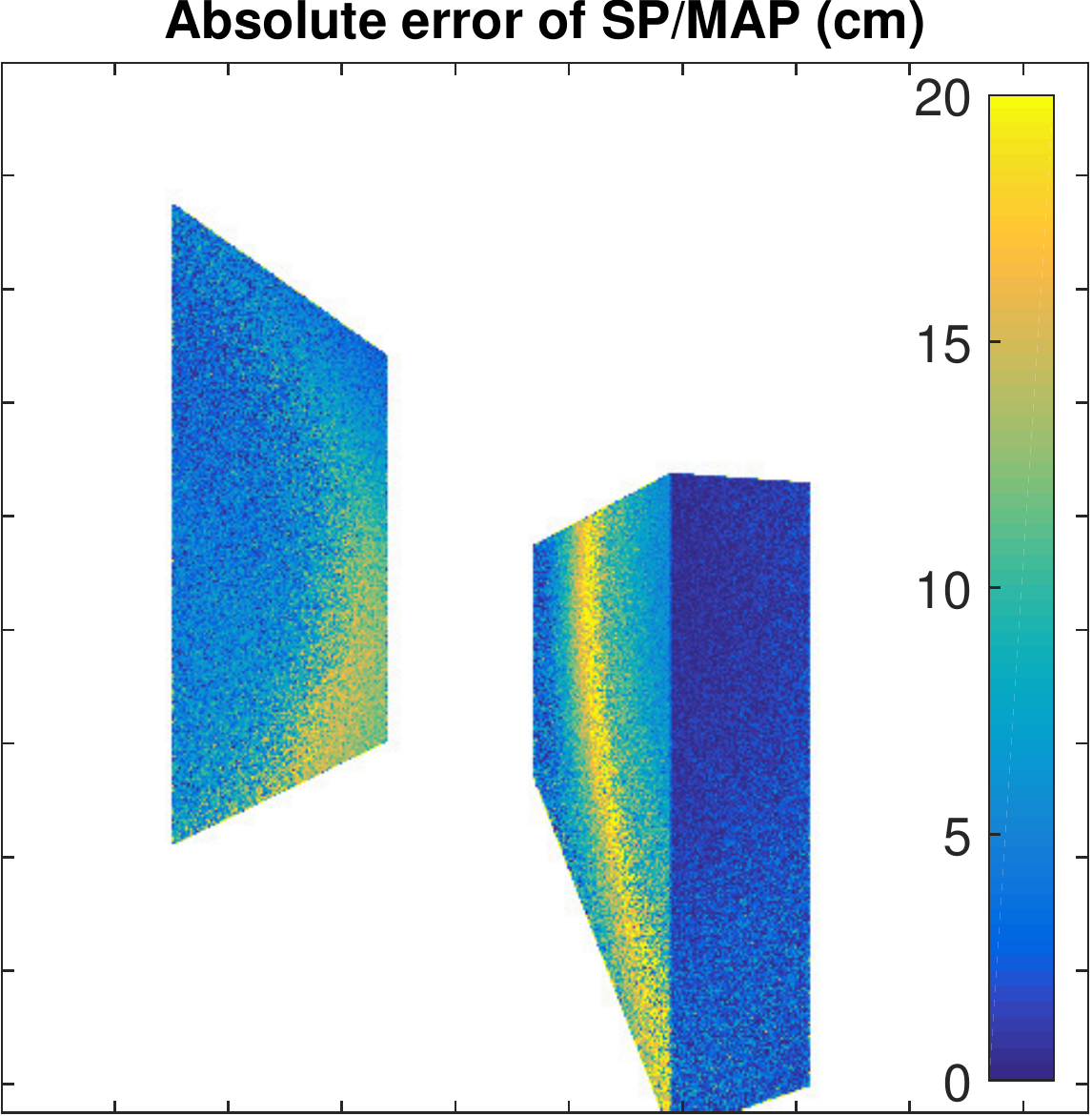}\label{fig:synth-withmp-err-spmap}%
	}\hfill%
	\subfigure[TP err (real)]{%
		\includegraphics[width=0.138\linewidth]{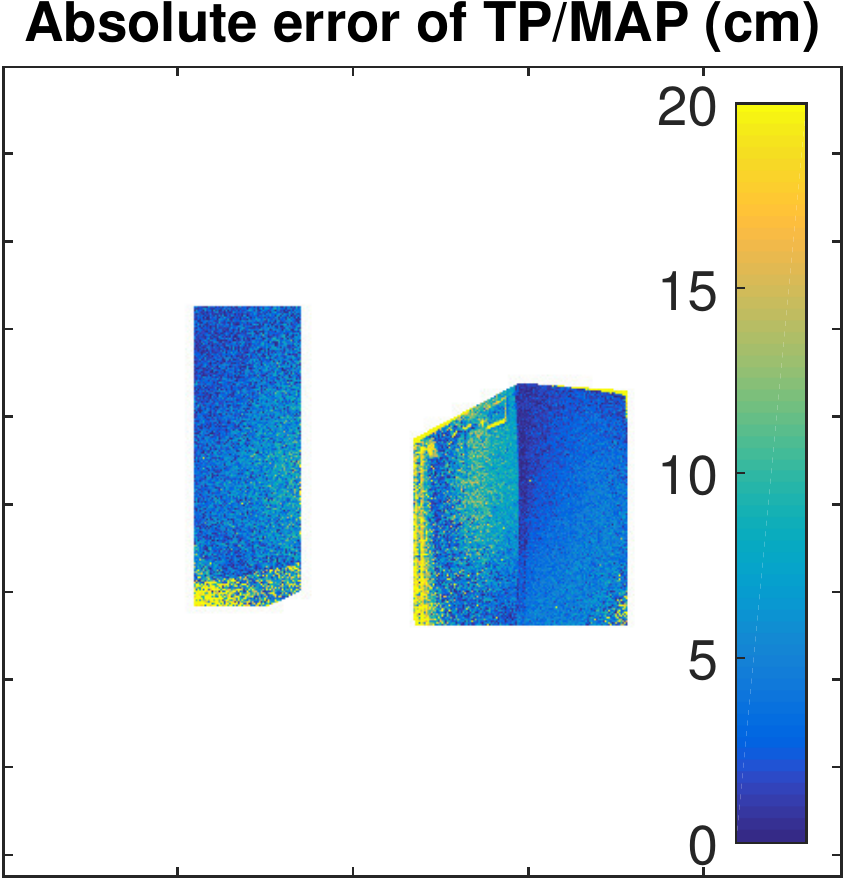}\label{fig:real-withmp-err-tpmap}%
	}\hfill%
	\subfigure[TP err (syn)]{%
		\includegraphics[width=0.138\linewidth]{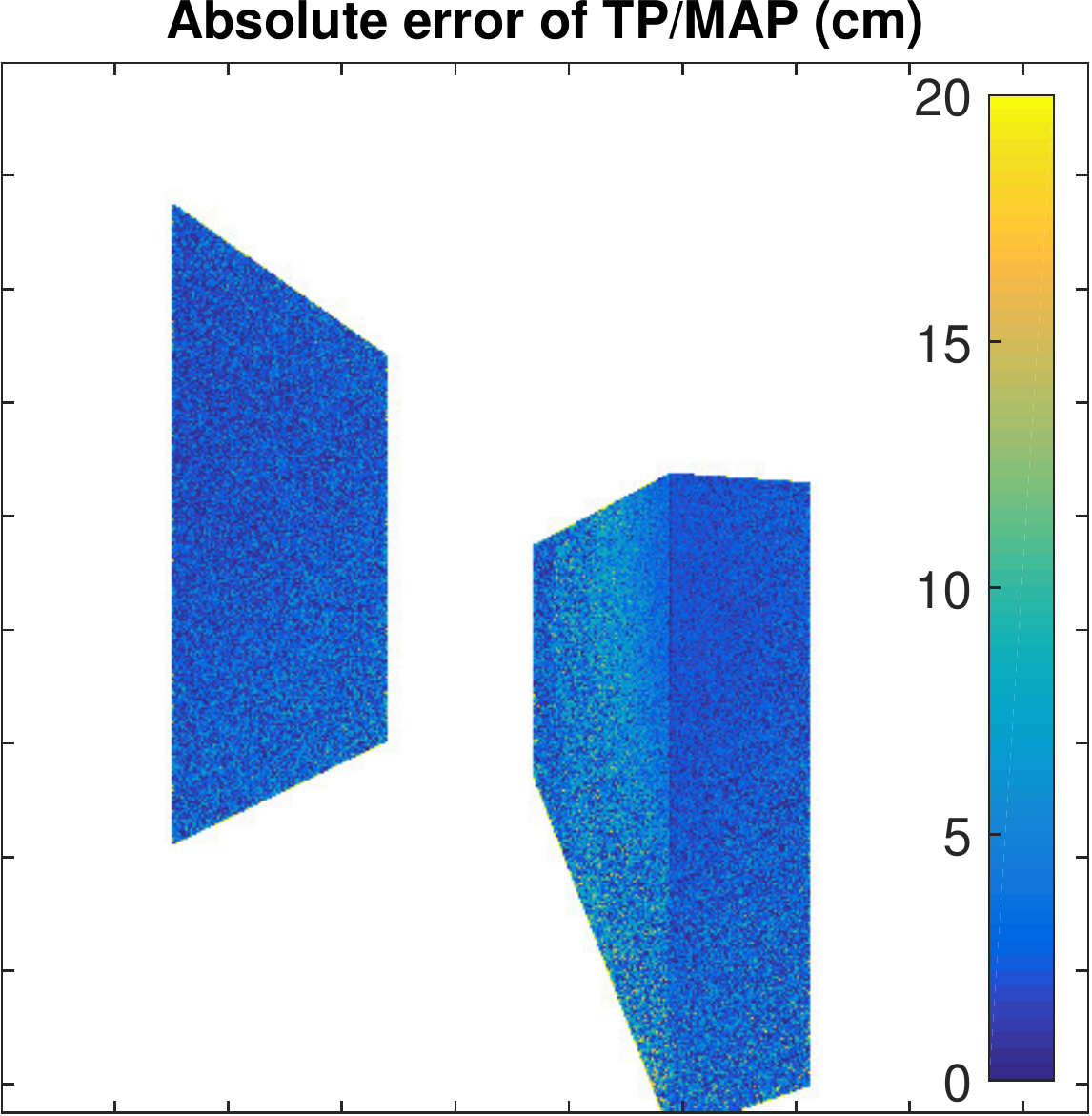}\label{fig:synth-withmp-err-tpmap}%
	}%
\vspace{-0.2cm}%
\caption{Validation of the accuracy of light transport simulation.
From the measured IR0 frames we use camera mapping to approximately
reconstruct the 3D scene geometry and surface properties manually.
The top row shows the no-multipath setting and we mask the
frame so that only the target object is shown.
The bottom row shows the multipath setting with a large white reflector added
to the scene (left side).
In the error results, each column corresponds to either real errors from the
measured IR frames or is entirely simulated.
Qualitatively there is an excellent agreement between real measurements and
synthetic simulations.
Small quantitative differences remain, for example in the no-multipath setting
with the two-path model.}
\label{fig:real-synth}%
\end{figure*}

To this end, we take images from two real scenes with a box and an optional
reflector, shown in Figure~\ref{fig:real-nomp-ir0} and~\ref{fig:real-withmp-ir0},
keeping the camera static between captures.
We then perform \emph{camera mapping} using the known camera intrinsics and
reconstruct a matching 3D scene for our simulator.  The synthetic scene allows
us to assess the agreement between qualitative effects in the real capture and
in the synthetic image.
In our comparison we mask the results to the area occupied by the box and
reflector, and in addition mask the bottom third of the sensor array because
this particular camera has no active illumination design in this region.

The top row in Figure~\ref{fig:real-synth} shows the scene with only the box,
the bottom row shows the multipath-corrupted scene with a large diffuse
reflector added.

The following important observations can be made: 1. Comparing each of the
four pairs of real and synthetic results the qualitative and quantitative
error agree between the actual recording and the simulation; 2. The multipath
corruption is clearly visible for the single-path (SP) model in Figure~\ref{fig:real-withmp-err-spmap}
and~\ref{fig:synth-withmp-err-spmap} and to a smaller extend in the
two-path (TP) models, Figure~\ref{fig:real-withmp-err-tpmap}
and~\ref{fig:synth-withmp-err-tpmap}.

Overall the simulation agrees very well with the real camera system.
We remark that beyond this single experiment we describe here, the simulator is
in daily use in our group and we have seen excellent agreement between
simulated results and live tests over many months of using it.

\subsection{Robust Model Invalidation}\label{sec:exp:invalidation}
To demonstrate the performance of the proposed invalidation mechanism we use
the same real experimental data as in the previous section.  We compute the
$\gamma$ score for both scenes and for the SP-Bayes and TP-Bayes models.

\begin{figure}[t!]%
	\centering%
	\subfigure[SP error]{%
		\includegraphics[width=0.245\linewidth]{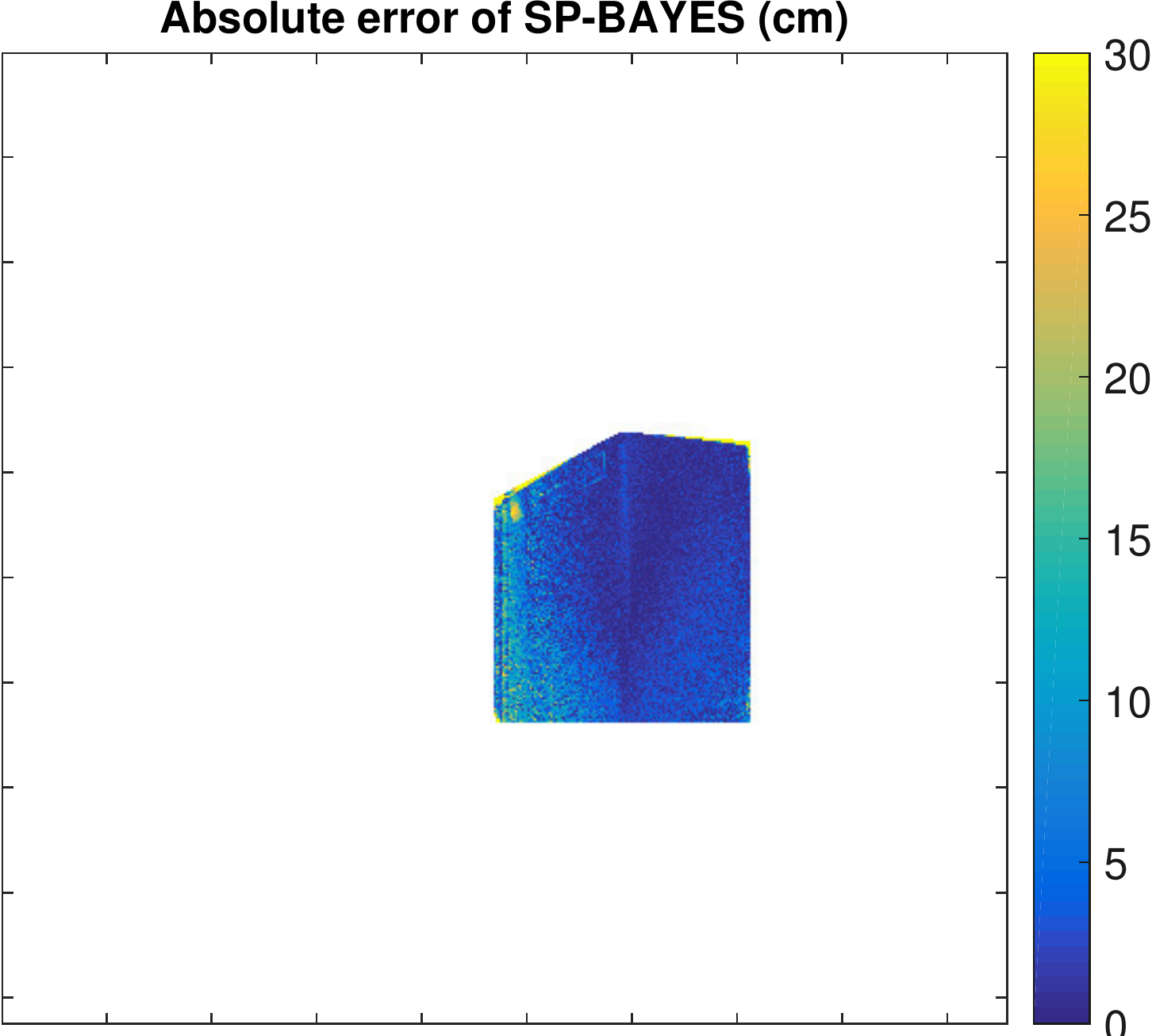}\label{fig:real-nomp-sp-terr}%
	}\hfill%
	\subfigure[SP $\gamma$]{%
		\includegraphics[width=0.245\linewidth]{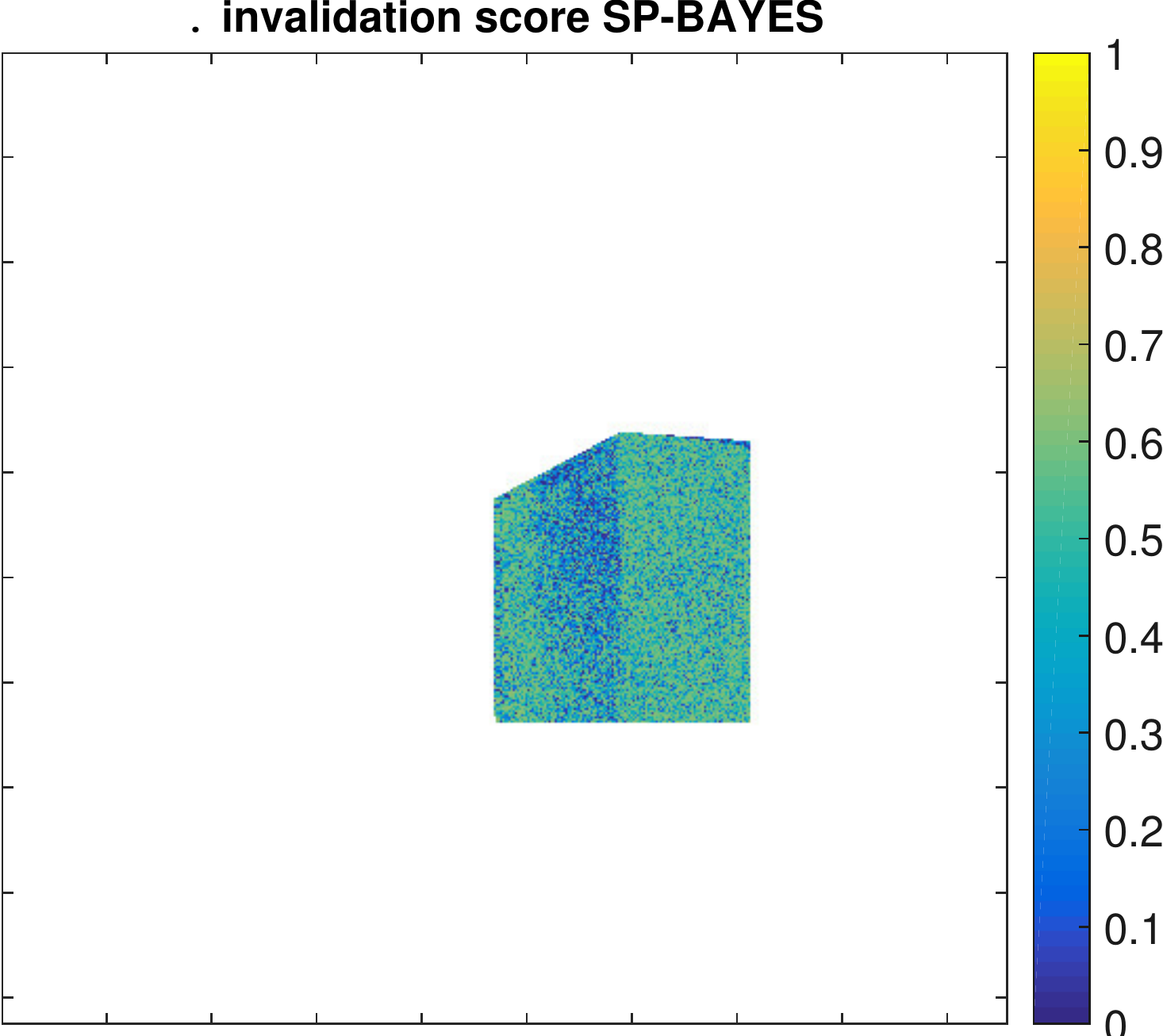}\label{fig:real-nomp-sp-gamma}%
	}\hfill%
	\subfigure[SP error]{%
		\includegraphics[width=0.245\linewidth]{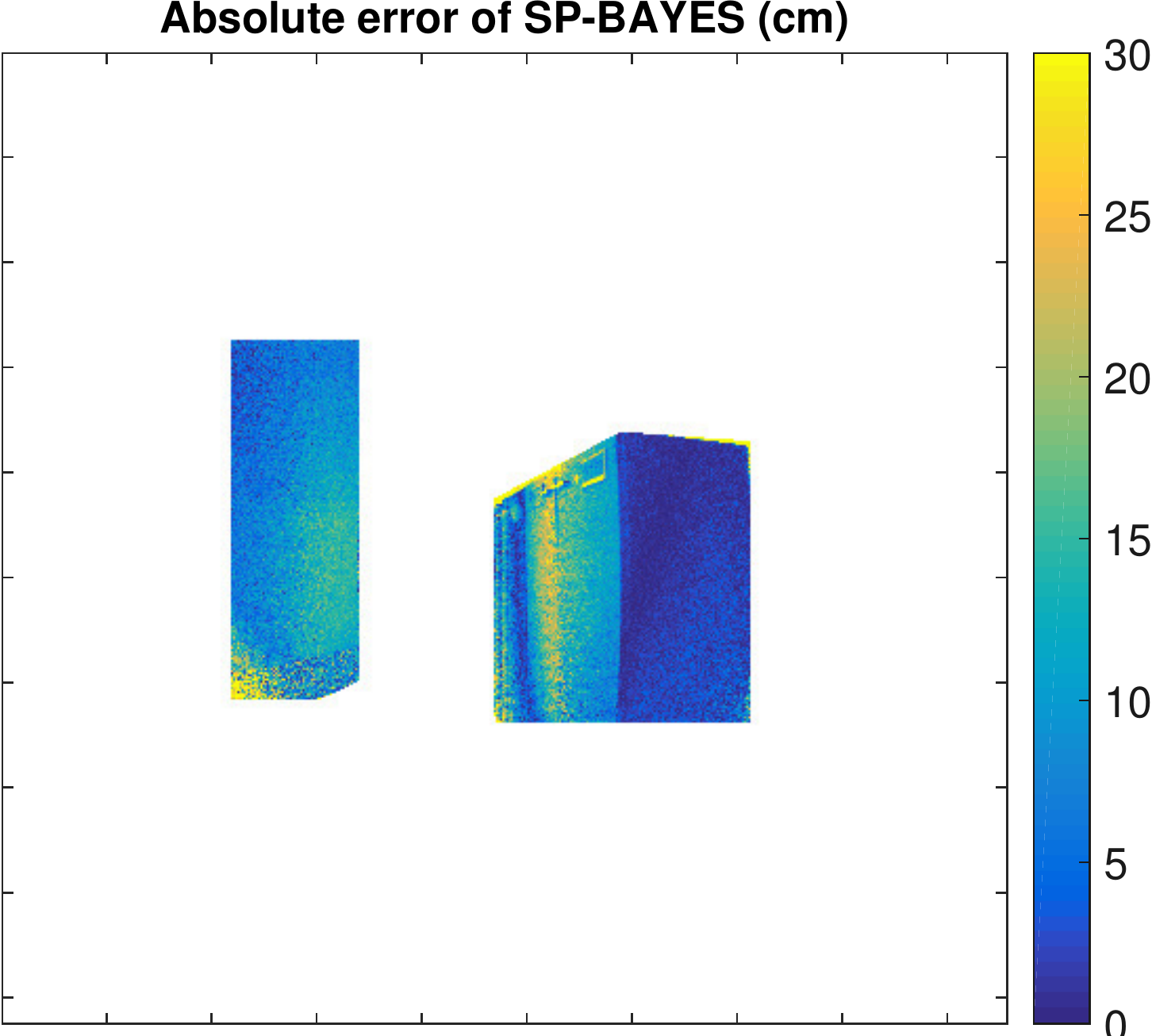}\label{fig:real-withmp-sp-terr}%
	}\hfill%
	\subfigure[SP $\gamma$]{%
		\includegraphics[width=0.245\linewidth]{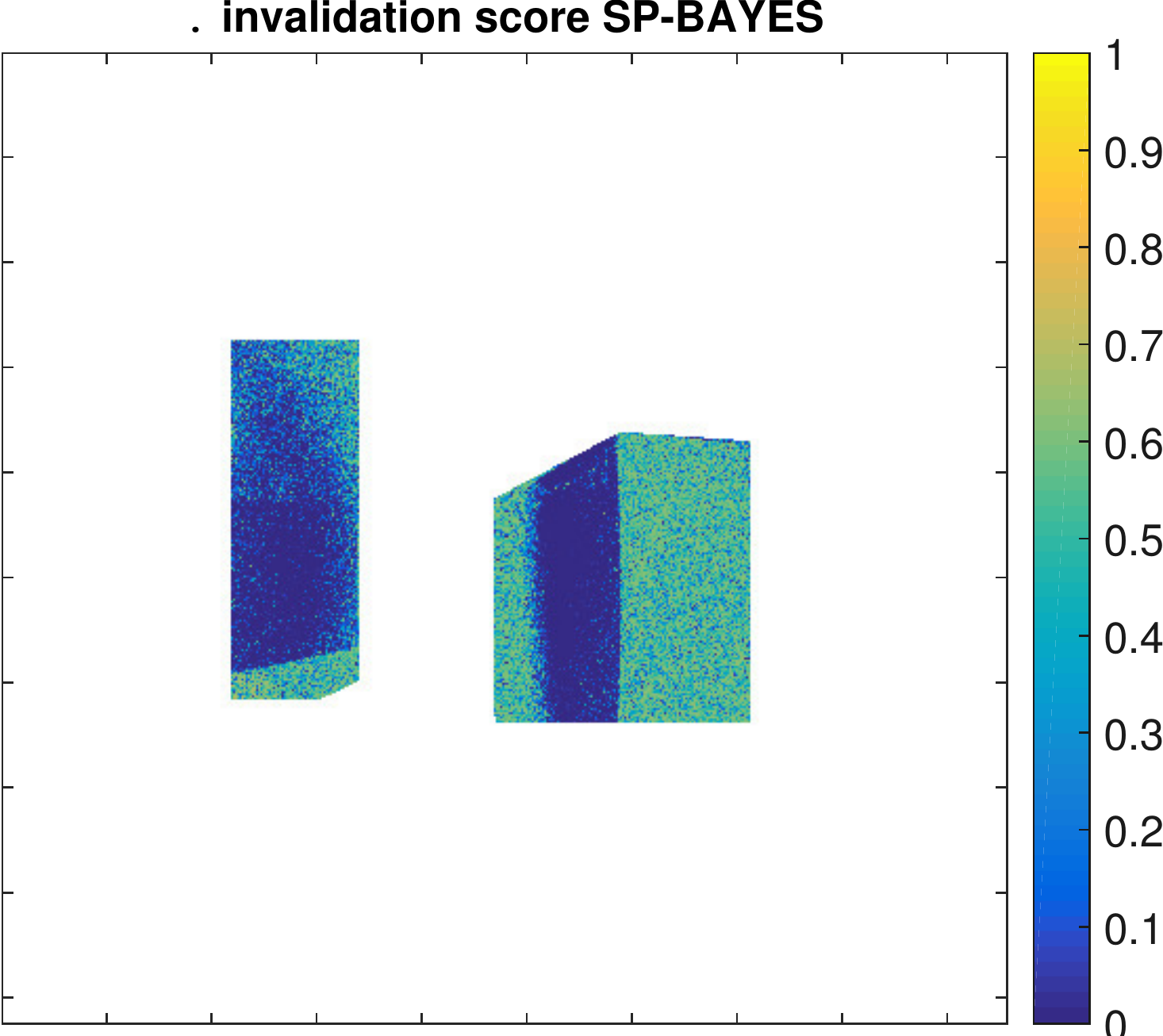}\label{fig:real-withmp-sp-gamma}%
	}\\%
	\subfigure[TP error]{%
		\includegraphics[width=0.245\linewidth]{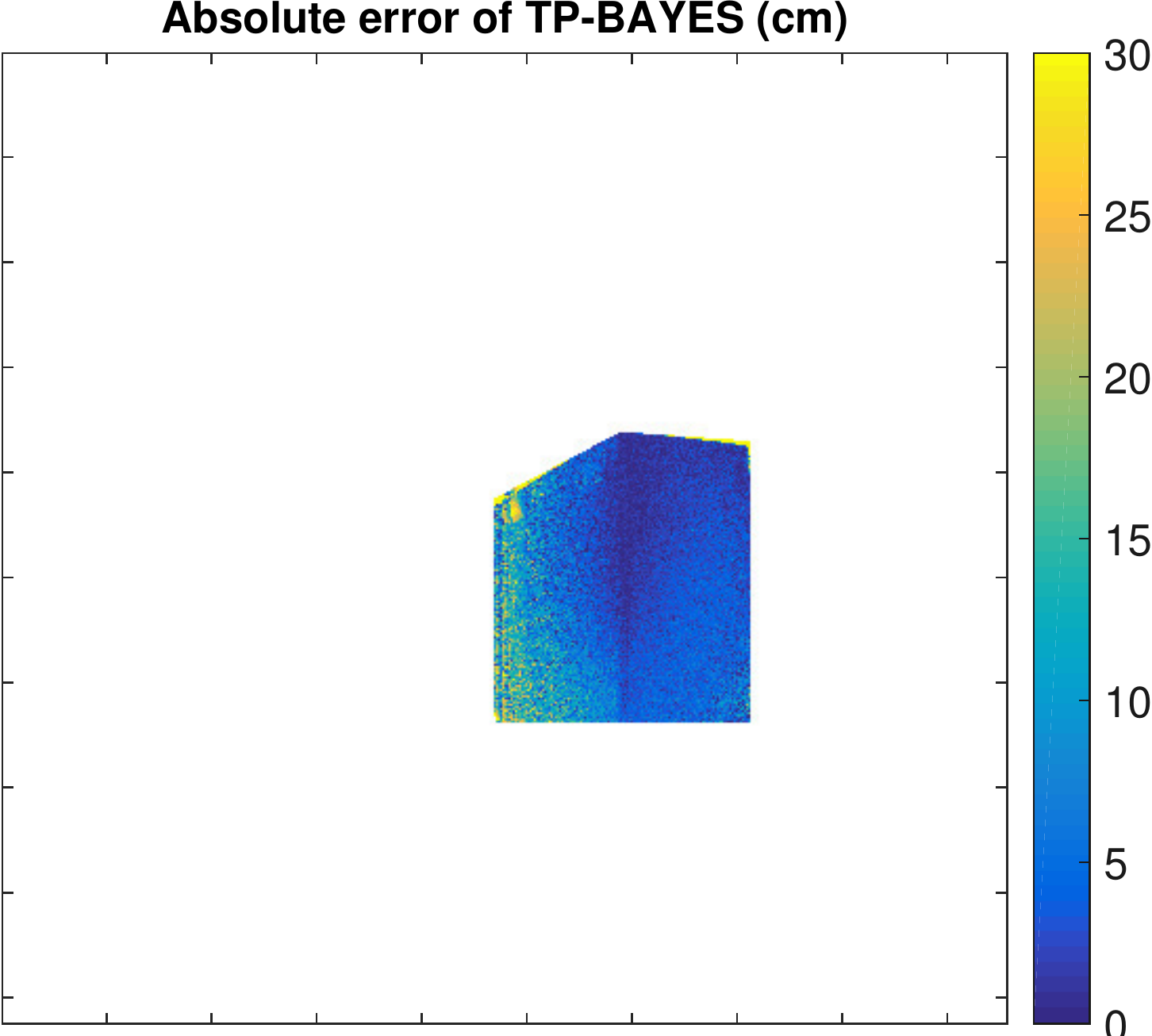}\label{fig:real-nomp-tp-terr}%
	}\hfill%
	\subfigure[TP $\gamma$]{%
		\includegraphics[width=0.245\linewidth]{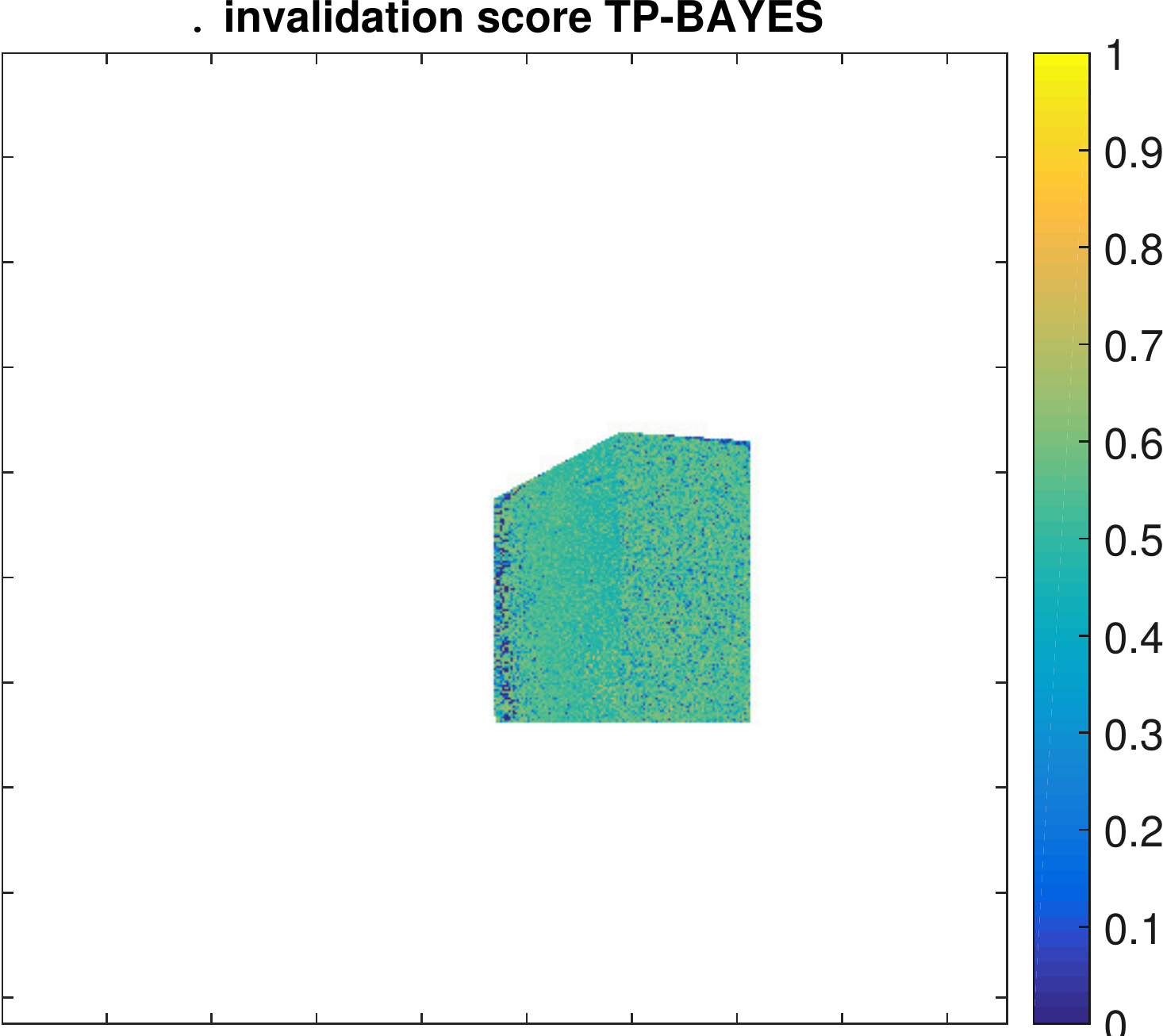}\label{fig:real-nomp-tp-gamma}%
	}\hfill%
	\subfigure[TP error]{%
		\includegraphics[width=0.245\linewidth]{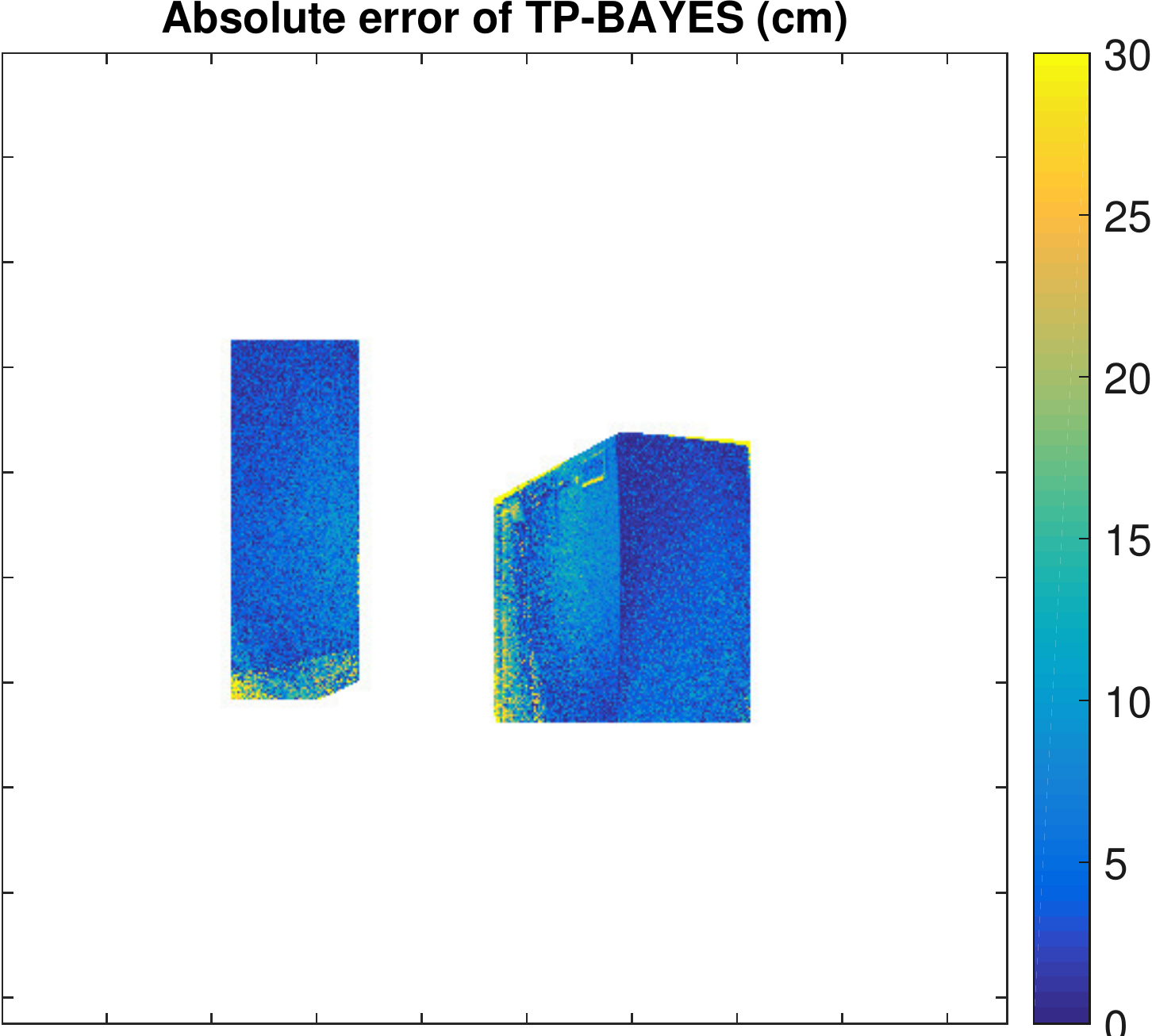}\label{fig:real-withmp-tp-terr}%
	}\hfill%
	\subfigure[TP $\gamma$]{%
		\includegraphics[width=0.245\linewidth]{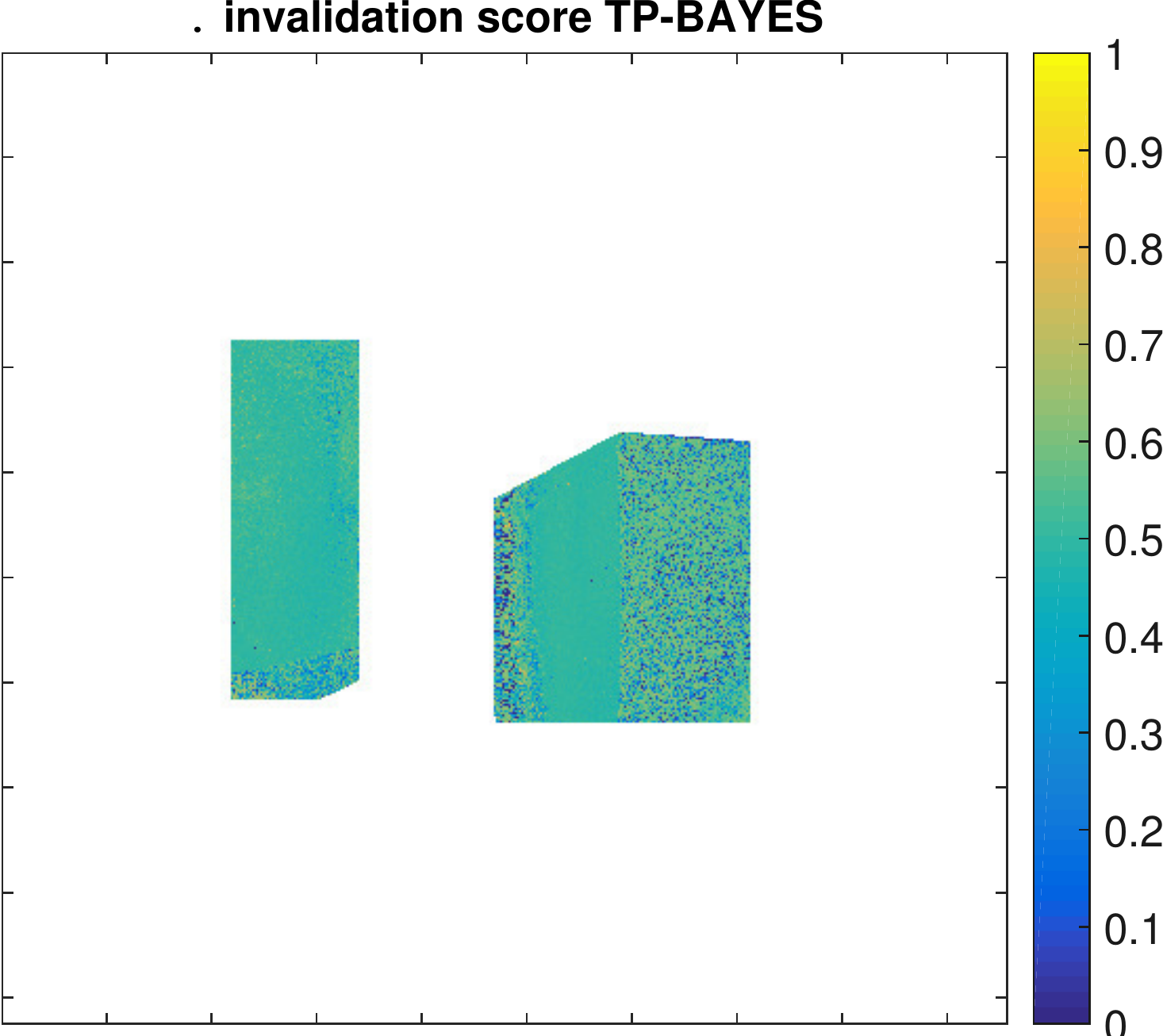}\label{fig:real-withmp-tp-gamma}%
	}%
\vspace{-0.2cm}%
\caption{Robust invalidation of measurements that violate our model
assumptions, as explained in Section~\ref{sec:gamma}.
In the top row we show the single path model SP-Bayes depth inference errors
on real data as with the previous experiment in Section~\ref{sec:real-synth},
whereas the bottom row shows the TP-Bayes model.
The $\gamma$ invalidation score robustly highlights
pixels affected by multipath in~\subref{fig:real-withmp-sp-gamma} (values
close to zero in dark blue) leading to large depth errors
in~\subref{fig:real-withmp-sp-terr}.
By thresholding the $\gamma$ score at a suitably chosen threshold these
measurements can be excluded from further processing.
Note that the two-path (TP) model handles multipath and the $\gamma$ score
in~\subref{fig:real-withmp-tp-gamma} does not invalidate affected pixels.}
\label{fig:real-invalidation}%
\end{figure}

The results are shown in Figure~\ref{fig:real-invalidation} and show that
robust invalidation is possible for this scene.
In the next section we also report $\gamma$ score images for simulation data.

\subsection{Benchmarking using Simulation Data}\label{sec:exp-sim}
In this final experiment we leverage the ability of our simulator to provide
ground truth depth.  This allows us to assess the depth inference performance
quantitatively.
We use five scenes adapted from blendswap.com for this purpose.  The depth
range in each of these scenes is within $50\textrm{cm}$ to $500\textrm{cm}$
and the scene surfaces represent a good variety in materials and convex and
concave geometries.

The results are visualized in Figures~\ref{fig:53420room}
to~\ref{fig:42851countrykitchen} and we report quantitative results for depth
reconstruction in Table~\ref{tab:sptp-sim}.  Two additional visualizations are
provided in the supplementary materials.
As error metric we use the 25/50/75 quantiles of absolute depth errors because
these approximately correspond to easy/medium/difficult surfaces.
We mask pixels in white for which no direct single-path response is created
during rendering.  These pixels typically correspond to either infinite rays
or to perfectly specular surfaces.  In both cases the time-of-flight operating
principle does not apply.

We again re-emphasize that the absolute magnitude of the errors is not material
and incomparable to other cameras for two reasons: first we report raw-depth errors
with absolutely no spatial or temporal filtering that is usually present. Second,
the jitter error highly depends on camera power, sensor size and other hardware
characteristics which are of no concern in this paper.

From the results we make the following observations:
\begin{enumerate}
\item Surfaces with low reflectivity have large depth errors but the model is
aware of this through a large inferred $\hat{\sigma}$ value.  For example, the
black floor in Figure~\ref{fig:53420room-spmap-terr}
and~\ref{fig:53420room-spmap-sigma}, or the cooking stove in
Figure~\ref{fig:42851countrykitchen-spmap-sigma}.
Improving on these regions would require increasing the light output or sensor
sensitivity.
\item Areas affected by strong multipath also have large depth errors; for
example the ceiling in Figure~\ref{fig:53420room-spmap-terr}.  The SP model
$\hat{\sigma}$ does not indicate a potential error, but the $\gamma$ score can
invalidate these observations, for example the ceiling in
Figure~\ref{fig:53420room-spmap-gamma}.
\item The two-path model (TP) improves depth accuracy in every scene but also
reports increased model $\hat{\sigma}$ compared to the single-path model.  For
example, compare the absolute errors between
Figures~\ref{fig:53420room-spmap-terr} and~\ref{fig:53420room-tpmap-terr}, and
the $\hat{\sigma}$ maps in Figure~\ref{fig:53420room-spmap-sigma}
and~\ref{fig:53420room-tpmap-sigma}.
\item Less invalidation happens in the two-path model.  In all scenes, the
TP-Bayes $\gamma$ invalidates less pixels compared to the SP-Bayes $\gamma$
map, because as a model it is a better representation for the physical
simulator.
\item In Table~\ref{tab:sptp-sim} the errors are significantly reduced by the
two-path model, typically by 40 percent.
\end{enumerate}

%
%

These results and insights agree with extensive live tests performed in the process
of productizing our system.

\begin{table}[t]
\centering
\resizebox{\columnwidth}{!}{%
\begin{tabular}{l|c|c|c|c}
      &       & \multicolumn{3}{c}{Absolute error quantile (cm)}\\
Scene & Model & $25\%$ & $50\%$ & $75\%$ \\
\hline
Sitting Room & SP-Bayes & 6.43 & 12.50 & 20.24\\
Fig.~\ref{fig:53420room} & TP-Bayes & 2.89 & 6.27 & 11.68\\
\hline
Breakfast Room & SP-Bayes & 3.09 & 5.69 & 10.39\\
(supp. mat.) & TP-Bayes & 1.60 & 3.40 & 6.55\\
\hline
Kitchen Nr 2 & SP-Bayes & 6.01 & 10.29 & 17.26\\
(supp. mat.) & TP-Bayes & 2.86 & 6.04 & 12.65\\
\hline
Country Kitchen & SP-Bayes & 4.69 & 9.63 & 17.20\\
Fig.~\ref{fig:42851countrykitchen} & TP-Bayes & 2.75 & 5.87 & 11.78\\
\hline
Wooden Staircase & SP-Bayes & 4.08 & 8.59 & 14.64\\
(supp. mat.) & TP-Bayes & 2.19 & 4.85 & 9.51\\
\hline
\end{tabular}%
}%
\caption{Predictive performance of the Bayesian single-path (SP) and two-path
(TP) models on realistic data obtained from physically-accurate light
transport simulation.  Across all scenes the 25/50/75 error quantiles are
significantly reduced by the two-path model. (raw-depth errors - no
spatial or temporal filtering
whatsoever)}
\label{tab:sptp-sim}
\end{table}


\newcommand{\maslulimresultfigure}[4][]{%
\begin{figure*}[t!]%
	\centering%
	\subfigure[Scene]{%
		\includegraphics[width=0.135\linewidth]{#2/#3/#3_rgb.png}\label{fig:#3-rgb}%
	}\hspace{0.55cm}
	\subfigure[Ground truth depth]{%
		\includegraphics[width=0.155\linewidth]{#2/#3/#3_gt.pdf}\label{fig:#3-gt}%
	}\hfill%
	\subfigure[SP-Bayes $\hat{t}$]{%
		\includegraphics[width=0.155\linewidth]{#2/#3/#3_sp-bayes_t.pdf}\label{fig:#3-spmap-t}%
	}\hfill%
	\subfigure[SP-Bayes error]{%
		\includegraphics[width=0.155\linewidth]{#2/#3/#3_sp-bayes_terr.pdf}\label{fig:#3-spmap-terr}%
	}\hfill%
	\subfigure[SP-Bayes $\hat{\sigma}$]{%
		\includegraphics[width=0.155\linewidth]{#2/#3/#3_sp-bayes_sigma.pdf}\label{fig:#3-spmap-sigma}%
	}\hfill%
	\subfigure[SP-Bayes $\gamma$]{%
		\includegraphics[width=0.155\linewidth]{#2/#3/#3_sp-bayes_gamma.pdf}\label{fig:#3-spmap-gamma}%
	}%
\\
\vspace{-0.2cm}%
	\subfigure[IR0]{%
		\includegraphics[width=0.155\linewidth]{#2/#3/#3_ir0.pdf}\label{fig:#3-ir0}%
	}\hfill%
	\subfigure[Multipath ratio]{%
		\includegraphics[width=0.155\linewidth]{#2/#3/#3_mpratio.pdf}\label{fig:#3-mpratio}%
	}\hfill%
	\subfigure[TP-Bayes $\hat{t}$]{%
		\includegraphics[width=0.155\linewidth]{#2/#3/#3_tp-bayes_t.pdf}\label{fig:#3-tpmap-t}%
	}\hfill%
	\subfigure[TP-Bayes error]{%
		\includegraphics[width=0.155\linewidth]{#2/#3/#3_tp-bayes_terr.pdf}\label{fig:#3-tpmap-terr}%
	}\hfill%
	\subfigure[TP-Bayes $\hat{\sigma}$]{%
		\includegraphics[width=0.155\linewidth]{#2/#3/#3_tp-bayes_sigma.pdf}\label{fig:#3-tpmap-sigma}%
	}\hfill%
	\subfigure[TP-Bayes $\gamma$]{%
		\includegraphics[width=0.155\linewidth]{#2/#3/#3_tp-bayes_gamma.pdf}\label{fig:#3-tpmap-gamma}%
	}%
\vspace{-0.2cm}%
\caption{Rendered simulation (scene adapted from #4, licensed CC-BY from
blendswap.com). #1}%
\label{fig:#3}%
\end{figure*}%
}%

\maslulimresultfigure[High errors are present due to low reflectivity surfaces
and multipath; the multipath errors are reduced by the two-path model
(ceiling, wall, floor).
The uncertainty estimate $\hat{\sigma}$ is higher for the two-path model,
reflecting the multipath awareness (compare the $\hat{\sigma}$ values at the
ceiling).]{figures/exp-maslulim}{53420room}{``sitting room'' by cenobi}
\maslulimresultfigure[Overall strong multipath error reduction across the
scene but higher overall single-frame jitter due to the strong ambient
lighting.]{figures/exp-maslulim}{42851countrykitchen}{``Country-Kitchen
Cycles'' by Jay-Artist}

\section{Conclusion}
Our presented approach is based on sound probabilistic modelling given our
understanding of the physical reality.  Bayesian inference naturally provides
a powerful formal calculus to perform depth inference given our modelling
assumptions.
We have shown that even a simple model of multipath enables significant
reductions in the depth error.
However, both parts of our approach---the \emph{prior} and \emph{model}---are
general and open to future extensions.
For the prior we plan to develop scene- and task-specific priors to be able to
improve performance in the presence of strong multipath and ambient light.
We envision more refined models of multipath, for example by
replacing the two-path pulse response by a more accurate analytic model of
diffuse Lambertian multipath.   This would require adding further latent
variables related to multipath responses and creating suitable priors for
them; this may be challenging but our simulation framework will likely enable
us to make progress in this direction in the future.
Our statistical view on time-of-flight enables all these extensions within a
principled framework.


\ifCLASSOPTIONcompsoc
  \section*{Acknowledgments}
\else
  \section*{Acknowledgment}
\fi
\small

We thank Michael Baltaxe, Yair Sharf, Sahar Vilan and Giora Yahav for their
contributions, long term support and commitment to this work.

Parts of this work have been done when Christoph Dann interned at Microsoft
Research Cambridge, UK.

\ifCLASSOPTIONcaptionsoff
  \newpage
\fi



{
\small
\bibliographystyle{IEEEtran}
\bibliography{BayesianTOF_arxiv}
}

%

\begin{IEEEbiography}[{\includegraphics[width=1in,clip,keepaspectratio,trim=0pt 0pt 0pt 0pt]{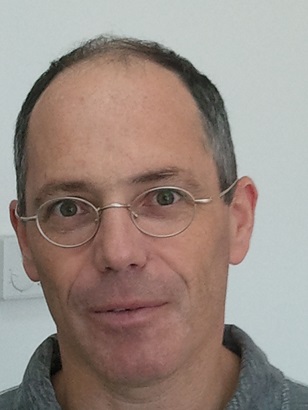}}]{Amit Adam}
Amit Adam received the PhD degree from the Technion-Israel Institute of
Technology in 2001 for a thesis on vision-based navigation.
Since his graduation, he has been working as an applied computer vision researcher in various
application areas such as medical navigation, video surveillance, and recognition.
Joining  Microsoft's Advanced Imaging Technologies Group (AIT) in 2012, he has
since worked on computational problems related to
time-of-flight depth cameras.
\end{IEEEbiography}

\begin{IEEEbiography}[{\includegraphics[width=1in,clip,keepaspectratio,trim=0pt 0pt 0pt 0pt]{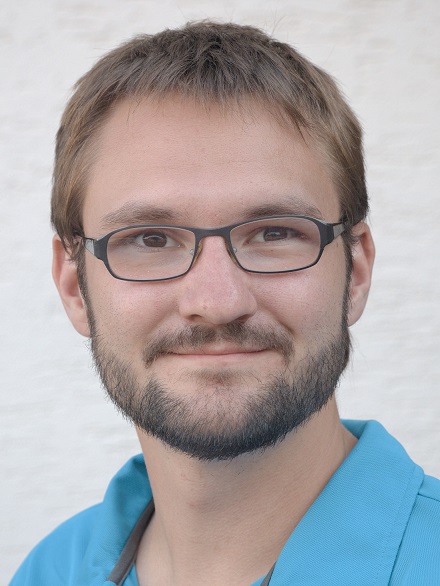}}]{Christoph Dann}
Christoph Dann obtained his B.Sc. and M.Sc. degree in Computer Science from
the Technical University of Darmstadt, Germany, in 2011 and 2014,
respectively. He is currently working toward a PhD degree in the Machine
Learning Department at Carnegie Mellon University, USA.  In the past,
Christoph worked as an undergraduate researcher at the Max-Planck Institute
for Informatics, the Intelligent Autonomous Systems group at the Technical
University of Darmstadt, the Aerospace Controls Laboratory at MIT and as a
research intern at Microsoft Research, Cambridge, UK.  His research primarily
focuses on sequential decision making under uncertainty including
reinforcement learning as well as applications in computer vision.
\end{IEEEbiography}

\begin{IEEEbiography}[{\includegraphics[width=1in,clip,keepaspectratio,trim=0pt 0pt 0pt 0pt]{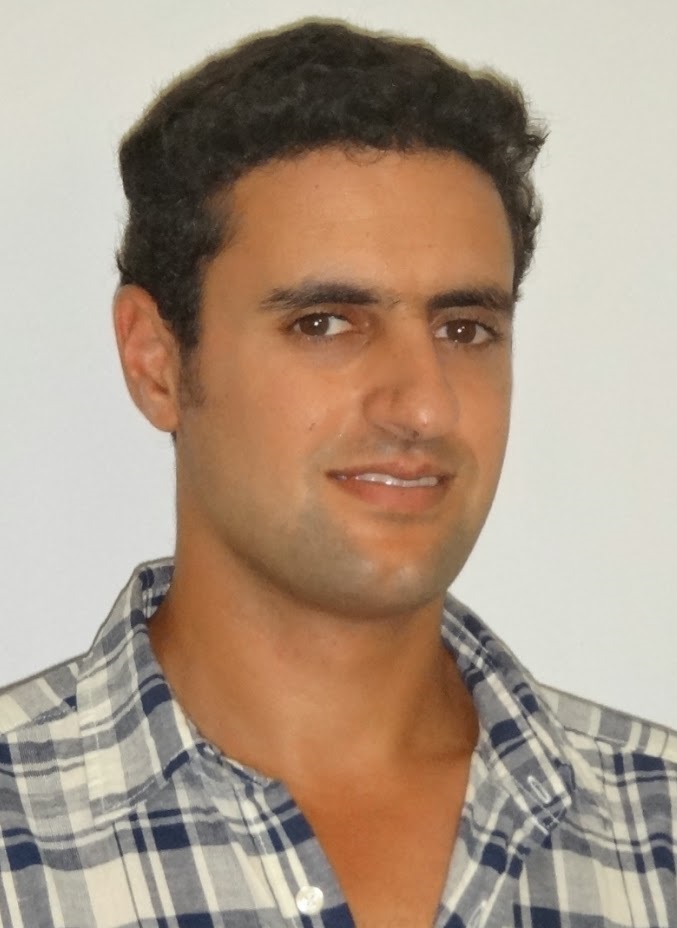}}]{Omer Yair}
received the BSc degree in Electrical Engineering (summa cum laude) and a BSc
in Physics (summa cum laude) from the Technion-Israel Institute of Technology
in 2011. He is currently with the Advance Imaging Technology Group at
Microsoft and pursuing a MSc degree in Physics at the Technion.
\end{IEEEbiography}

\begin{IEEEbiography}[{\includegraphics[width=1in,clip, keepaspectratio, trim=0pt 0pt 0pt 0pt]{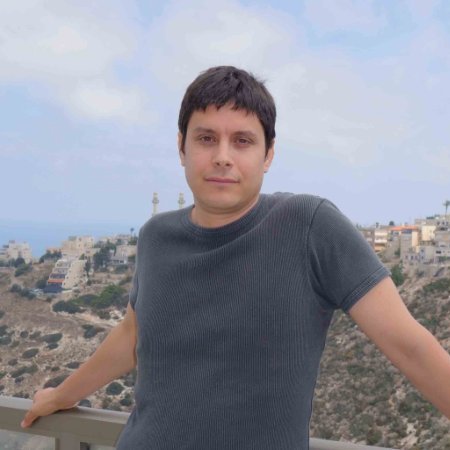}}]{Shai Mazor}
Shai received the B.Sc. and M.Sc. degrees in Electrical Engineering from the Technion-Israel Institute of Technology.
After graduating he worked as a developer and later program manager, gaining experience both in startup
companies and large corporations. He joined Microsoft's Advanced Imaging Technologies Group (AIT) in 2013, where he
is now a senior program manager responsible for incubation of new applications for AIT technology.
In addition to his engineering education, Shai holds an MBA from the IDC, where he was also an exchange
student at the Wharton Business School.

\end{IEEEbiography}

\begin{IEEEbiography}[{\includegraphics[width=1in,clip,keepaspectratio,trim=0pt 0pt 0pt 0pt]{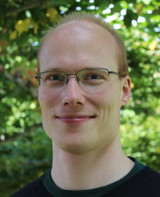}}]{Sebastian Nowozin}
is a researcher in the Machine Learning and Perception group at Microsoft
Research Cambridge.
He received his Master of Engineering degree from the Shanghai Jiaotong
University (SJTU) and his diploma degree in computer science with distinction
from the Technical University of Berlin in 2006.
He received his PhD degree summa cum laude in 2009 for his thesis on learning
with structured data in computer vision, completed at the Max Planck Institute
for Biological Cybernetics, T{\"u}bingen and the Technical University of
Berlin.
His research interest is at the intersection of computer vision and machine
learning.
%
He is associate editor for TPAMI, IJCV, and JMLR and regularly serves as
PC-member and reviewer for machine learning (NIPS, ICML, AISTATS, UAI, ECML,
JMLR) and computer vision (CVPR, ICCV, ECCV, PAMI, IJCV) conferences and
journals.
\end{IEEEbiography}

\vfill


\cleardoublepage


\begin{appendices}
\section{Video Demonstration}

We submit a short video showing the robustness of our approach. In the video
we show robust live (real-time) inference of depth, reflectance/albedo and
illumination. We demonstrate the effective separation between illumination and
reflectance by waving a powerful light projector and noting that the albedo
and depth map stay invariant to the changing illumination conditions.

Please note that the RGB stream was shot by the person holding the light
projector - and is unrelated to the depth camera stream (we added it for
general impression of the scene - it is slightly confusing).


\section{Inference Details}
For this section we will use the compound parameter vector
$\vec \theta = [t, \rho, \lambda]^T$, or
$\vec \theta = [t, \rho, \lambda, t_2, \rho_2]^T$
for the single- and two-paths model.
This unifies the notation for all unknown imaging conditions we would like to
infer.

The response curve function $\vec C(t)$ appearing in the expression
for the mean photon response $\vec\mu$ (see equation~(\ref{generative_mean}) in the main paper),
is obtained
from calibrated measurements of the actual camera, and then approximated by
Chebyshev polynomials of degree sixteen~\cite{trefethen2012approximation}.
Because the curves are smooth the Chebyshev approximation is compact yet very
accurate and evaluation of $\vec{C}(t)$ also provides the derivatives
$\frac{\partial}{\partial t} \vec{C}(t)$ and
$\frac{\partial}{\partial^2 t} \vec{C}(t)$ for no additional computational
cost.

\subsection{Maximum Likelihood Estimation (MLE)}
The standard maximum likelihood estimate are the imaging conditions $t$,
$\rho$, $\lambda$ which maximize the likelihood or equivalently minimize the
negative log-likelihood
{\small
\begin{eqnarray}
\!\!\!	& \!\!\underset{\vec \theta}{\argmin} & -\log P(\vec R | \vec \theta)\label{eqn:mle}\\
=\!\!\!	& \!\!\underset{\vec \theta}{\argmin}
		& \!\!\sum_{i=1}^n \left[\frac {(R_i - \mu_i(\vec \theta))^2}{2 (\alpha \mu_i(\vec \theta) + K)}
			\!+\! \frac{1}{2} \log(\alpha \mu_i(\vec \theta) + K )\right].\nonumber
\end{eqnarray}
}

With this Chebyshev polynomial approximation we can also compute derivatives with
respect to $\vec \theta$
of the log-likelihood function, and the entire
log-likelihood function becomes smooth and twice differentiable.

Solving the three-dimensional minimization problem in Equation~\eqref{eqn:mle}
with standard Quasi-Newton methods such as L-BFGS~\cite{liu1989lbfgs} is
possible but often yields unreasonable result if we do not constrain the
parameters.
For example, negative values of $\rho$ might have the lowest function value
but are physically impossible.
Another issue is that the response curves $\vec C$ are measured only within a
reasonable range.  Outside of this range, the Chebyshev approximations have
arbitrary behavior which leads to implausible  solutions.

We therefore constrain the range of parameters using log-barrier terms
{\small
\begin{align}
\underset{\vec \theta}{\argmin} \:
	& \:\sum_{i=1}^n \left[\frac {(R_i - \mu_i(\vec \theta))^2}{2 (\alpha \mu_i(\vec \theta) + K)}
		+ \frac{1}{2} \log(  \alpha \mu_i(\vec \theta) + K )\right]\\
& + \sum_j b \: (\log(\theta_j - \theta_{j,\min}) + \log(\theta_{j,\max} - \theta_j)).
\label{eqn:mle_reg}
\end{align}
}
The scalar $b = 10^{-2}$ is a \emph{barrier} coefficient and
$\vec \theta_{min}$, $\vec \theta_{max}$ are the smallest and largest values of
each parameter we want to consider.
The problem remains twice differentiable and quasi-Newton methods can be
applied for finding local minima reliably because any local optima has to
occur within the relative interior of the rectangle described by
$\theta_{j,\min}$ and $\theta_{j,\max}$.

To find the global optimum, we restart the quasi-Newton method ten times with
initialization sampled uniformly in the parameter ranges.
For producing labeled training data this is more than sufficient.
Even during exposure profile optimization,
experiments on good and mediocre shutter designs
have shown that after $10$ restarts in 97\% of the cases the same global
solution was found as with $100$ restarts.

\subsection{Maximum A-Posteriori Estimation (MAP)}\label{sec:map}
The maximum a-posteriori (MAP) estimate is similar to the maximum likelihood
estimate but also considers the prior instead of only the likelihood
distribution.
We determine the estimate by minimizing the negative log posterior
{\small
\begin{eqnarray}
\!\!\!\!\!	& \underset{\vec \theta}{\argmin} &  \!\!\!-\log P(\vec \theta | \vec R)\label{eqn:map}\\
=\!\!\!\!\!	& \underset{\vec \theta}{\argmin}
		& \!\!\!\sum_{i=1}^n \left[\frac{(R_i - \mu_i(\vec \theta))^2}{2 (\alpha \mu_i(\vec \theta) + K)}
			+ \frac{1}{2} \log(\alpha \mu_i(\vec \theta) + K )\right]\nonumber\\
	& & \!\!\! - \log p(\vec \theta).\nonumber
\label{eqn:map}
\end{eqnarray}
}
Due to the particular choices of twice differentiable prior distributions, we
can solve this problem right away with quasi-Newton methods.
The log-barrier terms used for the maximum likelihood estimate are now
implicitly defined in the prior.
In fact, the constrained maximum likelihood estimate in the previous
subsection can be understood as MAP estimate with an approximately uniform
prior on the ranges $\vec \theta_{min}$ to $\vec \theta_{max}$.

The advantage of the MAP estimate is that
when prior knowledge exists - for example
a strong belief on the ambient light intensity - then
we may incorporate it.
In contrast, the MLE does not encode any preference for certain parameter
values.

\subsection{Bayesian Posterior Inference}
The Bayesian point estimate is motivated by statistical decision
theory~\cite{berger1985statisticaldecisiontheory,robert2001bayesianchoice}.
The Bayes estimator yields the lowest expected error.
Assuming the squared loss function, the estimator is characterized as
\begin{equation}
\hat{\theta}_{\textrm{Bayes}}(\vec{R}) := \underset{\vec \theta}{\argmin} \:
	\E_{\tilde{\theta} \sim P(\tilde{\theta}|\vec{R})}[\| \vec \theta - \tilde{\theta}\|_2^2],
\end{equation}
where $\tilde{\theta}$ are the true but uncertain parameters.

This decision problem has a closed form solution: namely the mean parameters under
the marginal posterior distributions.  Because the squared loss decomposes
over parameters, so does the decision problem.

For example, the Bayes estimator $\hat{t}_{\textrm{Bayes}}$ for depth is given
by
\begin{equation}
	\hat{t}_{\textrm{Bayes}}(\vec{R}) = \E[ t | \vec R] = \int t \: p(t | \vec R) \,\textrm{d}t.
\end{equation}
The marginal posterior distribution $p(t | \vec R)$ can be written in terms of
the joint distribution as
\begin{eqnarray}
p(t | \vec R)
	& = & \int p(\vec \theta | \vec R) \,\textrm{d}\vec \rho \,\textrm{d}\lambda\nonumber\\
	& = & \int \frac{ p(\vec R | \vec \theta) p(\vec \theta)}{p(\vec R)}
		\,\textrm{d}\rho \,\textrm{d}\lambda.\nonumber
\end{eqnarray}
The Bayes estimator $\hat{t}_{\textrm{Bayes}}$ is therefore equal to
\begin{equation}
	\mathbb E[ t | \vec R] =
		\frac{\int t \: p(\vec \theta) p(\vec R | \vec \theta) \,\textrm{d}\vec \theta}{
			\int p(\vec \theta) p(\vec R | \vec \theta) \,\textrm{d}\vec \theta}.
	\label{eqn:bayes_est}
\end{equation}

One way of computing the Bayes estimator is solving the integrals in the
numerator and denominator for all parameters that we are interested in.
We use a state-of-the-art numerical quadrature method~\cite{genz1980remarks}
for vector-valued integrals over rectangular regions.

\begin{figure}
\centering\includegraphics[width=0.99\linewidth]{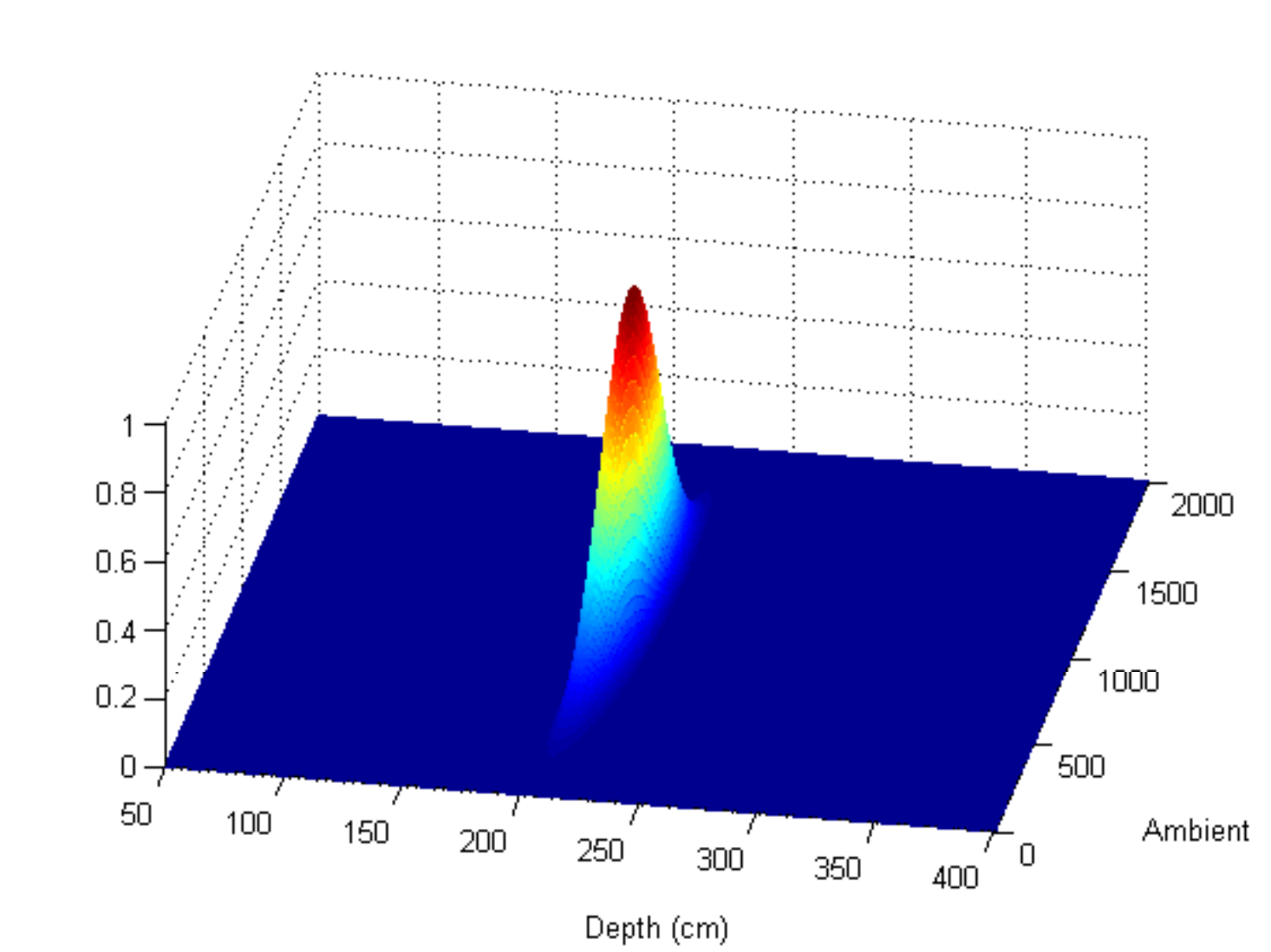}
\caption{Slice of the posterior distribution of the single-path model for
fixed values of reflectivity $\rho$.
In this case the posterior distribution of the single-path model is dominated
by a single Gaussian-like mode.}
\label{fig:post_slice}
\end{figure}
However, the numerical quadrature approach is very slow and has numerical
issues that yield sub-optimal solutions.
We therefore consider an alternative way to compute the Bayes estimators:
Monte Carlo using importance sampling~\cite{rubinstein2011simulation}.

We observed that the posterior distributions of the single-path model are
mostly dominated by a few important modes that often have symmetric shape, see
Figure~\ref{fig:post_slice}.
The posterior can therefore be approximated well by a mixture of Gaussians.
Using importance sampling with a mixture of Gaussians proposal distribution
should therefore yield fast convergence to the true Bayes estimator.

The proposal distribution is a mixture of $k$ Gaussians placed at the outputs
of $k$ local optima of the MAP problem obtained as described in
Section~(\ref{sec:map}).
The proposal distribution is
\begin{eqnarray}
q(\vec \theta) & \propto
	& \sum_{i = 1}^k p(\vec \theta^{(i)}) \: p(\vec R | \vec \theta^{(i)})
		\:\mathcal N(\vec \theta  | \vec \theta^{(i)}, H^{(i)}),
\end{eqnarray}
where $k$ is the number of mixture components used and $\vec \theta^{(k)}$ are
the locations of these mixtures.
For the covariance matrices $H^{(k)}$ we use the inverse Hessian of the
negative log-posterior (as in a Laplace approximation).
Due to the particular choice of twice differentiable priors, the Hessian of
the log-posterior are always positive definite in local optima.

We generate samples $\vec \eta_1 \dots \vec \eta_m$ from $q$ and re-weight
each sample by $w_i = \frac{p(\vec \eta_i)}{q(\vec \eta_i)}$ to account for
the errors in the approximation of the posterior by $q$.
These samples are then used to obtain Monte-Carlo estimates of the integrals
in equation~\eqref{eqn:bayes_est}.

We determine the number of samples required to approximate the integrals by
the \emph{effective sample size} (ESS)~\cite{liu1996metropolized,kong1992ess},
\begin{equation}
	\frac{\left(\sum_{i=1}^m w_i\right)^2}{\sum_{i=1}^m w_i^2}.
\end{equation}
We stop sampling as soon as the ESS exceeds a threshold (usually in the order
of $50-200$). In most cases this threshold is reached with a small number of
actual samples.
Empirically we observe the importance sampling approach to be much faster and
robust in practice than the numerical quadrature approach.

\section{Test-time Regression Tree Inference Details}
The regression trees approach we described in the main paper has an advantage in
terms of flexibility.
We used this flexibility to solve several issues that we encountered during
development of the prototype camera.
While a full description of the issues and their seamless solution within this
framework is beyond the scope of this paper, we want to provide one important
example.

One notes that all inference is based on the response curve $\vec{C}(\cdot)$
which characterizes the pixel's response to depth.
In the physical camera, due to various optical and semiconductor effects, this
response curve varies between sensor elements, and this variation is smooth
with the position of the pixel on the image sensor.
As a result, instead of having a single curve $\vec{C}(\cdot)$ as we described
so far, we actually have a set of response curves $\vec{C}_{x,y}(\cdot)$, one
for each pixel in the image.
Using the regression tree framework, we had a simple seamless solution for
this issue as follows:
\begin{itemize}
\item During training, instead of sampling responses from a single curve
$\vec{C}(\cdot)$, we sample responses from multiple response curves
corresponding to different parts of the image.
To obtain the label $\hat{t}(\vec{R}_i)$, use slow inference with the actual (position dependent)
curve from which $\vec{R}_i$ was sampled.
\item We augment the feature vector to include pixel position in addition to
the response $\vec{R}$.
\item We extend the leaf model and add linear terms in pixel coordinates $x$
and $y$.
\item Train the regression tree as usual.
\item During runtime: just add pixel position to the feature vector used to traverse the tree
\end{itemize}

This example serves to show the added benefit of a flexible regression
mechanism in extending the model to solve new and unexpected problems.

\section{Model Checking and P-Values}
This discussion provides additional background for our choice of the posterior
predictive p-value used in the main paper.
P-values are highly controversial in the field of statistics, in particular
for formal hypothesis testing.  For example,
in~\cite{meng1994posteriorpredictivepvalue} Xiaoli Meng writes about the
p-value,
\begin{quote}
``There is perhaps no single notion in statistics, other than the $p$-value,
that has been so widely used and yet so seriously criticized for so long.''
\end{quote}
Many works have discussed this controversy and Berger~\cite{berger2003testing}
provides a nice formal summary of the issues of disagreement.

For models that are fully observed the choice of a test statistic
unambiguously defines the p-value.
However, our model involves unobserved quantities, the imaging conditions $t$,
$\rho$, and $\lambda$, as well as $t_2$ and $\rho_2$ for the two-path model.
In this case, there is no single p-value to be used and this case is known as
``composite null hypothesis'' or models with ``nuisance parameters''.
Also, in this case the test statistic in the classical p-value generally does
not depend on the nuisance parameter, which is a drawback as it limits the
choice of useful test statistics.

The posterior predictive
p-value~\cite{meng1994posteriorpredictivepvalue,bayarri2000compositenull}
addresses both problems by integrating the test statistic over the Bayesian
posterior of the unknown variables.
As test statistic we choose the likelihood of the observation given the
unknown parameters, yielding equation~(\ref{eqn:gamma-exp}) in the main paper.

\begin{wrapfigure}[20]{r}{0.45\linewidth}
	\vspace{-0.35cm}%
	\begin{center}%
	\includegraphics[width=0.99\linewidth]{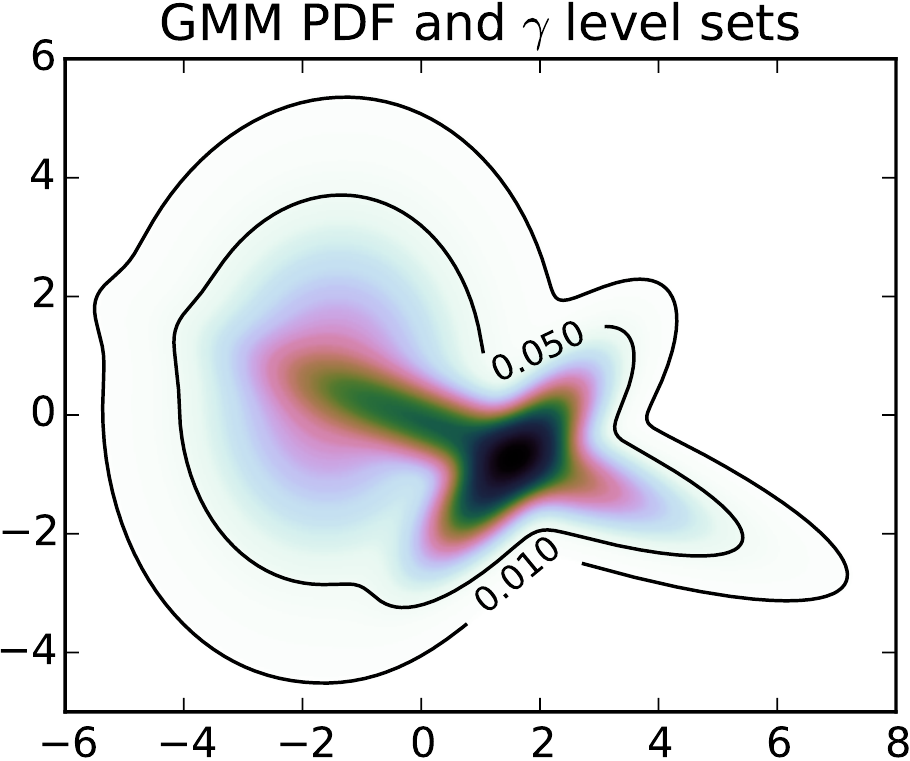}%
	\end{center}%
	\caption{\footnotesize Invalidation score example.
For a Gaussian mixture model level sets at significance levels
$\gamma \in \{0.01, 0.05\}$ separate the total probability
mass such that outside the level set the integrated probability is $\gamma$.
Samples outside these sets are rejected.}%
	\label{fig:gamma-gmm}%
\end{wrapfigure}

It is easier to understand the usefulness of the likelihood as a test
statistic on a model that is fully observed.
We visualize such an example with a simple Gaussian mixture model in
Figure~\ref{fig:gamma-gmm}.
For the case with unobserved variables the situation is similar except that
the distribution changes as a function of the observation $\vec{R}$.

\setlength{\columnsep}{0.5cm}%

The posterior predictive p-value has known drawbacks, analyzed
in~\cite{bayarri2000compositenull,robins2000asymptoticpvalue}.  In particular
it is known that it can be too conservative in rejecting the null hypothesis.
In our application this implies that we may not detect all detectable
deviation from the assumed model.
Intuitively the reason for this lack of power is that the p-value is not
Bayesian and actually does use the observation twice, once to define the
posterior, and once in the computation that defines the p-value, leading to
overly optimistic agreement with the model.
The so called \emph{partial} posterior predictive
p-value~\cite{bayarri2000compositenull} does successfully address this issue
but it is much more difficult to compute; in fact, although desirable, we have
not found a practical method to compute it in our application.

\section{Exposure Profiles Design Details}
To optimize the design objective~(\ref{prob:wpdesign}) in the main paper, we use a simulated
annealing approach as follows.
Let us abbreviate the objective function~(\ref{prob:wpdesign}) as
\begin{equation}
	f(\Z) = \E_{t,\rho,\lambda} \: \E_{\vec{R} \sim P(\vec{R} | t,\rho,\lambda,\Z)}
		\left[\ell(\hat{t}(\vec{R}), t)\right].
\end{equation}
We introduce an auxiliary Gibbs distribution, parametrized by a
\emph{temperature} $T > 0$,
\begin{equation}
	r(\Z,T) \propto \exp\left(-\frac{1}{T} f(\Z)\right).
\end{equation}
We use a sequence of temperature parameters that is slowly decreased for a
finite number of steps, that is, $T_0 > T_1 > \dots > T_K$,
starting from an initial temperature $T_0 = T_{\textrm{start}}$ down to a
final temperature $T_K = T_{\textrm{final}}$.
The smaller $T$ gets, the more peaked the distribution $r(\cdot,T)$ becomes
around the minimum of $f$.
Given a Markov chain sampler on $r$, this approach converges to the global
minimum of $f$.

We first discuss the Markov chain that we use, then give details about the
temperature schedule.

\subsection{Markov Chain}
To account for the sparsity constraints on $\Z$, our Markov chain uses an
augmented state space~\cite{liu2001montecarlo} to avoid measure-theoretic
difficulties of asserting reversibility in the context of changing
dimensionality~\cite{green1995reversiblejump}.

We decompose $\Z$ into a binary matrix $\B \in \{0, 1\}^{m \times n}$ and a
value matrix $\VM \in \mathbb R^{m \times n}$ with $Z_{ji} = B_{ji} V_{ji}$.
This allows us to easily set weights to zero by setting $B_{ji} = 0$ and have
the reversible proposal readily available by setting $B_{ji} = 1$.
Our MCMC sampler is a reversible Metropolis-Hastings sampler and consists of
the following transition kernels (moves):
\begin{enumerate}
\item \emph{Move mass:}
Choose two matrix entries $V_{ji}$, $V_{kl}$ randomly (uniform) and move a
uniformly sampled value from one entry to another such that their total value
stays the same and both are still positive.  This kernel is reversible with
itself.
\item \emph{Swap values:}
Choose two matrix entries $W_{ji}$, $W_{ji}$ randomly (uniform) and swap their
values $V$ and binary indicator value $B$.  This kernel is reversible with
itself.
\item \emph{Set a weight to zero:}
Choose a matrix entry with $B_{ji} = 1$ randomly (uniform) and set it to zero.
This kernel is reversible with the following kernel.
\item \emph{Set a weight to nonzero:}
Choose a matrix entry with $B_{ji} = 0$ randomly (uniform) and set its binary
indicator value to one.  This kernel is reversible with the previous
set-to-zero kernel.
\item \emph{Perturb weight value:}
Choose a matrix entry $V_{ji} = 0$ randomly (uniform) and rescale its value
with a log-normal sampled factor.  This kernel is reversible with itself.
\item \emph{Scale all weight values:}
Rescale all values $\VM$ with a log-normal sampled scalar.
This kernel is reversible with itself.
\end{enumerate}

The above kernels are combined with the following probabilities:
20\% for the \emph{move mass} kernel;
20\% for the \emph{swap values} kernel;
10\% for the \emph{set-to-zero} and \emph{set-to-nonzero} kernels, each;
30\% for the \emph{perturb weight} kernel;
10\% for the \emph{global scaling} kernel.

\subsection{Temperature Schedule}
\begin{figure}
\centering\includegraphics[width=0.99\linewidth]{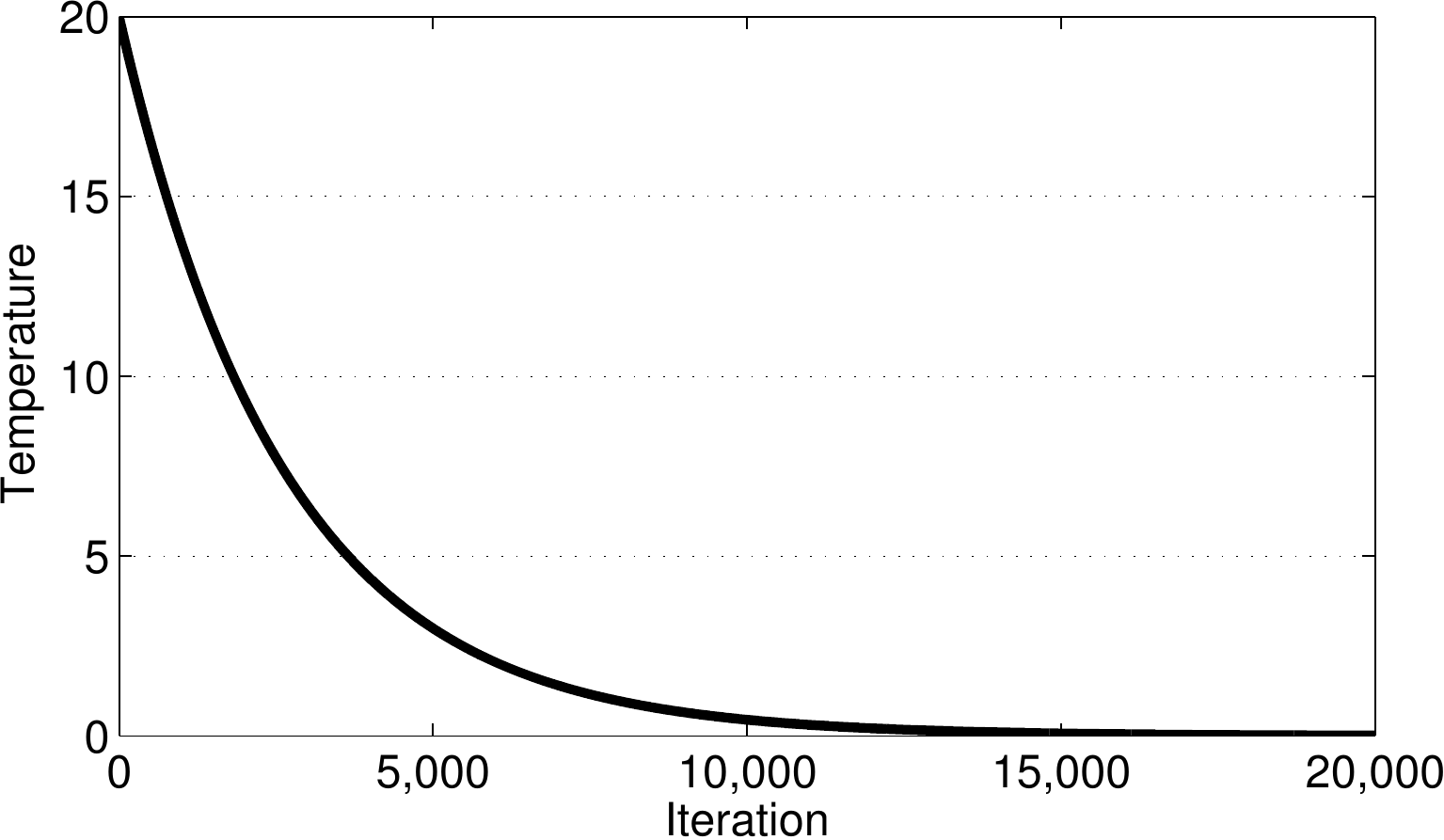}
\caption{Simulated annealing schedule used during shutter profile design
optimization, here with $K=20,000$ iterations.}
\label{fig:sa-schedule}
\end{figure}

For simulated annealing we use a geometric temperature
schedule~\cite{kirkpatrick1983simulatedannealing}, with the temperature
at iteration $k$ being
	\[T_k = T_{\textrm{start}} \: \beta^k,\]
where we use the initial temperature $T_{\textrm{start}}=20$ and a target
temperature of $T_{\textrm{final}}=0.01$, so that
	\[\beta = \exp\left(\frac{1}{K} \left[\log T_{\textrm{final}}
		- \log T_{\textrm{final}}\right]\right).\]
This leads to the schedule as shown in Figure~\ref{fig:sa-schedule}.
We typically use a $K=20,000$ or $K=100,000$ iterations.

\section{Time of Flight Simulation Details}
We now discuss details of the physically accurate light simulation that we use
to simulate multipath phenomena.
First we recap the basis of both the bidirectional path tracer (BDPT) and the
Metropolis light transport (MLT)
algorithms~\cite{veach1995bidirectional,veach1997mlt} and then provide
information about the variance reduction techniques we use.

\subsection{Light Transport Formulation}
Assuming a geometric light model where light travels in staight lines and only
interacts with surfaces, the measured light intensity at a pixel in a static
scene without active illumination can be formulated as a path integral. This
integral accumulates the intensity from light paths $x_0, x_1 \dots x_{k+1}$ that start in a point
$x_0$ on an emitting surface and end in a point $x_{k+1}$ on the pixel's
sensor surface. The intermediate nodes of this path $x_1 \dots x_k$ are surfaces in the scene.
The integral can be formulated (see \cite{veach1997mlt} for details) as
\begin{align}
\sum_{k=0}^\infty \int_{\mathcal M^{k+1}}
&  L_e(x_0 \rightarrow x_1) G(X_0 \leftrightarrow x_1)\nonumber \\
&\prod_{i=1}^k \left( f(x_{i-1} \rightarrow x_i \rightarrow x_{i+1}) G(x_i \leftrightarrow x_{i+1}) \right) \nonumber \\
& L_s (x_k \rightarrow x_{k+1})\,\textrm{d}A(x_0) \dots \textrm{d} A(x_{k+1}). \label{eq:mlt_integral}
\end{align}
In this equation,
\begin{itemize}
\item $\mathcal M$ is the set of all surfaces in the scene including emitters
    and the camera sensor and $A$ is the area measure on $\mathcal M$;
\item $L_e(x_0 \rightarrow x_1)$ is a function representing emitters. It is
    proportional to the light that is emitted from point $x_0$ in the direction
    of $x_1$. It takes only non-zero values if $x_0$ is on emitter surfaces;
\item $L_s(x_k \rightarrow x_{k+1})$ is the equivalent of $L_e$ for the sensor.
    $L_s$ specifies how sensitive the sensor is for photons arriving at
    $x_{k+1}$ from the direction of $x_{k}$.
\item $f(x_{i-1} \rightarrow x_i \rightarrow x_{i+1})$ is the bidirectional
    scattering distribution function (BSDF) describing how much light is
    scattered at surface point $x_i$ in direction $x_{i+1}$ of an incoming ray
    from the direction of $x_{i-1}$;
\item $G(x_i \leftrightarrow x_{i+1}) = V(x_i \leftrightarrow x_{i+1}) \frac {|
    \cos \phi_i \cos \phi_{i+1} |}{\| x_{i+1} - x_{i}\|^2}$ is the throughput
    of a differential beam between $\textrm{d}A(x_i)$ and
    $\textrm{d}A(x_{i+1})$.  $V(x_i \leftrightarrow x_{i+1})$ is an indicator
    function for mutual visibility of $x_i$ and $x_{i+1}$, which means $V$ is
    zero if the direct path between the two inputs is blocked, otherwise $1$;
    The variables $\phi_i$, $\phi_{i+1}$ denote the angle between the beam and
    the surface normals at $x_i$ and $x_{i+1}$.
\end{itemize}

The observed response in a specific pixel of our time-of-flight camera from the
emitted light pulse can be modelled by extending the path integral formulation
above to
{\small \begin{align}
        R_\textrm{active} = \int &  \sum_{k=0}^\infty \int_{\mathcal M^{k+1}}  P(u) L_e(x_0 \rightarrow x_1) G(X_0 \leftrightarrow x_1)\nonumber \\
       &\prod_{i=1}^k \left( f(x_{i-1} \rightarrow x_i \rightarrow x_{i+1}) G(x_i \leftrightarrow x_{i+1}) \right) \nonumber \\
       & L_s (x_k \rightarrow x_{k+1}) S_j(u + t_l) \nonumber \\
&\textrm{d}A(x_0) \dots \textrm{d} A(x_{k+1}) \textrm{d} u.
\end{align}}
We additionally integrate over time $u$ and include the intensity of the
emitted pulse $P(t)$ as well as the shutter function $S_j(t + t_l)$.
The time delay $t_l = c \: l $
of emitted light arriving at the sensor is the total path length,
	\[l = \sum_i \| x_{i+1} - x_i\|,\]
times the speed of light $c$.
All terms involving time can be group together into the expression
\[
   \int P(u) S_j(u + t_l) \,\textrm{d}u = \frac {C_j(t_l)}{d(t_l)}
\]
that only depends on the time delay $t_l$ corresponding to total path length.
It corresponds to the curve $C_j$ without the decay of light $d(t_l)$ due to
distance $l$ (The decay of light is already accounted for in the $G$ terms of
the integral).  The measured response is then
{\small \begin{align}
        R_\textrm{active} = &  \sum_{k=0}^\infty \int_{\mathcal M^{k+1}}  \frac{C_j(t_l)}{d(t_l)} L_e(x_0 \rightarrow x_1) G(X_0 \leftrightarrow x_1)\nonumber \\
       & \quad \prod_{i=1}^k \left( f(x_{i-1} \rightarrow x_i \rightarrow x_{i+1}) G(x_i \leftrightarrow x_{i+1}) \right) \nonumber \\
& \quad L_s (x_k \rightarrow x_{k+1}) \nonumber \\
& \quad \textrm{d}A(x_0) \dots \textrm{d} A(x_{k+1}).
\end{align}}
This formulation is identical to the path integral
Equation~\eqref{eq:mlt_integral} but with the additional $C_j(t_l) / d(t_l)$
term.

We modified the bidirectional path tracer (BDPT)
algorithm~\cite{veach1995bidirectional} and the Metropolis light transport
(MLT) algorithm~\cite{veach1997mlt} in the Mitsuba renderer \cite{wenzel2010mitsuba}
to produce a weighted set of samples $\{(w_i,L_i,t_i)\}_{i=1,\dots,N}$ of the
path integral in Equation~\eqref{eq:mlt_integral}.  The weight of the path
sample is $w_i$, $L_i$ is the number of edges and $t_i$ is the time
corresponding to the total path distance. We can generate samples of
$R_\textrm{active}$ by
\[
\sum_{i=1}^N \frac{w_i}{d(t_i)} C_j(t_i).
\]
Considering all shutters $C_1, C_2, \dots$ and adding constant ambient light
$\tau$ to account for $R_\textrm{ambient}$, we may obtain realistic estimates of the
mean response vector
\begin{equation}
	\vec{\mu} = \tau \vec{A} + \sum_{i=1}^N \frac{w_i}{d(t_i)} \vec{C}(t_i).
\end{equation}

\subsection{Variance Reduction}
The BDPT and MLT rendering techniques are Monte Carlo methods and therefore
estimates obtained from them will have a Monte Carlo variance.
This variance does not originate with the underlying mechanism that is being
simulated, but is due to the finite number of samples that are used for
estimation.
Whereas normal light transport rendering in computer graphics applications is
targeted at estimating mean intensities in three spectral bands (RGB), we are
instead interested in the time-of-flight density.
Because it is a function instead of a small number of values, it is more
difficult to obtain reliable estimates of this function.

To improve the accuracy of our estimate with the given time and memory
constraints, we use two variance reduction techniques: \emph{stratification}
and \emph{priority sampling}.

The starting point for both methods is a stream of weighted samples
$(w_i,L_i,t_i)$ being generated for each pixel.

\subsubsection{Stratification}
Stratification is a classic variance reduction technique based on prior
knowledge of subpopulations which have lower within-population variation.
It works by breaking up the estimation problem into one estimation problem per
subpopulation and combining the individual estimates into one joint estimate.
This reduces the variance of the joint estimate compared to lumping all
subpopulations together in only one population and sampling and estimating
from only this one population~\cite[Section 5.5]{rubinstein2011simulation}.

We \emph{stratify} the incoming stream of samples into two sets.
The first stratum is the set of samples $L_i=2$ and the second stratum is the
set of samples for which $L_i > 2$.  For both sets we keep an equal number of
samples, typically a few thousand.  The following priority sampling is then
performed on each of these two sets separately.

\subsubsection{Priority Sampling}
The output of the simulation is a set of path samples $(w_i,L_i,t_i)$ for each
pixel.  For large image sizes this can require tens of gigabytes of storage.
In particular for the MLT sampler many of these samples contain
partially redundant information due to correlated sampling via runnign a
Markov chain, and storing all of them is wasteful in terms
of storage.  Because for MLT the samples are correlated in time, one really
does need to generate a large number of samples to get good estimates; it is
only the storage that is wasteful.
For BDPT the samples are uncorrelated but we can still
improve estimates by replacing low-weight samples with
more important ones due to the inefficiencies of importance sampling.

Typically, in general Markov chain Monte Carlo simulations the samples are
unweighted and one can simply \emph{thin} the samples by taking, for example,
every 10th sample only, or by using reservoir sampling to keep a random
subset.
Here, however, the samples from both BDPT (importance sampling) and MLT
(Markov chain simulations) are weighted, and this naive strategy---while still
valid in terms of providing an unbiased estimate---yields a high variance
estimate because it discards important samples with high weights.

To obtain low-variance estimates from few samples we use \emph{priority
sampling}~\cite{duffield2007prioritysampling}, a close to optimal method
addressing the above subsampling problem.
Intuitively, priority sampling generalizes reservoir sampling to the case of
weighted samples.
It processes the input sample stream one sample at a time and keeps a fixed
number of samples with adjusted weights.  The weights are adjusted such that
the estimate of any subset sum is unbiased, and the variance of weight subset
sums is almost optimal uniformly over all possible subsets.

We use priority sampling to thin the two sample streams for each stratum and
after rendering is finished we simply output the kept samples and adjusted
weights.

Overall we found that the bidirectional path tracer (BDPT) often produces
better results with lower variance and all simulation results in the main
paper are obtained by running BDPT with 8192 samples per pixel.

\section{Noise Model Validation}
\begin{figure}[t!] \centering
\includegraphics[width=0.99\linewidth]{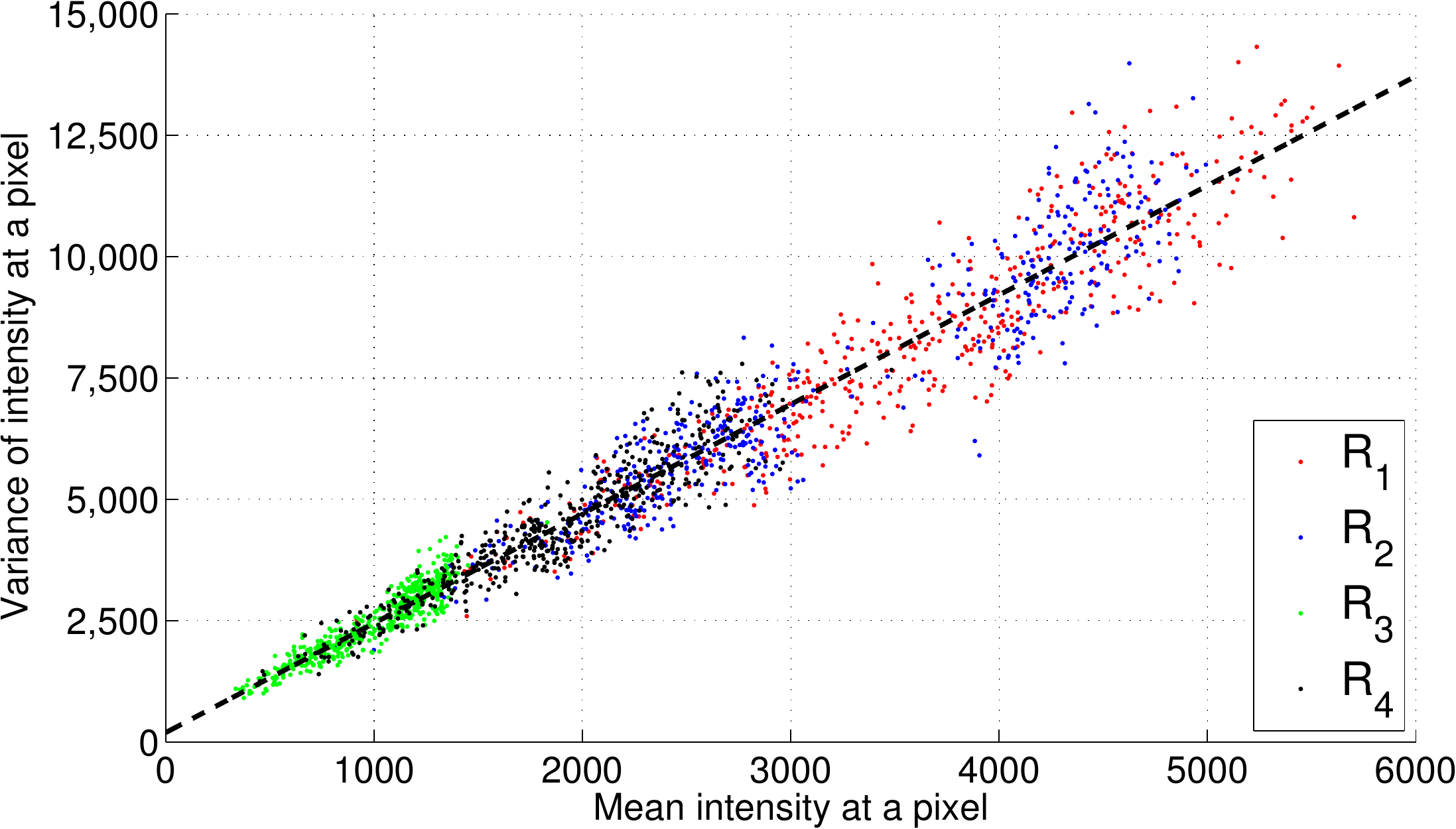}%
	\caption{Validation of the noise model~(\ref{generative_cov}).
The variance of the response is approximately a linear function of the mean
response~\cite{DBLP:journals/tip/FoiTKE08}.  We show samples from four
responses in different colors and superimpose the least squares fit.}
	\label{fig:noisemodel}%
\end{figure}
To verify the noise model we assumed in Section~\ref{prob_model} (equation~\ref{generative_cov}) of the main paper,
we used
the experimental setup  as described in Section~\ref{sec:exp-depthuncertainty}
of the main paper:
we sample 500 random pixels and capture 200 frames from a static scene.
We then measure the variance of each pixel's response, as well as estimate the
mean response; this provides empirical data about the actual noise present in
the input signal.

Figure~\ref{fig:noisemodel} shows the noise model results
in the form of a scatter plot of the variance of responses versus their mean.
The data clearly validates the assumed noise model~(\ref{generative_cov}) from
the main paper, and shows that the signal-dependent Poissonian shot noise
component dominates except for very small intensities.

\section{The scene used as an extended generative model}

\begin{figure}[t!] \centering
\includegraphics[width=0.99\linewidth]{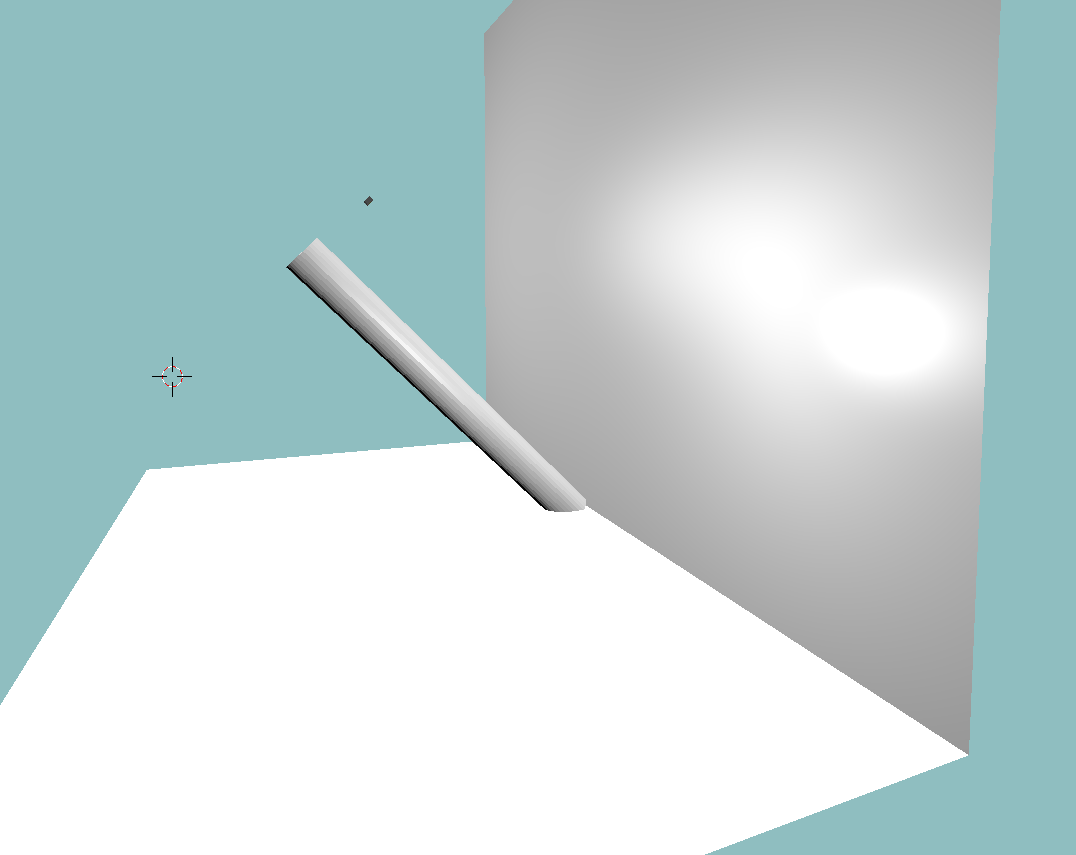}%
	\caption{The scene used for designing a multipath-resistant exposure profile.}
	\label{fig:supp_mp_scene}%
\end{figure}

As described in Section~\ref{sec:mp-robust} in the main paper, we may use a realistic light transport simulation as a more complex generative model $Q$
in order to generate responses containing also multipath components. We used this method to design a multipath-resistant exposure profile. The scene we used
is depicted in Figure~\ref{fig:supp_mp_scene}. The camera is pointed towards a reflective wall,
and the responses sampled from the cylinder and the floor contain both a direct component and multipath components. We used two copies of this scene at two
different scales.
The exposure profile designed using this scene  was tested on a very different test scene as described in
Section~\ref{sec:wpdesign-experiments}.

\section{Additional Rendered Results}
In Section~\ref{sec:exp-sim} of the main paper we omitted three scenes for
space reasons; the results are provided here in
Figure~\ref{fig:75431breakfast}, Figure~\ref{fig:70272cocina},
and Figure~\ref{fig:77668staircase} and qualitatively agree with the results
shown in the main paper.

\maslulimresultfigure[Significant multipath error reductions due to the
two-path model are visible (wall, table, floor, chairs).
Specular surfaces (lamp shade) remain problematic.]{figures/exp-maslulim}{75431breakfast}{``The Breakfast Room'' by Wig42}
\maslulimresultfigure{figures/exp-maslulim}{70272cocina}{``Kitchen Nr 2'' by oldtime}
\maslulimresultfigure{figures/exp-maslulim}{77668staircase}{``The Wooden Staircase'' by Wig42}

\subsection{Example Problem with $\gamma$}
We now give an example where we suffer high depth errors but have no
operational method to recognize this: the estimated uncertainty
$\hat{\sigma}$ is not overly large, and the invalidation score $\gamma$ does
not indicate deviation from the model.

\begin{figure}[t!]%
	\centering%
	\subfigure[Problematic area with high depth error (see
Fig.~\ref{fig:70272cocina-tpmap-terr}) that is not recognized by
$\hat{\sigma}$ (see Fig.~\ref{fig:70272cocina-tpmap-sigma}), nor by $\gamma$
(see Fig.~\ref{fig:70272cocina-tpmap-gamma}).]{%
		\includegraphics[width=0.425\linewidth]{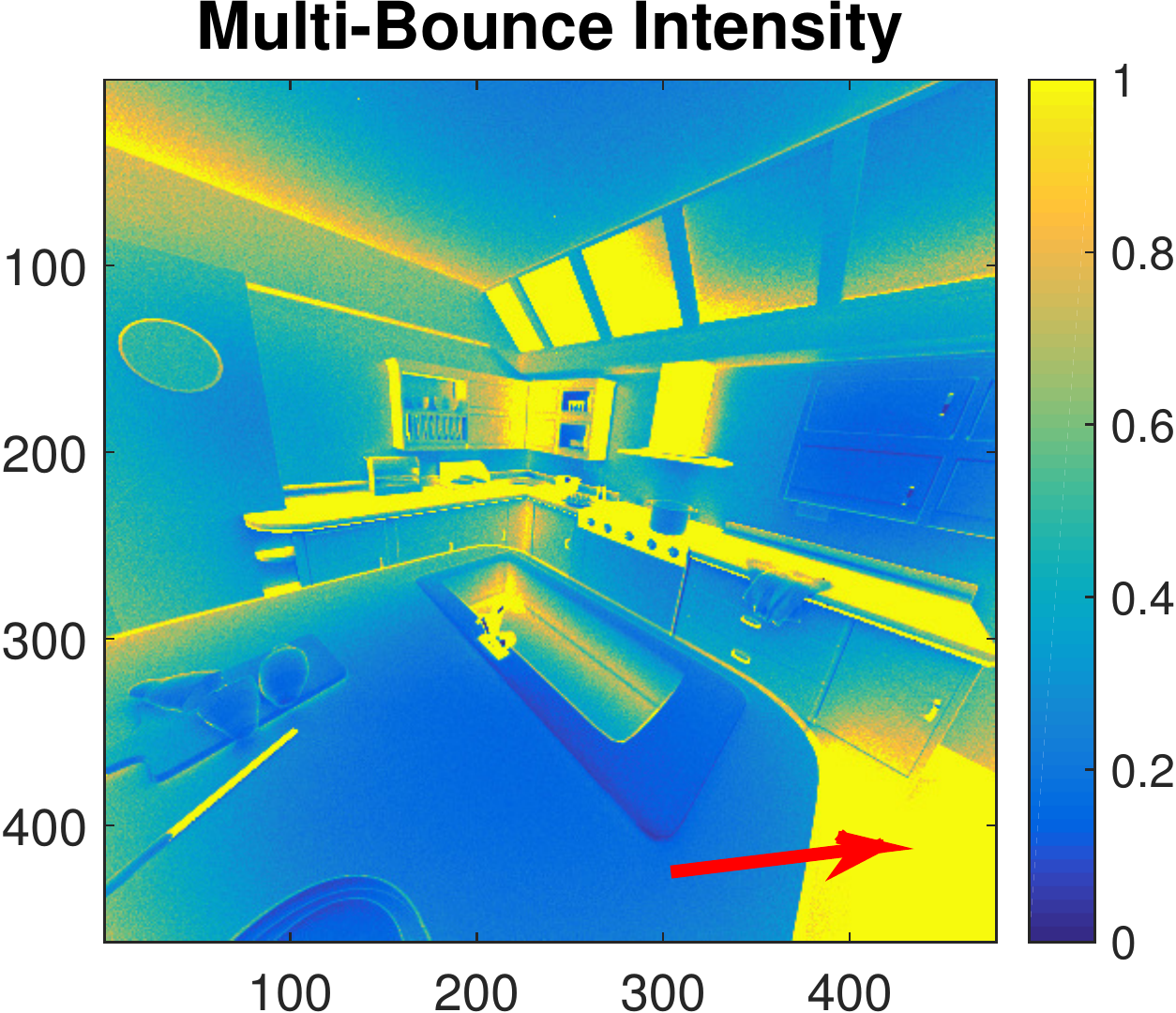}\label{fig:cocina2-mpratio-highlight}%
	}\hfill%
	\subfigure[Intensity measure at the highlighted pixel showing complex
multipath phenomena with more than two-path responses and large diffuse
multipath.]{%
		\includegraphics[width=0.525\linewidth]{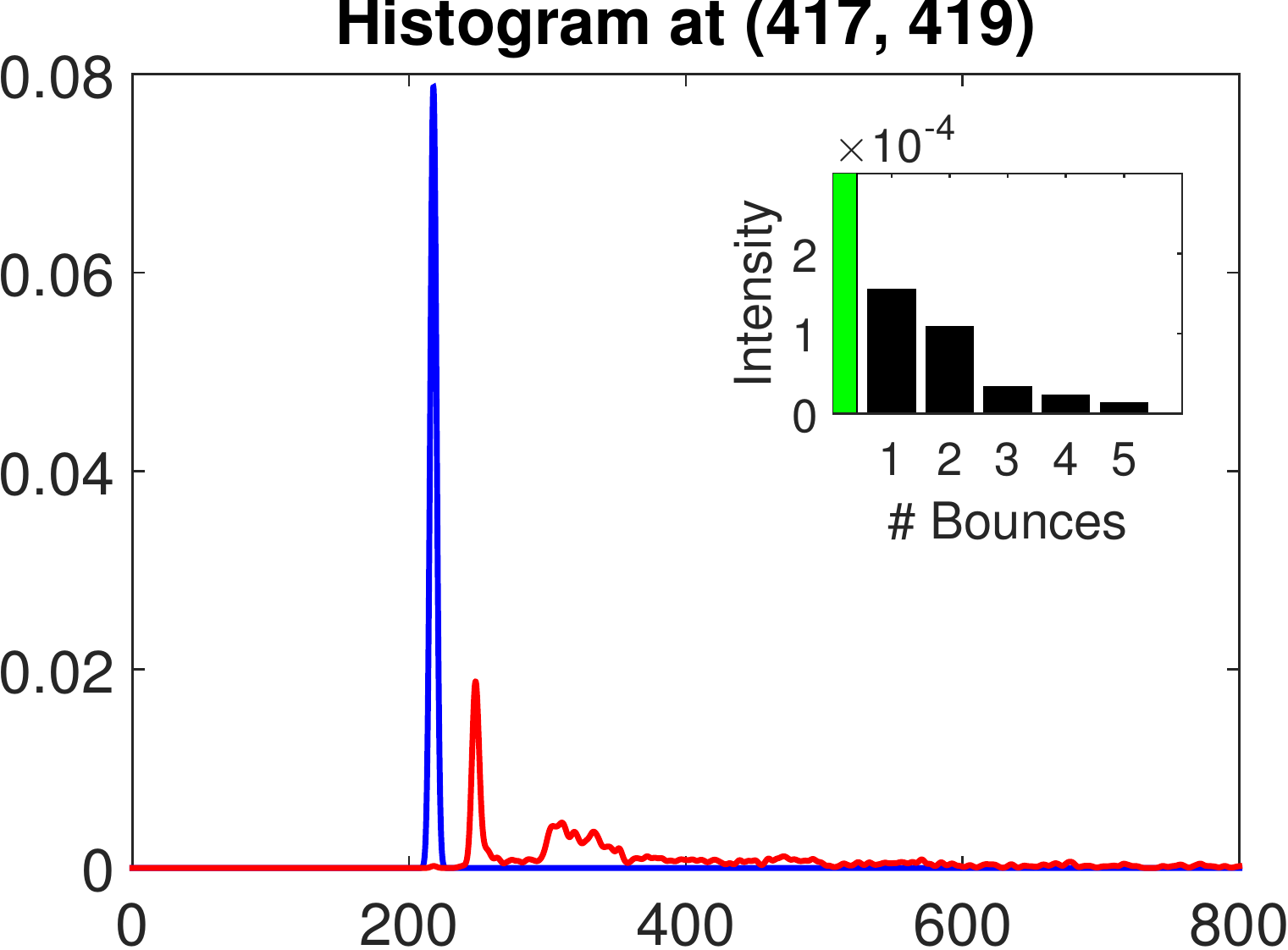}\label{fig:cocina2-dh-pixel}%
	}%
\vspace{-0.2cm}%
\caption{Explaining a failure of $\hat{\sigma}$ and $\gamma$ to recognize
areas of large depth errors.}
\label{fig:cocina2-problem}%
\end{figure}

The situation occurs in the scene shown in Figure~\ref{fig:70272cocina}, and
we highlight the area in Figure~\ref{fig:cocina2-problem}.
In essence the problem is due to complex multipath phenomena involving diffuse
multipath and multiple bounces as can be seen from
Figure~\ref{fig:cocina2-dh-pixel}, which leads to an observed response that is
well within a high-probability region of the assumed two-path model (no
invalidation) and has a strong direct response component (leading to low
$\hat{\sigma}$).

In order to improve depth accuracy in regions such as the highlighted one,
several approaches are relevant.
%
%
We are currently considering temporal integration
of the observed response and an imaging model that does not assume conditional
independence among different sensor elements.  For example, many surfaces are
planar and recognizing deviations from planarity over multiple pixels could
potentially provide a strong cue to recognize and correct for multipath
interference.

\end{appendices}

\end{document}